\pdfoutput=1

\documentclass[11pt]{article}

\usepackage{emnlp2021}

\usepackage{times}
\usepackage{latexsym}

\usepackage[T1]{fontenc}

\usepackage[utf8]{inputenc}

\usepackage{microtype}

\usepackage{microtype}
\usepackage{times}
\usepackage{latexsym}
\usepackage{xcolor}
\usepackage[show]{inotes}
\usepackage{amsmath,amssymb,amsfonts}
\usepackage{algorithmic}
\usepackage{graphics}
\usepackage{graphicx}
\usepackage{textcomp}
\usepackage{xcolor}
\usepackage{multirow}
\usepackage{pbox}
\usepackage{booktabs} 
\usepackage[normalem]{ulem}

\usepackage[normalem]{ulem}
\useunder{\uline}{\ul}{}
\usepackage{amssymb}
\usepackage{float}
\usepackage{amssymb}
\usepackage{enumitem}
\usepackage{listings}
\usepackage{xstring}
\usepackage{graphicx}
\usepackage{pbox}
\usepackage{subcaption}
\usepackage{epstopdf}
\usepackage{xstring}

\usepackage{balance}
\usepackage{pbox}
\usepackage{multirow}
\usepackage{enumitem}

\usepackage{multicol}

\usepackage{tabularx}
\usepackage{tcolorbox}
\usepackage{color, colortbl}
\usepackage[show]{inotes}
\usepackage{booktabs}
\usepackage{multirow}
\usepackage[switch]{lineno}
\usepackage[maxfloats=256]{morefloats}

\usepackage{comment}

\newcommand{\lbl}[1]{\textbf{\texttt{{#1}}}}

\newcommand*{\img}[1]{%
   \raisebox{-.3\baselineskip}{%
       \includegraphics[
       height=1.2 cm, 
       width=1.2 cm, 
       keepaspectratio,
       ]{#1}%
   }%
}

\title{Fighting the COVID-19 Infodemic:\\
Modeling the Perspective of Journalists, Fact-Checkers,\\ Social Media Platforms, Policy Makers, and the Society}

\author{Firoj Alam,\textsuperscript{\rm 1}  Shaden Shaar,\textsuperscript{\rm 1}  Fahim Dalvi,\textsuperscript{\rm 1} Hassan Sajjad,\textsuperscript{\rm 1} Alex Nikolov,\textsuperscript{\rm 2}  \\ 
{\bf Hamdy Mubarak,\textsuperscript{\rm 1}  Giovanni Da San Martino,\textsuperscript{\rm 3}  Ahmed Abdelali,\textsuperscript{\rm 1} Nadir Durrani,\textsuperscript{\rm 1}  } \\ 
{\bf Kareem Darwish,\textsuperscript{\rm 1}  Abdulaziz Al-Homaid,\textsuperscript{\rm 1} Wajdi Zaghouani,\textsuperscript{\rm 4}  Tommaso Caselli,\textsuperscript{\rm 5}  } \\
{\bf Gijs Danoe,\textsuperscript{\rm 5} Friso Stolk,\textsuperscript{\rm 5} Britt Bruntink\textsuperscript{\rm 5} and Preslav Nakov}\textsuperscript{\rm 1} \\
  \textsuperscript{\rm 1} Qatar Computing Research Institute, HBKU, Qatar, 
  \textsuperscript{\rm 2} Sofia University, Sofia, Bulgaria\\
  \textsuperscript{\rm 3} University of Padova, Italy, 
  \textsuperscript{\rm 4} Hamad Bin Khalifa University, Qatar\\
  \textsuperscript{\rm 5} University of Groningen, The Netherlands\\
  \{fialam, faimaduddin, hsajjad, hmubarak, aabdelali, ndurrani\}@hbku.edu.qa, \\
   \{abalhomaid, kdarwish, pnakov\}@hbku.edu.qa,\\
  alexnickolow@gmail.com, dasan@math.unipd.it, wzaghouani@hbku.edu.qa\\
  t.caselli@rug.nl, \{g.danoe, b.m.bruntink, f.r.p.stolk\}@student.rug.nl
}

\begin{document}
\maketitle
\begin{abstract}
With the emergence of the COVID-19 pandemic, the political and the medical aspects of disinformation merged as the problem got elevated to a whole new level to become \emph{the first global infodemic}. Fighting this infodemic has been declared one of the most important focus areas of the World Health Organization, with dangers ranging from promoting fake cures, rumors, and conspiracy theories to spreading xenophobia and panic. Addressing the issue requires solving a number of challenging problems such as identifying messages containing claims, determining their check-worthiness and factuality, and their potential to do harm as well as the nature of that harm, to mention just a few. To address this gap, we release a large dataset of 16K manually annotated tweets for fine-grained disinformation analysis that (\emph{i})~focuses\ on COVID-19, (\emph{ii})~combines the perspectives and the interests of journalists, fact-checkers, social media platforms, policy makers, and society, and (\emph{iii})~covers Arabic, Bulgarian, Dutch, and English. Finally, we show strong evaluation results using pretrained Transformers, thus confirming the practical utility of the dataset in monolingual \textit{vs.} multilingual, and single task \textit{vs.} multitask settings. 
\end{abstract}

\section{Introduction}
\label{sec:introduction}

The rise of social media has made them one of the main channels for information dissemination and consumption. As a result, nowadays, many people rely on social media as their primary source of news \citep{perrin2015social}, attracted by the broader choice of information sources and by the ease for anybody to become a news producer. 

Unfortunately, the democratic nature of social media has raised questions about the quality and the factuality of the information that is shared on these platforms. Eventually, social media have become one of the main channels to spread disinformation. 

\begin{figure*}[h]
\centering
\includegraphics[width=\textwidth]{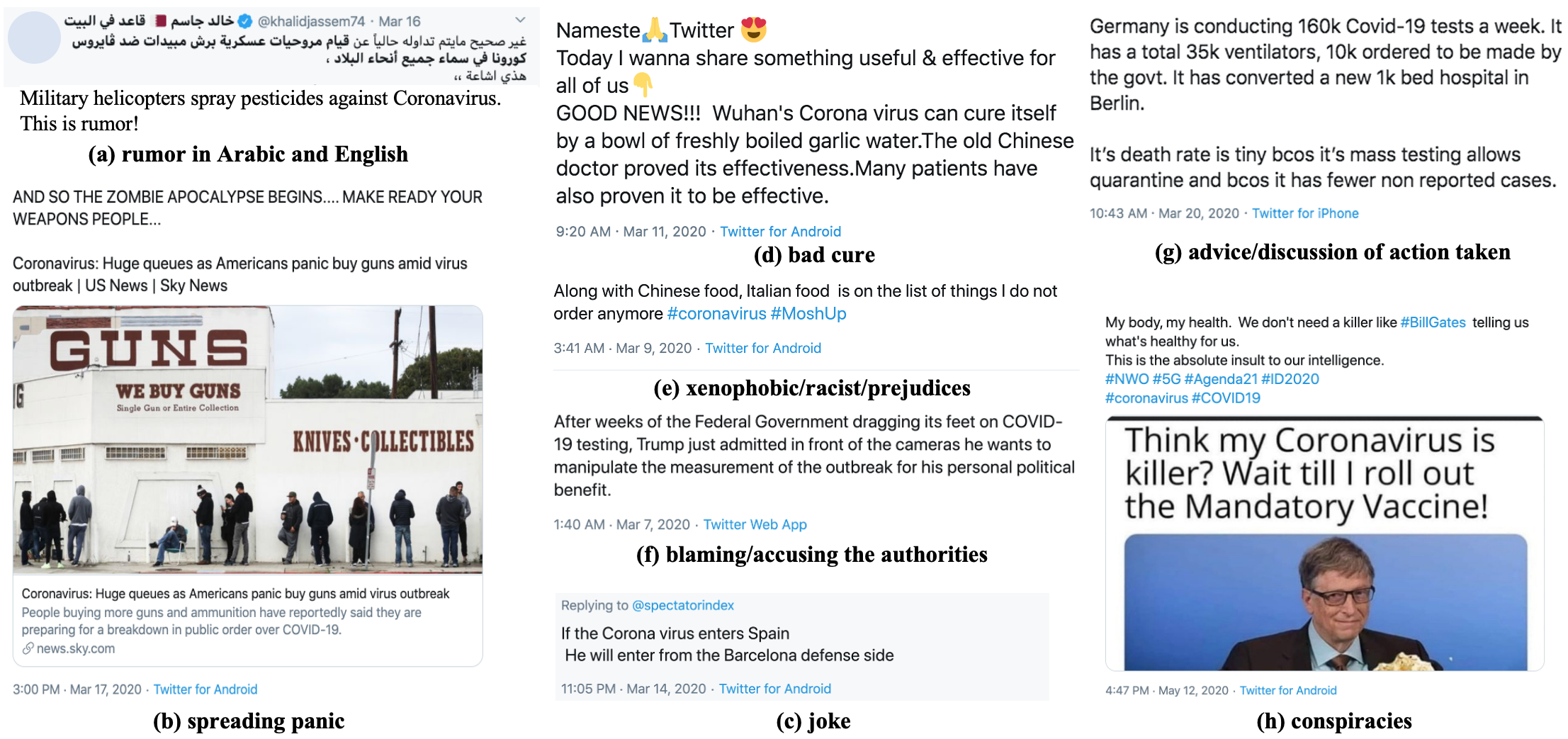}
\caption{Examples of tweets that would be of potential interest to journalists, fact-checkers, social media platforms, policy makers, government entities, and the society as a whole.}
\label{fig:tweet_examples}
\end{figure*}

Figure \ref{fig:tweet_examples} demonstrates how online users discuss topics related to COVID-19 in social media. We can see that the problem goes beyond factuality: there are tweets spreading rumors (Figure 1a), instilling panic (Figure 1b), making jokes (Figure~1c), promoting fake cures (Figure 1d), spreading xenophobia, racism, and prejudices (Figure 1e), or promoting conspiracy theories (Figure 1h). 

Other examples in Figure \ref{fig:tweet_examples} contain information that could be potentially useful and might deserve the attention of government entities. For example, the tweet in Figure 1f blames the authorities for their inaction regarding COVID-19 testing. The tweet in Figure 1g is useful both for policy makers and for the general public as it discusses action taken and suggest actions that probably should be taken elsewhere to fight the pandemic.

For the tweets in Figure \ref{fig:tweet_examples}, it is necessary to understand whether the information is correct, harmful, calling for action to be taken by relevant authorities, etc. Rapidly sorting these questions is crucial to help organizations channel their efforts, and to counter the spread of disinformation, which may cause panic, mistrust, and other problems.

Addressing these issues requires significant effort in terms of (\emph{i})~defining comprehensive annotation guidelines, (\emph{ii})~collecting tweets about COVID-19 and sampling from them, (\emph{iii})~annotating the tweets, and (\emph{iv})~training and evaluating models.
Given the interconnected nature of these issues, it is more efficient to address them simultaneously. 

With this consideration in mind, we adopt a \emph{multifaceted} approach, which is motivated by engaging with different stakeholders such as journalists and policy makers. We focused on three key aspects, which are formulated into seven questions: (\emph{i})~Check worthiness and veracity of the tweet (Q1-4 and Q5). (\emph{ii})~Harmfulness to society (Q6); and (\emph{iii})~Call for action addressing a government / policy makers (Q7).
Q1--Q5 were motivated by conversations with journalists and professional fact-checkers, while Q6-Q7 were formulated in conversations with a Ministry of Public Health.

Our contributions can be summarized as follows:
\begin{itemize}
        \item We develop a large manually annotated dataset of 16K tweets related to the COVID-19 infodemic in four languages (Arabic, Bulgarian, Dutch, and English), using a schema that combines the perspective of journalists, fact-checkers, social media platforms, policymakers, and the society.
        \item We demonstrate sizable performance gains over popular deep contextualized text representations (such as BERT), when using multitask learning, cross-language learning, and when modeling the social context of the tweet, as well as the propagandistic nature of the language used.
        \item We make our data and code freely available.\footnote{\url{https://github.com/firojalam/COVID-19-disinformation}}
\end{itemize}

\section{Related Work}
\label{sec:related_work}

\paragraph{Fact-Checking}
Research on fact-checking claims is largely based on datasets mined from major fact-checking organizations. Some of the larger datasets include the \emph{Liar, Liar} dataset of 12.8K claims from PolitiFact \citep{wang:2017:Short}, the \emph{ClaimsKG} dataset and system~\citep{ClaimsKG} of 28K claims from eight fact-checking organizations, the \emph{MultiFC} dataset of 38K claims from 26 fact-checking organizations \citep{augenstein-etal-2019-multifc}, 
and the 10K claims \emph{Truth of Various Shades} dataset \citep{rashkin-EtAl:2017:EMNLP2017}.
There have been also datasets for other languages, created in a similar fashion, e.g.,~for Arabic \citep{baly-EtAl:2018:N18-2,alhindi-etal-2021-arastance}.

A number of datasets were created as part of shared tasks. In most cases, they performed their own annotation, either (a)~manually, e.g.,~the SemEval tasks on determining the veracity of rumors \citep{derczynski-EtAl:2017:SemEval,gorrell-etal-2019-semeval}, propaganda detection in news articles and memes \cite{DBLP:conf/semeval/MartinoBWPN20,dimitrov2021detecting,SemEval2021-6-Dimitrov}, fact-checking in community question answering forums~\citep{mihaylova-etal-2019-semeval}, the CLEF CheckThat! lab on identification and verification of claims~\citep{clef2018checkthat:overall,CheckThat:ECIR2019,CheckThat:ECIR2020,clef-checkthat-en:2020,clef-checkthat:2021:LNCS,clef-checkthat:2021:task2,clef-checkthat:2021:task1}, or (b)~using crowdsourcing, e.g.,~the FEVER task on fact extraction and verification, focusing on claims about Wikipedia content~\citep{thorne-EtAl:2018:N18-1,thorne-etal-2019-fever2}. 

Unlike our work, the above datasets did not focus on tweets (they used claims from news, speeches, political debates, community question answering fora, or were just made up by human annotators; RumourEval is a notable exception), targeted factuality only (we cover a number of other issues), were limited to a single language (typically English; except for CLEF), and did not focus on COVID-19.

\paragraph{Check-Worthiness Estimation} 
Another relevant research line is on detecting check-worthy claims in political debates using manual annotations~\citep{Hassan:15} or by observing the selection of fact-checkers
~\citep{gencheva-EtAl:2017:RANLP,Patwari:17,NAACL2018:claimrank,RANLP2019:checkworthiness:multitask}.

\paragraph{COVID-19 Research}
There are a number of COVID-19 Twitter datasets: some unlabeled~\cite{info:doi/10.2196/19273,Banda:2020,haouari2020arcov19}, some automatically labeled with location information~\cite{abdul2020mega,Umair2020geocovid19}, some labeled using distant supervision~\cite{cinelli2020covid19,zhou2020repository}, and some manually annotated~\cite{song2020classification,vidgen2020detecting,shahi2020fakecovid,pulido2020covid,dharawat2020drink}.

There is also work on credibility \cite{cinelli2020covid19,pulido2020covid,zhou2020repository}, racial prejudices and fear \cite{Medford2020.04.03.20052936,vidgen2020detecting}, as well as situational information, e.g.,~caution and advice \citep{li2020characterizing}, as well as on detecting mentions and stance with respect to known misconceptions \cite{hossain-etal-2020-covidlies}.

The closest work to ours is that of \citet{song2020classification}, who
collected false and misleading claims about COVID-19 from IFCN Poynter, and annotated them as
(1)~Public authority,
(2)~Community spread and impact, 
(3)~Medical advice, self-treatments, and virus effects,
(4)~Prominent actors,
(5)~Conspiracies,
(6)~Virus transmission,
(7)~Virus origins and properties,
(8)~Public reaction, and
(9)~Vaccines, medical treatments, and tests.
These categories partially overlap with ours, but account for less perspectives. Moreover, we cover both true and false claims, we focus on tweets (while they have general claims), and we cover four languages.

Last but not least, we have described the general \emph{annotation schema} in previous work 
\cite{alam2020fighting}. Unlike that work, here we focus on the \emph{dataset}, which is much larger and covers four languages, and we present a rich set of experiments.

\section{Dataset}
\label{sec:dataset}

\subsection{Data Collection}

\begin{figure}[h]
\centering
\includegraphics[width=0.40\textwidth]{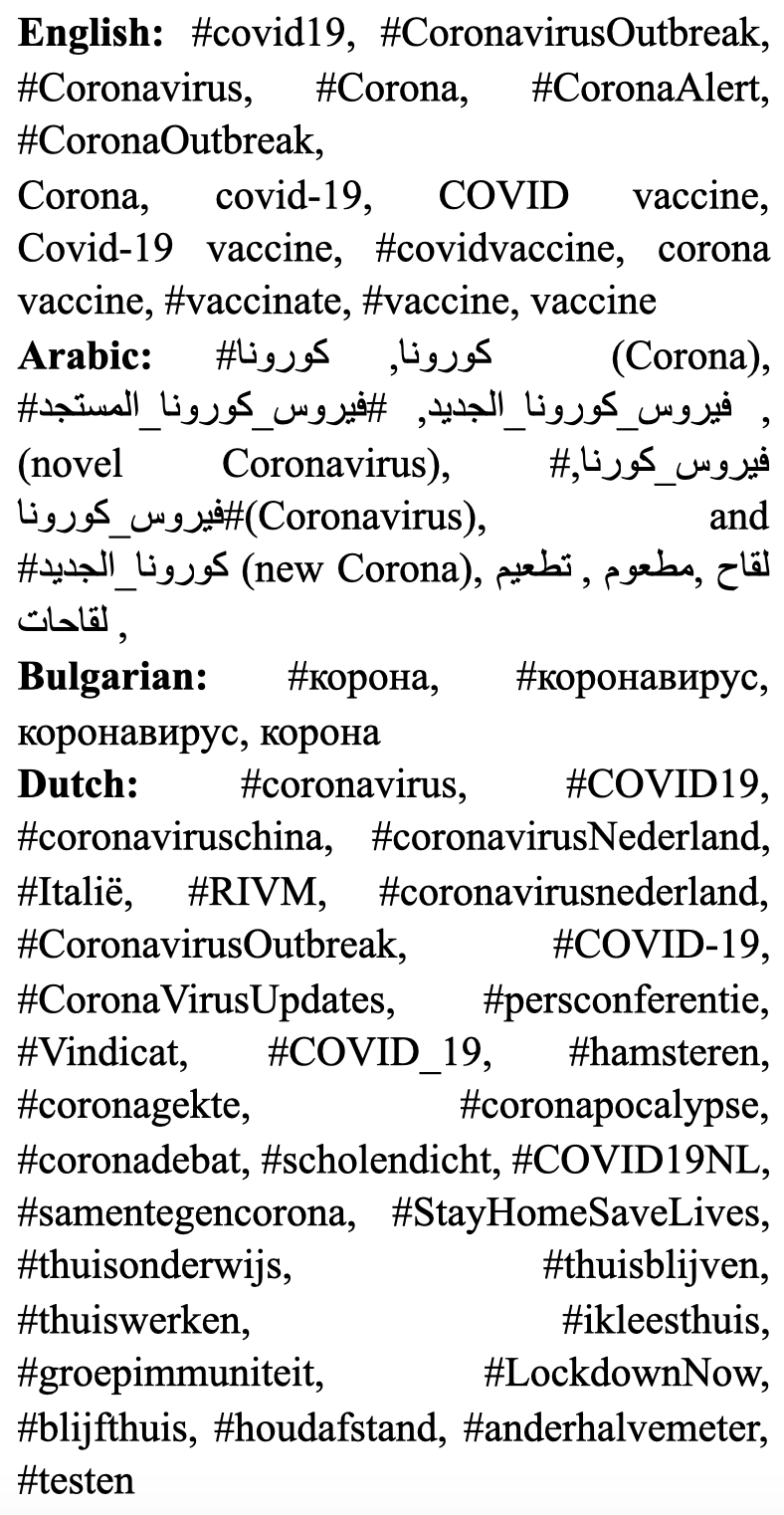}
\caption{The keywords used to collect the tweets.}
\label{fig:tweet_keywords}
\end{figure}

We collected tweets by specifying a target language (English, Arabic, Bulgarian, or Dutch), a set of COVID-19 related keywords, as shown in Figure~\ref{fig:tweet_keywords}, and different time frames: from January 2020 till March 2021.
We collected original tweets (no retweets or replies), we removed duplicates using a similarity-based approach \cite{alam2020standardizing}, and we filtered out tweets with less than five words. Finally, we selected the most frequently liked and retweeted tweets for annotation.

\subsection{Annotation Task}
\label{sec:annotation_instruc_concise}

The annotation task consists of determining whether a tweet contains a factual claim, as well as its veracity, its potential to cause harm (to the society, to a person, to an organization, or to a product), whether it needs verification, and how interesting it is for policy makers. These are then formulated into seven questions presented in Table~\ref{tab:class_label_distribution}. 

The full annotation instructions we gave to the annotators, together with examples, can be found in Appendix~\ref{sec:annotation_instruc_detail}. To facilitate the annotation task, we used the annotation platform described in \citet{alam2020fighting}.  There were 10, 14, 5, and 4 annotators for English, Arabic, Bulgarian, and Dutch, respectively. We used three annotators per tweet, native speakers or fluent in the respective language, male and female, with qualifications ranging from undergrads to PhDs in various disciplines. We resolved the cases of disagreement in a consolidation discussion including external consolidators.

Table \ref{tab:annotated_tweets} shows two tweets, annotated for all questions. The first tweet contains a harmful factual claim with a causal argument of interest to the public and requiring urgent fact-checking. Moreover, it appears to spread rumors. It also attacks government officials, and thus might need the attention of government entities. The second tweet contains a non-harmful factual claim of interest to the general public, which is probably true, but should be fact-checked urgently. It might be of interest to policy makers as it discusses protection from COVID-19.

\begin{table}[htb!]
\centering
\setlength{\tabcolsep}{2.0pt}
\scalebox{0.73}{
\begin{tabular}{@{}llrrrr@{}}
\toprule
\textbf{Exp.} & \multicolumn{1}{c}{\textbf{Class labels}} & \multicolumn{1}{c}{\textbf{En}} & \multicolumn{1}{c}{\textbf{Ar}} & \multicolumn{1}{c}{\textbf{Bg}} & \multicolumn{1}{c}{\textbf{Nl}} \\ \midrule
\multicolumn{2}{l}{\textbf{\begin{tabular}[c]{@{}l@{}}Q1: Does the tweet contain \\ a verifiable factual claim?\end{tabular}}} & \textbf{4,542} & \textbf{4,966} & \textbf{3,697} & \textbf{2,665} \\ \midrule
\multirow{2}{*}{Bin} & No & 1,651 & 1,527 & 1,130 & 1,412 \\
 & Yes & 2,891 & 3,439 & 2,567 & 1,253 \\ \midrule
\multicolumn{2}{l}{\textbf{\begin{tabular}[c]{@{}l@{}}Q2: To what extent does the tweet \\ appear to contain false information?\end{tabular}}} & \textbf{2,891} & \textbf{3,439} & \textbf{2,567} & \textbf{1,253} \\ \midrule
\multirow{5}{*}{Multi} & No, definitely contains no false info & 222 & 137 & 102 & 190 \\
 & No, probably contains no false info & 2,272 & 2,465 & 2,166 & 718 \\
 & not sure & 213 & 22 & 219 & 113 \\
 & Yes, probably contains false info & 142 & 764 & 5 & 162 \\
 & Yes, definitely contains false info & 42 & 51 & 75 & 70 \\ \cmidrule{2-6}
\multirow{2}{*}{Bin} & No & 2,494 & 2,602 & 2,268 & 908 \\
 & Yes & 184 & 815 & 80 & 232 \\ \midrule
\multicolumn{2}{l}{\textbf{\begin{tabular}[c]{@{}l@{}}Q3: Will the tweet's claim have \\ an impact on or be of interest to\\ the general public?\end{tabular}}} & \textbf{2,891} & \textbf{3,439} & \textbf{2,567} & \textbf{1,253} \\ \midrule
\multirow{5}{*}{Multi} & No, definitely not of interest & 11 & 9 & 2 & 108 \\ 
 & No, probably not of interest & 94 & 120 & 68 & 181 \\
 & not sure & 8 & 14 & 0 & 21 \\
 & Yes, probably of interest & 2,481 & 2,047 & 2,000 & 645 \\
 & Yes, definitely of interest & 297 & 1249 & 497 & 298 \\\cmidrule{2-6}
\multirow{2}{*}{Bin} & No & 105 & 129 & 70 & 289 \\
 & Yes & 2,778 & 3,296 & 2,497 & 943 \\ \midrule
\multicolumn{2}{l}{\textbf{\begin{tabular}[c]{@{}l@{}}Q4: To what extent does the tweet \\ appear to be harmful to the society, \\ a person(s),  a company(s) \\ or a product(s)?\end{tabular}}} & \textbf{2,891} & \textbf{3,439} & \textbf{2,567} & \textbf{1,253} \\ \midrule
\multirow{5}{*}{Multi} & No, definitely not harmful & 1,107 & 1,591 & 437 & 520 \\ 
 & No, probably not harmful & 1,126 & 1,088 & 1,876 & 449 \\
 & not sure & 21 & 22 & 17 & 23 \\
 & Yes, probably harmful & 505 & 433 & 196 & 204 \\
 & Yes, definitely harmful & 132 & 305 & 41 & 57 \\ \cmidrule{2-6}
\multirow{2}{*}{Bin} & No & 2,233 & 2,233 & 2,313 & 969 \\
 & Yes & 637 & 637 & 237 & 261 \\ \midrule
\multicolumn{2}{l}{\textbf{\begin{tabular}[c]{@{}l@{}}Q5: Do you think that a professional \\ fact-checker should verify \\ the claim in the tweet?\end{tabular}}} & \textbf{2,891} & \textbf{3,439} & \textbf{2,567} & \textbf{1,247} \\ \midrule
\multirow{4}{*}{Multi} & No, no need to check & 472 & 163 & 721 & 410 \\
 & No, too trivial to check & 1,799 & 1,948 & 1,326 & 330 \\
 & Yes, not urgent & 513 & 1086 & 422 & 309 \\
 & Yes, very urgent & 107 & 242 & 98 & 198 \\ \cmidrule{2-6}
\multirow{2}{*}{Bin} & No & 2,271 & 2,111 & 2,047 & 740 \\
 & Yes & 620 & 1,328 & 520 & 507 \\ \bottomrule
\end{tabular}
}
\caption{Statistics about Q1--Q5.
In rows with a question, the number refers to the total number of tweets for the respective language. Bin: binary, Multi: multiclass.}
\label{tab:class_label_distribution}
\end{table}

\begin{table}[!tbh]
\centering
\setlength{\tabcolsep}{2.0pt}
\scalebox{0.78}{
\begin{tabular}{@{}llrrrr@{}}
\toprule
\textbf{Exp.} & \multicolumn{1}{c}{\textbf{Class labels}} & \multicolumn{1}{c}{\textbf{En}} & \multicolumn{1}{c}{\textbf{Ar}} & \multicolumn{1}{c}{\textbf{Bg}} & \multicolumn{1}{c}{\textbf{Nl}} \\ \midrule
\multicolumn{2}{l}{\textbf{\begin{tabular}[c]{@{}l@{}}Q6: Is the tweet harmful to the \\ society and why?\end{tabular}}} & \textbf{4,542} & \textbf{4,966} & \textbf{3,697} & \textbf{2,665} \\ \midrule
\multirow{8}{*}{Multi} & No, joke or sarcasm & 95 & 155 & 200 & 162 \\
 & No, not harmful & 4,040 & 3,872 & 3,017 & 2,254 \\
 & not sure & 2 & 12 & 4 & 9 \\
 & Yes, bad cure & 4 & 6 & 7 & 10 \\
 & Yes, other & 33 & 23 & 4 & 7 \\
 & Yes, panic & 90 & 347 & 305 & 35 \\
 & Yes, rumor conspiracy & 246 & 425 & 151 & 159 \\
 & \begin{tabular}[c]{@{}l@{}}Yes, xenophobic racist \\ prejudices or hate speech\end{tabular} & 32 & 126 & 9 & 29 \\ \cmidrule{2-6}
\multirow{2}{*}{Bin} & No & 4,135 & 4,027 & 3,217 & 2,416 \\
 & Yes & 405 & 927 & 476 & 240 \\ \midrule
\multicolumn{2}{l}{\textbf{\begin{tabular}[c]{@{}l@{}}Q7: Do you think that this tweet \\ should get the attention of \\a government entity?\end{tabular}}} & \textbf{4,542} & \textbf{4,966} & \textbf{3,697} & \textbf{2,665} \\ \midrule
\multirow{10}{*}{Multi} & No, not interesting & 3,892 & 1,598 & 3,186 & 2,092 \\
 & not sure & 6 & 12 & 0 & 4 \\
 & Yes, asks question & 7 & 129 & 1 & 116 \\
 & Yes, blame authorities & 181 & 93 & 51 & 177 \\
 & Yes, calls for action & 63 & 61 & 8 & 43 \\
 & Yes, classified as in question 6 & 249 & 725 & 333 & 136 \\
 & Yes, contains advice & 18 & 102 & 10 & 50 \\
 & Yes, discusses action taken & 35 & 695 & 25 & 32 \\
 & Yes, discusses cure & 60 & 1,536 & 79 & 8 \\
 & Yes, other & 31 & 15 & 4 & 7 \\ \cmidrule{2-6}
\multirow{2}{*}{Bin} & No & 3,892 & 1,598 & 3,186 & 2092 \\
 & Yes & 644 & 3,356 & 511 & 569 \\ \bottomrule
\end{tabular}
}
\caption{Statistics about Q6--Q7.}
\label{tab:class_label_distribution_q6_7}
\end{table}

\begin{table}[tbh]
\centering
\scalebox{0.91}{
\begin{tabular}{@{}l@{}}
\toprule
\parbox{8.0cm}{\textbf{Tweet 1:} \textit{This is unbelievable. It reportedly took Macron's threat to close the UK border for Boris Johnson to finally shutdown bars and restaurants. The Elysee refers to UK policy as `benign neglect'. This failure of leadership is costing lives.}} \\ \midrule
Q1: Yes \\
Q2: No, probably contains no false info \\
Q3: Yes, probably of interest \\
Q4: Yes, definitely harmful \\
Q5: Yes, very urgent \\
Q6: Yes, rumor, or conspiracy \\
Q7: Yes, blames authorities \\ 
\toprule
\parbox{8.0cm}{\textbf{Tweet 2:} \textit{An antiviral spray against novel \#coronavirus has developed in Shanghai Public Health Clinical Center, which can be put into throat as shield from virus. The spray can greatly help protect front-line medical staff, yet mass-production for public use is not available for now. https://t.co/bmRzCssCY5}} \\ \midrule
Q1: Yes \\
Q2: not sure \\
Q3: Yes, definitely of interest \\
Q4: No, definitely not harmful \\
Q5: Yes, very urgent \\
Q6: No, not harmful \\
Q7: Yes, discusses cure \\ 
\bottomrule
\end{tabular}
}
\caption{Examples of annotated English tweets.}
\label{tab:annotated_tweets}
\end{table}

\subsection{Labels}
\label{ssec:label}

The annotation was designed in a way that the fine-grained multiclass labels can be easily transformed into binary labels by mapping all \textit{\textbf{Yes*}} into \textbf{Yes}, and all \textit{\textbf{No*}} into \textbf{No}, and dropping the \textit{not sure} tweets.

Although some of the questions are correlated (for Q1-Q5, this is on purpose), the annotation instructions are designed, so that the dataset can be used independently for different tasks. Questions Q2-Q4 (see Table~\ref{tab:class_label_distribution}) can be seen as categorical or numerical (i.e.,~on a Likert scale), and thus can be addressed in a classification or in an ordinal regression setup. Below, we will use classification.

\subsection{Statistics}
\label{ssec:stat}

We annotated a total of 4,542, 4,966, 3,697, and 2,665 tweets for English, Arabic, Bulgarian, and Dutch, respectively. Table~\ref{tab:class_label_distribution} shows the distribution of the class labels for all languages. 

The distribution for Q1 is quite balanced: 64\% \emph{Yes} vs. 36\% \emph{No}. Only tweets that contain factual claims were annotated for Q2--Q5.

For question Q2, 81\% of the tweets were judged to contain no false information, for 6\% the judges were unsure, and 13\% were suspected to possibly contains false information. Note that this is not fact-checking, but just a subjective judgment about whether the claim seems credible.

For Q3, which asks whether the tweet is of potential interest to the general public, the distribution is quite skewed towards \emph{Yes}: 94\% of the examples. This can be attributed to the fact that we selected the tweets based on frequency of retweets and likes, and these would be the interesting tweets.

For Q4, which asks whether the tweet is harmful to the society, we can see that the labels vary widely from not harmful to harmful; yet, most are not harmful.

For Q5, which asks whether a professional fact-checker should verify the claim, the majority of the cases were either \emph{Yes, not urgent} (23\%) or \emph{No, no need to check} (17\%). It appears that a professional fact-checker should verify the claim urgently in a relatively small number of cases (6\%). 

For questions Q2-4, the \emph{not sure} cases are very rare. However, they are substantially more prevalent for Q2 (6\%), which is hard to annotate, as in many cases, it requires access to external information. When annotating Q2 (as well as Q3--Q7, but not Q1), the annotators were presented the tweet as it appears in Twitter, which allows them to see some context, e.g.,~the user identifier, a snapshot of linked webpage, a video, an image, etc. 

For Q6, most of the tweets were considered \emph{not harmful} for the society or a \emph{joke}. However, 1\% of the tweets were found to be \emph{xenophobic, racist, prejudices or hate speech}, 6\% to be \emph{rumor conspiracy}, and 5\% to be \emph{spreading panic}. 

For Q7, the vast majority of the tweets were not interesting for policy makers and government entities. However, 3\% blamed the authorities.

\subsection{Inter-Annotation Agreement}
\label{ssec:anno_agr}

We assessed the quality of the annotations by computing inter-annotator agreement. As mentioned earlier, three annotators independently annotated each tweet, following the provided annotation instructions, and the cases of disagreement were resolved in a consolidation discussion including external consolidators. We computed the Fleiss Kappa ($\kappa$) between each annotator and the consolidated label, using (a)~the original multiclass labels, and (b)~binary labels. The results for the English dataset are shown in Table~\ref{tab:annotation_agr}, where we can see that overall, there is moderate to substantial agreement.\footnote{Recall that values of Kappa of 0.21--0.40, 0.41--0.60, 0.61--0.80, and 0.81--1.0 correspond to fair, moderate, substantial and perfect agreement, respectively~\cite{landis1977measurement}.} 
The Kappa value is higher for objective questions such as Q1, and it is lower for subjective and partially subjective questions;\footnote{Our agreement is much higher than for related tasks \cite{Roitero2020}: Krippendorff's $\alpha$ in [0.066; 0.131].} the number of labels is also a factor. The agreement for the other languages is also moderate to substantial for all questions and also both for binary and for multiclass labels; see Appendix~\ref{sec:appendix_anno_agreement} for more detail.

\begin{table}[]
\centering
\scalebox{0.95}{
\setlength{\tabcolsep}{2.5pt}
\begin{tabular}{@{}lrrrrrrr@{}}
\toprule
\multicolumn{1}{c}{\textbf{Agree. Pair}} & \multicolumn{1}{c}{\textbf{Q1}} & \multicolumn{1}{c}{\textbf{Q2}} & \multicolumn{1}{c}{\textbf{Q3}} & \multicolumn{1}{c}{\textbf{Q4}} & \multicolumn{1}{c}{\textbf{Q5}} & \multicolumn{1}{c}{\textbf{Q6}} & \multicolumn{1}{c}{\textbf{Q7}} \\ \midrule
\multicolumn{8}{c}{\textbf{Multiclass}} \\ \midrule
A1 - C & 0.81 & 0.73 & 0.59 & 0.74 & 0.79 & 0.67 & 0.73 \\
A2 - C & 0.67 & 0.53 & 0.44 & 0.45 & 0.39 & 0.65 & 0.47 \\
A3 - C & 0.78 & 0.58 & 0.63 & 0.61 & 0.70 & 0.17 & 0.42 \\
\textbf{Avg} & \bf 0.75 & \bf 0.61 & \bf 0.55 & \bf 0.60 & \bf 0.63 & \bf 0.50 & \bf 0.54 \\\midrule
\multicolumn{8}{c}{\textbf{Binary}} \\\midrule
A1 - C & 0.81 & 0.73 & 0.77 & 0.85 & 0.84 & 0.77 & 0.92 \\
A2 - C & 0.67 & 0.58 & 0.53 & 0.43 & 0.52 & 0.33 & 0.57 \\
A3 - C & 0.78 & 0.70 & 0.63 & 0.70 & 0.74 & 0.11 & 0.57 \\ 
\textbf{Avg} & \bf 0.75 & \bf 0.67 & \bf 0.64 & \bf 0.66 & \bf 0.70 & \bf 0.40 & \bf 0.69 \\ \bottomrule
\end{tabular}
}
\caption{Inter-annotator agreement using Fleiss Kappa ($\kappa$) for the English dataset. \emph{A} refers to annotator, and \emph{C} refers to consolidation.}
\label{tab:annotation_agr}
\end{table}
\begin{table*}[tbh]
\footnotesize
\centering
\setlength{\tabcolsep}{2.0pt}  
\scalebox{0.95}{
\begin{tabular}{cc@{ }@{ }|@{ }@{ }cccc@{ }@{ }|@{ }@{ }cccc@{ }@{ }|@{ }@{ }cccc@{ }@{ }|@{ }@{ }cccc}
\toprule
\multicolumn{1}{c}{\textbf{}} & \multicolumn{1}{c}{\textbf{}} & \multicolumn{4}{c}{\textbf{English}} & \multicolumn{4}{c}{\textbf{Arabic}} & \multicolumn{4}{c}{\textbf{Bulgarian}} & \multicolumn{4}{c}{\textbf{Dutch}} \\
\midrule
\multicolumn{1}{c}{\textbf{Q.}} & \multicolumn{1}{c@{ }@{ }|@{ }@{ }}{\textbf{Cls.}} & \multicolumn{1}{c}{\textbf{Maj.}} & \multicolumn{1}{c}{\textbf{FT}} & \multicolumn{1}{c}{\textbf{BT}} & \multicolumn{1}{c@{ }@{ }|@{ }@{ }}{\textbf{RT}} & \multicolumn{1}{c}{\textbf{Maj.}} & \multicolumn{1}{c}{\textbf{FT}} & \multicolumn{1}{c}{\textbf{ArBT}} & \multicolumn{1}{c@{ }@{ }|@{ }@{ }}{\textbf{XLM-r}} & \multicolumn{1}{c}{\textbf{Maj.}} & \multicolumn{1}{c}{\textbf{FT}} & \multicolumn{1}{c}{\textbf{mBT}} & \multicolumn{1}{c@{ }@{ }|@{ }@{ }}{\textbf{XLM-r}} & \multicolumn{1}{c}{\textbf{Maj.}} & \multicolumn{1}{c}{\textbf{FT}} & \multicolumn{1}{c}{\textbf{BTje}} & \multicolumn{1}{c}{\textbf{XLM-r}} \\ \midrule
\multicolumn{18}{c}{\textbf{Binary (Coarse-grained)}} \\ \midrule
Q1 & 2 & 48.7 & \textbf{77.7} & \textbf{76.5} & \underline{\textbf{78.6}} & 56.8 & \textbf{63.1} & \textbf{83.8} & \underline{\textbf{84.2}} & 58.3 & \textbf{75.5} & \textbf{84.0} & \underline{\textbf{87.6}} & 36.5 & \textbf{61.9} & \textbf{75.4} & \underline{\textbf{80.0}} \\
Q2 & 2 & 91.6 & 89.0 & \textbf{92.1} & \underline{\textbf{92.7}} & 68.3 & \textbf{81.7} & \underline{\textbf{84.0}} & \textbf{83.1} & 95.0 & 85.2 & 94.7 & \underline{\textbf{95.0}} & 64.9 & \underline{\textbf{87.9}} & \textbf{75.1} & \textbf{83.1} \\
Q3 & 2 & 96.3 & 69.3 & \textbf{96.4} & \underline{\textbf{96.9}} & 96.3 & 82.0 & 96.0 & \underline{\textbf{96.3}} & 96.5 & 79.3 & 96.0 & \underline{\textbf{96.5}} & 62.3 & \textbf{69.9} & \textbf{76.9} & \underline{\textbf{78.3}} \\
Q4 & 2 & 66.7 & \underline{\textbf{96.3}} & \textbf{85.6} & \textbf{89.0} & 67.2 & \underline{\textbf{96.2}} & \textbf{90.3} & \textbf{89.0} & 86.8 & \underline{\textbf{96.5}} & \textbf{87.7} & \textbf{88.4} & 63.9 & \textbf{72.7} & \textbf{77.1} & \underline{\textbf{83.9}} \\
Q5 & 2 & 67.7 & \textbf{83.8} & \textbf{80.6} & \underline{\textbf{84.4}} & 46.8 & \underline{\textbf{74.0}} & \textbf{65.9} & \textbf{66.7} & 70.5 & \textbf{81.5} & \textbf{80.5} & \underline{\textbf{82.9}} & 44.4 & \underline{\textbf{75.3}} & \textbf{66.8} & \textbf{70.9} \\
Q6 & 2 & 86.7 & \underline{\textbf{92.1}} & \textbf{88.9} & \textbf{90.5} & 72.5 & \textbf{79.3} & \textbf{88.9} & \underline{\textbf{89.8}} & 83.2 & \underline{\textbf{95.0}} & \textbf{84.5} & \textbf{85.1} & 84.7 & 74.9 & \textbf{86.9} & \underline{\textbf{88.1}} \\
Q7 & 2 & 78.3 & \textbf{80.6} & \textbf{85.5} & \underline{\textbf{86.1}} & 57.7 & \underline{\textbf{81.6}} & \textbf{77.4} & \textbf{77.4} & 80.1 & \underline{\textbf{87.2}} & \textbf{81.6} & \textbf{81.7} & 65.6 & \textbf{74.1} & \textbf{78.3} & \underline{\textbf{79.6}} \\ \cmidrule{3-18}
\textbf{Avg.} &  & 76.6 & \textbf{84.1} & \textbf{86.5} & \underline{\textbf{88.3}} & 66.5 & \textbf{79.7} & \underline{\textbf{83.8}} & \textbf{83.7} & 81.5 & \textbf{85.8} & \textbf{87.0} & \underline{\textbf{88.2}} & 60.3 & \textbf{73.8} & \textbf{76.6} & \underline{\textbf{80.5}} \\ \midrule
\multicolumn{18}{c}{\textbf{Multiclass (Fine-grained)}} \\ \midrule
Q2 & 5 & 67.9 & 44.7 & \textbf{69.2} & \underline{\textbf{70.6}} & 62.9 & 53.3 & \textbf{75.6} & \underline{\textbf{76.2}} & 77.3 & \textbf{78.8} & \textbf{77.8} & \underline{\textbf{79.3}} & 36.5 & \textbf{39.7} & \textbf{45.7} & \underline{\textbf{51.1}} \\
Q3 & 5 & 78.9 & 57.4 & \textbf{82.5} & \underline{\textbf{82.8}} & 44.4 & \underline{\textbf{75.6}} & \textbf{53.7} & \textbf{59.5} & 64.2 & \underline{\textbf{78.2}} & \textbf{68.1} & \textbf{68.8} & 32.0 & \underline{\textbf{77.7}} & \textbf{50.9} & \textbf{53.9} \\
Q4 & 5 & 19.9 & \underline{\textbf{69.2}} & \textbf{56.0} & \textbf{58.0} & 28.1 & \underline{\textbf{54.2}} & \textbf{46.9} & \textbf{50.6} & 58.8 & \underline{\textbf{69.0}} & \textbf{65.6} & \textbf{67.1} & 21.0 & \textbf{42.9} & \textbf{46.3} & \underline{\textbf{53.1}} \\
Q5 & 5 & 46.8 & \underline{\textbf{84.9}} & \textbf{62.0} & \textbf{70.0} & 41.2 & \underline{\textbf{52.6}} & \underline{\textbf{52.6}} & \textbf{52.4} & 36.0 & \underline{\textbf{81.5}} & \textbf{58.0} & \textbf{61.6} & 18.4 & \underline{\textbf{69.6}} & \textbf{40.7} & \textbf{46.4} \\
Q6 & 8 & 84.0 & 71.7 & \textbf{86.5} & \underline{\textbf{87.7}} & 68.7 & \textbf{71.5} & \textbf{82.2} & \underline{\textbf{84.8}} & 76.6 & \underline{\textbf{79.6}} & \textbf{77.2} & \textbf{78.8} & 74.4 & 46.0 & \underline{\textbf{76.7}} & \textbf{76.3} \\
Q7 & 10 & 78.1 & \textbf{82.4} & \textbf{83.4} & \underline{\textbf{85.3}} & 13.8 & \textbf{40.8} & \textbf{57.5} & \underline{\textbf{61.6}} & 80.1 & \textbf{66.8} & \textbf{81.7} & \underline{\textbf{81.8}} & 65.4 & 45.3 & \textbf{72.2} & \underline{\textbf{74.1}} \\ \cmidrule{3-18} 
\textbf{Avg.} & & 62.6 & \textbf{68.4} & \textbf{73.3} & \underline{\textbf{75.8}} & 43.2 & \textbf{58.0} & \textbf{61.4} & \underline{\textbf{64.2}} & 65.5 & \underline{\textbf{75.6}} & \textbf{71.4} & \textbf{72.9} & 41.3 & \textbf{53.5} & \textbf{55.4} & \underline{\textbf{59.1}} \\ \bottomrule
\end{tabular}
}
\caption{\textbf{Monolingual experiments.} We report weighted F$_1$ for binary (top) and multiclass (bottom) experiments for English, Arabic, Bulgarian, and Dutch using various Transformers and FastText (\textbf{FT}). The results that improve over the majority class baseline (\emph{Maj.}) are in \textbf{bold}, and the best system is \underline{underlined}. Legend: Q. -- question, Cls -- number of classes. \textbf{BT}: BERT, \textbf{ArBT}: Monolingual BERT in Arabic (AraBERT), \textbf{RT}: RoBERTa. \textbf{mBT:} multilingual BERT, \textbf{BTje}: Monolingual BERT in Dutch (BERTje), \textbf{XLM-r}: XLM-RoBERTa.}
\label{tab:results_binary_multiclass_tasks}
\end{table*}

\section{Experimental Setup}
\label{sec:experiments}

We experimented with binary and multiclass settings for all languages, using deep contextualized text representations based on large-scale pretrained transformer models such as BERT, mBERT, RoBERTa, XLM-R, etc. We further performed multitask and cross-language learning, and we modeled the social context of the tweet, as well as the propagandistic nature of the language used.

\subsection{Data Preprocessing}
The preprocessing includes removal of hash-symbols and non-alphanumeric symbols, case folding, URL replacement with a URL tag, and username replacement with a user tag. We generated a stratified split \cite{sechidis2011stratification} of the data into 70\%/10\%/20\% for training/development/testing. We used the development set to tune the model hyper-parameters.

\paragraph{Models}

Large-scale pretrained Transformer models have achieved state-of-the-art performance for several NLP tasks. We experimented with several such models to evaluate their efficacy under various training scenarios such as, binary vs. multiclass classification, multilingual setup, etc. 

We used BERT \citep{devlin2018bert} and RoBERTa for English, AraBERT \citep{baly2020arabert} for Arabic, and  BERTje \cite{devries2019bertje} for Dutch.
We further used multilingual transformers such as \citep{liu2019roberta}, multilingual BERT (mBERT) and XLM-r~\citep{conneau2019unsupervised}.
Finally, we used static embeddings from FastText~\citep{joulin2017bag}.

For Transformer models, we used the Transformer toolkit~\cite{Wolf2019HuggingFacesTS}. We fine-tuned each model using the default settings for ten epochs as described in \citep{devlin2018bert}. Due to instability, we performed ten reruns for each experiment using different random seeds, and we picked the model that performed best on the development set. 

For FastText, we used embeddings pretrained on Common Crawl, which were released by FastText for different languages.

\subsection{Multitask Learning}

While question Q1, Q2, $\ldots$, Q7 can be deemed as independent tasks, some questions are interrelated and information in one can help improve the predictive performance for another task. For example, Q5 asks whether the claim in a tweet should be checked by a professional fact-checker. A tweet is more likely to be worth fact-checking if its factuality is under question (Q2), if it is interesting for the general public (Q3), and, more importantly, if it is harmful (Q4). This interdependence between the tasks (which was by design) motivated multitask learning with the goal of improving the performance of the classifier on Q5 using Q2, Q3, and Q4 as auxiliary tasks. We applied multitask learning by aggregating task-specific dense layers of transformers.
More specifically, for the four questions, we computed the cross-entropy loss for each task independently and we then combined them linearly:
$L = \lambda_1 L_1 + \lambda_2 L_2+\lambda_3 L_3+ \lambda_4 L_4$
where the lambdas sum up to 1.

\subsection{Twitter/Propagandistic/Botometer Features}

Previous work has demonstrated the utility of modeling the social context for related tasks such as predicting factuality \cite{canini2011finding,baly2020written}, and thus we extracted context features from the Twitter object. We further modeled the degree of propagandistic content in the tweet, and we also used bot-related features.

The features from the Twitter object include general information about the tweet's content, as well as about its author, i.e.,~whether the account is verified, whether it uses the default profile picture, the number of years since the account's creation, the number of followers, statuses, and friends, whether the tweet contains quotes, media or a URL, and the factuality of the website it points to.\footnote{From \url{http://mediabiasfactcheck.com}}

The propagandistic features include two scores modeling the degree to which the message is propagandistic: one from the Proppy~\cite{AAAI2019:proppy,BARRONCEDENO20191849} and one from the Prta~\cite{da-san-martino-etal-2020-prta} systems, as implemented in Tanbih \cite{zhang-etal-2019-tanbih}.

We extracted bot-related features  using the Botometer~\cite{10.1145/2872518.2889302}.
This includes a score about whether the tweet author is likely to be a bot, as well as content-, network- and friend-related scores. 
These features are summarized in Appendix (Table~\ref{tab:social_features}).

\subsection{Baseline}
\label{ssec:baseline}

For all tasks, we use a majority class baseline. Note that for questions with highly imbalanced class distribution, this baseline could be very high, which can make it hard for models to improve upon (see Table~\ref{tab:results_binary_multiclass_tasks}). For example, in the Arabic dataset for Q3 in the binary setting, the tweets from the \emph{Yes} category comprise 96\% of the total.

\subsection{Evaluation Measures}

We report weighted F$_1$ score, which takes into account class imbalance. In Appendix \ref{sec:appendix_results}, we further report some other evaluation measures such as accuracy and macro-average F$_1$ score.

\section{Evaluation Results}
\label{sec:eval_results}

\subsection{Binary Classification}
\label{ssec:binary_classification}

The evaluation results for binary classification are shown in the first half of Table~\ref{tab:results_binary_multiclass_tasks}.

\paragraph{English}
Most models outperformed the baseline. RoBERTa outperformed the other models in five of the seven tasks, and FastText was best on the remaining two.

\paragraph{Arabic}
In all the cases except for Q3 (which has a very skewed distribution as we mentioned above), all models performed better than the baseline. The strongest models were FastText and XLM-r, each winning 3 of the seven tasks. AraBERT was best on one of the tasks.

\paragraph{Bulgarian}
For Bulgarian, most models outperformed the baselines. We also have a highly imbalanced distribution for Q2 (96.6\% `No') and for Q3 (97.3\% `Yes'), which made for a very hard to beat baseline. XLM-r was best for four out of seven tasks, and FastText was best on the remaining three.

\paragraph{Dutch}
For Dutch, all models managed to outperform the majority class baseline, except for FastText on Q6 (due to class imbalance). XLM-r performed best in five out of the seven tasks, and FastText was best on the other two.

\subsection{Multiclass Classification}
\label{ssec:multiclass_classification}

The bottom part of Table \ref{tab:results_binary_multiclass_tasks} shows the multiclass results. The \emph{Cls} column shows the number of classes per task. We can see that this number ranges in [5,10], and thus the multiclass setup is a much harder compared to binary classification. This explains the much lower results compared to the binary case (including for the baseline). 

\paragraph{English}
Most models outperformed the baseline. The most successful model was RoBERTa, which was best for four out of the six tasks; FastText was best on the remaining two tasks.

\paragraph{Arabic}
Almost all models outperformed the majority class baseline for all tasks (except for FastText on Q2). FastText was best for three of the six tasks, XLM-r was best on two, and AraBERT was best on the remaining one.

\paragraph{Bulgarian}
All models outperformed the baselines for all tasks. FastText was best for four tasks, and XLM-r was best for the remaining two.

\paragraph{Dutch}
Most models outperformed the majority class baseline. 
XLM-r was best for three of the six tasks, FastText was best on two, and BERTje won the remaining one.

\subsection{Discussion}

Overall, the experimental results above have shown that there is no single model that performs universally best across all languages, all tasks, and all class sizes. We should note, however, the strong performance of RoBERTa for English, and of XLM-r for the remaining languages. 

Interestingly, language-specific models, such as AraBERT for Arabic and BERTje for Dutch, were not as strong as multilingual ones such as XLM-r. This could be partially explained by the fact that for them we used a base-sized models, while for XLM-r we used a large model. 

Finally, we should note the strong performance of context-free models such as FastText. We believe that it is suitable for the noisy text of tweets due to its ability to model not only words but also character $n$-grams.
In future work, we plan to try transformers specifically trained on tweets and/or on COVID-19 related data such as BERTweet~\cite{nguyen-etal-2020-bertweet} and COVID-Twitter-BERT \cite{CT_BERT}.

\section{Advanced Experiments}
\label{sec:additional_exp}

Next, we performed some additional, more advanced experiments, including multilingual training, modeling the Twitter context, the use of propagandistic language, and whether the user is likely to be a bot, as well as multitask learning. We describe each of these experiments in more detail below.

\subsection{Multilingual Training}
\label{sec:multiling_exp}

We experimented with a multilingual setup, where we combined the data from all languages. We fine-tuned a multilingual model (mBERT),\footnote{We also tried XLM-r, but it performed worse.} separately for each question. The results are shown in Table~\ref{tab:results_multilingual_tasks}, where the \emph{Mul} columns shows the multilingual fine-tuning results, which are to be compared to the monolingual fine-tuning results in the previous respective columns. We can see that the differences are small and that the results are mixed. Multlingual fine-tuning helps a bit in about half of the cases, but it also hurts a bit in the other half of the cases. This is true both in the binary and in the multiclass setting.

\begin{table}[h]
\centering
\setlength{\tabcolsep}{2.5pt}
\scalebox{0.87}{
\begin{tabular}{ll|rr|rr|rr|rr}
\toprule
\multicolumn{1}{c}{\textbf{}} & \multicolumn{1}{c|}{\textbf{}} & \multicolumn{2}{c|}{\textbf{English}} & \multicolumn{2}{c|}{\textbf{Arabic}} & \multicolumn{2}{c|}{\textbf{Bulgarian}} & \multicolumn{2}{c}{\textbf{Dutch}} \\ \midrule
\multicolumn{1}{c}{\textbf{Q.}} & \multicolumn{1}{c|}{\textbf{Cls.}} & \multicolumn{1}{c}{\textbf{EN}} & \multicolumn{1}{c|}{\textbf{Mul}} & \multicolumn{1}{c}{\textbf{AR}} & \multicolumn{1}{c|}{\textbf{Mul}} & \multicolumn{1}{c}{\textbf{BG}} & \multicolumn{1}{c|}{\textbf{Mul}} & \multicolumn{1}{c}{\textbf{NL}} & \multicolumn{1}{c}{\textbf{Mul}} \\ \midrule
\multicolumn{10}{c}{\textbf{Binary (Coarse-grained)}} \\ \midrule
Q1 & 2 & 76.5 & \textbf{77.5} & \textbf{82.6} & 81.5 & \textbf{84.0} & 81.8 & \textbf{76.6} & \textbf{76.6} \\
Q2 & 2 & 92.1 & \textbf{92.6} & \textbf{81.4} & 78.8 & \textbf{94.7} & 94.4 & \textbf{73.4} & 71.3 \\
Q3 & 2 & \textbf{96.4} & \textbf{96.4} & 96.1 & \textbf{96.5} & 96.0 & \textbf{96.5} & \textbf{78.6} & 77.2 \\
Q4 & 2 & \textbf{85.6} & 83.9 & \textbf{87.7} & 87.2 & \textbf{87.7} & 87.2 & \textbf{75.7} & 74.7 \\
Q5 & 2 & \textbf{80.6} & 78.6 & 63.1 & \textbf{66.5} & 80.5 & \textbf{83.2} & 64.3 & \textbf{68.7} \\
Q6 & 2 & \textbf{88.9} & 85.6 & 84.6 & \textbf{85.6} & 84.5 & \textbf{85.6} & \textbf{87.5} & 85.6 \\
Q7 & 2 & \textbf{85.5} & 79.9 & 73.4 & \textbf{79.9} & \textbf{81.6} & 79.9 & 77.7 & \textbf{79.9} \\ \cmidrule{3-10}
\textbf{Avg.} &  & 86.5 & 84.9 & 81 & 82.4 & 87.5 & 87.8 & 76.2 & 76.2 \\ \midrule
\multicolumn{10}{c}{\textbf{Multiclass (Fine-grained)}} \\ \midrule
Q2 & 5 & 69.2 & \textbf{70.2} & 70.8 & \textbf{72.0} & \textbf{77.8} & \textbf{77.8} & 46.1 & \textbf{47.6} \\
Q3 & 5 & 82.5 & \textbf{82.9} & 55.8 & \textbf{55.9} & 68.1 & \textbf{68.3} & \textbf{49.7} & 47.1 \\
Q4 & 5 & 56.0 & \textbf{56.3} & \textbf{48.2} & 43.8 & 65.6 & \textbf{68.9} & 47.9 & \textbf{48.5} \\
Q5 & 5 & \textbf{62.0} & 61.2 & \textbf{56.0} & 54.6 & \textbf{58.0} & 56.3 & 40.8 & \textbf{42.4} \\
Q6 & 8 & \textbf{86.5} & 84.8 & \textbf{79.0} & 78.9 & 77.2 & \textbf{77.8} & \textbf{78.1} & 75.6 \\
Q7 & 10 & \textbf{83.4} & \textbf{83.4} & \textbf{54.7} & 53.5 & \textbf{81.7} & 80.2 & \textbf{69.2} & 68.3 \\ \cmidrule{3-10}
\textbf{Avg.} &  & 73.3 & 73.1 & 60.7 & 59.8 & 71.4 & 71.5 & 55.3 & 54.9 \\ 
\bottomrule
\end{tabular}
}
\caption{\textbf{Multilingual experiments using mBERT.} Shown are results for monolingual vs. multilingual models (weighted F$_1$). \textbf{Mul} is trained on the combined English, Arabic, Bulgarian, and Dutch data.}
\label{tab:results_multilingual_tasks}
\end{table}

\subsection{Twitter/Propagandistic/Botometer}
\label{sec:social_feat}

We conducted experiments with Twitter, propaganda, and botness features alongside the posteriors from the BERT classifier, which we combined using XGBoost~\cite{chen2016xgboost}. The results are shown in Table~\ref{tab:results_social_features}. We can see that many of the combinations yielded improvements, with botness being the most useful, followed by propaganda, and finally by the Twitter object features.

\begin{table}[h]
\centering
\setlength{\tabcolsep}{2.0pt}
\scalebox{0.85}{
\begin{tabular}{cc|ccccc}
\toprule
\multicolumn{7}{c}{\textbf{Binary (Coarse-grained)}}\\ 
\midrule
\bf Q. & \bf Cls &
\multicolumn{1}{c}{\textbf{BERT}} & \multicolumn{1}{c}{\textbf{B+TF}} & \multicolumn{1}{c}{\textbf{B+Prop}} & \multicolumn{1}{c}{\textbf{B+Bot}} &
\multicolumn{1}{c}{\textbf{B+All}} \\  \midrule
Q1 & 2 & 76.5 & \textbf{76.9} & \textbf{77.1} & \textbf{\underline{77.8}} & \textbf{76.8} \\
Q2 & 2 & 92.1 & 91.8 & \textbf{92.3} & \textbf{92.3} & \textbf{\underline{92.4}} \\
Q3 & 2 & \underline{96.4} & 96.3 & \underline{96.4} & \underline{96.4} & 96.3 \\
Q4 & 2 & 85.6 & \textbf{86.5} & \textbf{86.5} & \textbf{\underline{86.7}} & \textbf{86.4} \\
Q5 & 2 & 80.6 & \textbf{\underline{82.0}} & \textbf{81.5} & \textbf{81.9} & \textbf{81.4} \\
Q6 & 2 & 88.9 & 88.9 & \textbf{89.6} & \textbf{\underline{89.4}} & 87.6 \\
Q7 & 2 & 85.5 & 84.1 & \textbf{85.6} & \textbf{\underline{86.2}} & 83.9 \\
\midrule
\multicolumn{7}{c}{\textbf{Multiclass (Fine-grained)}} \\ \midrule
Q2 & 5 & 69.2 & \textbf{69.4} & \textbf{70.0} & \textbf{\underline{70.3}} & 69.1 \\
Q3 & 5 & 82.5 & 81.2 & 82.2 & 82.2 & 81.6 \\
Q4 & 5 & 56.0 & 52.7 & 55.9 & \textbf{\underline{56.8}} & 53.4 \\
Q5 & 4 & 62.0 & 60.9 & \textbf{\underline{63.2}} & \textbf{62.8} & 58.2 \\
Q6 & 8 & 86.5 & 84.3 & 86.4 & \textbf{\underline{86.6}} & 84.1 \\
Q7 & 10 & 83.4 & 79.6 & \textbf{83.7} & \textbf{\underline{83.9}} & 80.8 \\
\bottomrule
\end{tabular}
}
\caption{\textbf{Experiments with social features and BERT (weighted F$_1$)}. Improvements over BERT (\textbf{B}) are shown in \textbf{bold}, while the highest scores for each question are \underline{underlined}. \textbf{TF}: Tweet features, \textbf{Prop}: propaganda features, \textbf{Bot}: Botometer features.}
\label{tab:results_social_features}
\end{table}

\subsection{Multitask Learning}
\label{sec:multitask_learn}

For the multitask learning experiments, we used BERT and RoBERTa on the English dataset, in a multiclass setting, fine-tuned with a multiclass objective on Q2--Q5. The results are shown in Table~\ref{tab:multitask_results}. We achieved sizable improvements for Q2, Q4, and Q5 over the single-task setup. However, performance degraded for Q3, probably due to the skewed label distribution for this question.

\begin{table}[h]
\centering
\scalebox{0.85}{
\setlength{\tabcolsep}{2.0pt}
\begin{tabular}{ccccc}
\toprule
\multicolumn{5}{c}{\textbf{English, multiclass}} \\ \midrule
 & \multicolumn{1}{c}{\textbf{BERT(S)}} & \multicolumn{1}{c}{\textbf{BERT(M)}} & \multicolumn{1}{c}{\textbf{RoBERTa(S)}} & \multicolumn{1}{c}{\textbf{RoBERTa(M)}} \\ \midrule
Q2 & 69.2 & \textbf{72.9} & 70.62 & \textbf{73.85} \\
Q3 & \textbf{82.5} & 71.6 & \textbf{82.84} & 67.34 \\
Q4 & 56.0 & \textbf{67.9} & 58.04 & \textbf{66.95} \\
Q5 & 62.0 & \textbf{76.8} & 70.02 & \textbf{75.75} \\ \midrule
\end{tabular}
}
\caption{\textbf{Multitask learning experiments} (weighted F$_1$). \textbf{S}: Single task, \textbf{M}: Multitask.} 
\label{tab:multitask_results}
\end{table}

\section{Conclusion and Future Work}
\label{sec:conclutions}

We presented a large manually annotated dataset of COVID-19 tweets, aiming to help in the fight against the COVID-19 infodemic. The dataset combines the perspectives and the interests of journalists, fact-checkers, social media platforms, policymakers, and society as a whole. It includes tweets in Arabic, Bulgarian, Dutch, and English, and we are making it freely available to the research community. We further reported a number of evaluation results for all languages using various transformer architectures. Moreover, we performed advanced experiments, including multilingual training, modeling the Twitter context, the use of propagandistic language, and whether the user is likely to be a bot, as well as multitask learning.

In future work, we plan to explore multimodality and explainability \cite{RANLP2021:propaganda:interpretable}. We further want to model the task as a multitask ordinal regression \cite{baly-etal-2019-multi}, as Q2--Q5 are defined on an ordinal scale. Moreover, we would like to put the data and the system in some practical use; in fact, we have already used them to analyze disinformation about COVID-19 in Bulgaria \cite{RANLP2021:COVID19:Bulgarian} and Qatar \cite{RANLP2021:COVID19:Qatar}. Finally, the data will be used in a shared task at the CLEF-2022 CheckThat! lab; part of it was used for the NLP4IF-2021 shared task \cite{shaar-etal-2021-findings}.

\section*{Acknowledgments}

We thank Akter Fatema, Al-Awthan Ahmed, Al-Dobashi Hussein, El Messelmani Jana, Fayoumi Sereen, Mohamed Esraa, Ragab Saleh, and Shurafa Chereen for helping with the Arabic annotations.

We also want to thank the Atlantic Club in Bulgaria and DataBee for their support for the Bulgarian annotations.

This research is part of the Tanbih mega-project,
developed at the Qatar Computing Research Institute, HBKU, which aims to limit the impact of ``fake news,'' propaganda, and media bias by making users aware of what they are reading.

This material is also based upon work supported by the US National Science Foundation under Grants No. 1704113 and No. 1828199.

This publication was also partially made possible by the innovation grant No. 21 -- Misinformation and Social Networks Analysis in Qatar from Hamad Bin Khalifa University’s (HBKU) Innovation Center. The findings achieved herein are solely the responsibility of the authors.
\section*{Ethics Statement}

\subsection*{Dataset Collection}

We collected the dataset using the Twitter API\footnote{\url{http://developer.twitter.com/en/docs}} with keywords that only use terms related to COVID-19, without other biases. We followed the terms of use outlined by Twitter.\footnote{\url{http://developer.twitter.com/en/developer-terms/agreement-and-policy}} Specifically, we only downloaded public tweets, and we only distribute dehydrated Twitter IDs. 

\subsection*{Biases}

We note that some of the annotations are subjective, and we have clearly indicated in the text which these are. Thus, it is inevitable that there would be biases in our dataset. Yet, we have a very clear annotation schema and instructions, which should reduce biases.

\subsection*{Misuse Potential}
Most datasets compiled from social media present some risk of misuse. We, therefore, ask researchers to be aware that our dataset can be maliciously used to unfairly moderate text (e.g.,~a tweet) that may not be malicious based on biases that may or may not be related to demographics and other information within the text. Intervention with human moderation would be required in order to ensure this does not occur. 

\subsection*{Intended Use}
Our dataset can enable automatic systems for analysis of social media content, which could be of interest to practitioners, professional fact-checker, journalists, social media platforms, and policymakers. Such systems can be used to alleviate the burden for social media moderators, but human supervision would be required for more intricate cases and in order to ensure that the system does not cause harm.

Our models can help fight the infodemic, and they could support analysis and decision making for the public good. However, the models can also be misused by malicious actors. Therefore, we ask the potential users to be aware of potential misuse. With the possible ramifications of a highly subjective dataset, we distribute it for research purposes only, without a license for commercial use. Any biases found in the dataset are unintentional, and we do not intend to do harm to any group or individual.

\bibliography{bib/main,bib/checkthat19}
\bibliographystyle{acl_natbib}

\newpage
\clearpage
\section*{Appendix}
\label{sec:appendix}
\appendix

\section{Experimental Setup}
\label{sec:appendix_exp_param}

\subsection{Transformer Parameters}
Below, we list the values of the hyper-parameters that we used for fine-tuning the Transformer models we used. We further release all our scripts, together with the data.

\begin{itemize}
    \item Batch size: 32;
    \item Learning rate (Adam): 2e-5;
    \item Number of epochs: 10;
    \item Max seq length: 128.
\end{itemize}

\textbf{Models and Number of Parameters:}
\begin{itemize}
    \item \textbf{BERT} (bert-base-uncased): L=12, H=768, A=12, total parameters: 110M; where \textit{L} is the number of layers (i.e.,~Transformer blocks), \textit{H} is the hidden size, and \textit{A} is the number of self-attention heads;
    \item \textbf{RoBERTa} (roberta-base): similar to BERT-base, but with a higher number of parameters (125M);
    \item \textbf{AraBERT} (bert-base-arabert): same number as BERT (110M);
    \item \textbf{BERTje} (bert-base-dutch-cased): same number as BERT (110M);
    \item \textbf{RoBERTa for Bulgarian} (roberta-base-bulgarian): L=12, H=768, A=12, parameters=125M;
    \item \textbf{BERT Multilingual} (bert-base-multilingual-uncased) (mBERT): similar to BERT-base with a higher number of parameters (172M);
    \item \textbf{XLM-RoBERTa} (xlm-roberta-base): L=12, H=768, A=12; the total number of parameters is 270M.
\end{itemize}

\newpage
\subsection{FastText Parameters}
We release all the FastText parameters with our released packages. We have not listed them here due to the length of the resulting list.   

\subsection{XGBoost Parameters}
We used XGBoost to run experiments with Twitter, Propaganda, Botometer, and BERT model predictions. We release the scripts with our code repository, which contains detailed the parameter settings. 

\subsection{Computing Infrastructure and Runtime}
We used a server with NVIDIA Tesla V100-SXM2-32 GB GPU, 56 cores, and 256GB CPU memory. To perform an experiment for a question, on average the computing time took 40 minutes using the BERT base model. This means about four hours for all seven questions using one Transformer architecture.

\clearpage
\newpage

\section{Twitter/Propagandistic/Botometer Features Types}
\label{sec:appendix_social_feat}

In the additional experiments in Section~\ref{sec:additional_exp}, we extracted features from the Twitter object, botness scores from the Botometer API, and propaganda scores from the Tanbih API. We have already described the experiments with these features in Section~\ref{sec:additional_exp}, but we did not have enough space in the main text of the paper to describe the features themselves. Table~\ref{tab:social_features} aims to address this. It lists the features, offers a brief description for each one, and specifies its type, which can be one of the following:

\begin{itemize}
    \item \emph{Boolean features} take a value of either 0 or 1; we use them directly.
    \item \emph{Categorical features} take a fixed number of possible values, and we encode them using one-hot representation.
    \item \emph{Numerical features} are continuous and may take an infinite number of values; we transform the value $x$ of such a feature according to the formula $x^{\prime} = ln(x+1)$. 
\end{itemize}

\begin{table}[tbh]
\centering
\scalebox{0.73}{
\begin{tabular}{@{}lcl@{}}
\toprule
\multicolumn{1}{c}{\textbf{Tweet-Specific}} &  & \multicolumn{1}{c}{\textbf{Description}} \\ \midrule
URL & B & \begin{tabular}[c]{@{}l@{}} Is there aa URL \\is included in the tweet?\end{tabular} \\
Reply & B & Is the tweet a reply? \\
Quotes & B & Is this a quoted tweet? \\
URL & B & Does the tweet contain a URL? \\
Media & B & Does the tweet contain media? \\
Source & C & \begin{tabular}[c]{@{}l@{}}Tools/devices used to post the tweet,\\ as an HTML-formatted string. 
\end{tabular} \\
Domain & C & Domain of the included URL. \\
Num media & N & Number of media mentioned in the tweet. \\
Media type & C & Type of included media, e.g., image.\\
Fact & C & \begin{tabular}[c]{@{}l@{}}A label (unknown, high, mixed, or low) \\ for factuality of the linked information, \\e.g., if it is a news medium.
\end{tabular} \\
\midrule
\multicolumn{1}{c}{\textbf{User-Specific}} & & \multicolumn{1}{c}{\textbf{Description}} \\\midrule
Statuses & N & Number of tweets (incl. retweets) posted. \\
Followers & N & \begin{tabular}[c]{@{}l@{}}The number of followers. 
\end{tabular} \\
Friends & N & \begin{tabular}[c]{@{}l@{}}The number of following.
\end{tabular} \\
Favorites & & \begin{tabular}[c]{@{}l@{}}The number of liked tweets. 
\end{tabular} \\
Listed & N & The number of subscriptions to public lists.
\\
Default profile & B & \begin{tabular}[c]{@{}l@{}}Has the user altered the theme\\or the background of the profile?\end{tabular}. \\
Profile img & B
& \begin{tabular}[c]{@{}l@{}}Has the user uploaded\\ a profile image?
\end{tabular} \\
Verified & B & Is it a verified account?\\
Protected & B & \begin{tabular}[c]{@{}l@{}}Has the user chosen \\to protect their tweets?\end{tabular} \\
GEO-enabled & B & Is geotagging enabled?\\ 
\midrule
\multicolumn{1}{c}{\textbf{Botometer}} & & \multicolumn{1}{c}{\textbf{Description}} \\
\midrule
Content & N & \begin{tabular}[c]{@{}l@{}} Score of the length of tweets \\ and frequency of part-of-speech tags.\end{tabular} \\ 
Network & N & \begin{tabular}[c]{@{}l@{}} Score about retweets, mentions, \\ and hashtags that a user tweeted in the past.\end{tabular} \\ 
Temporal & N & \begin{tabular}[c]{@{}l@{}} Score about time patterns of tweets.\end{tabular} \\ 
Sentiment & N & \begin{tabular}[c]{@{}l@{}} Score about the sentiment of the user.\end{tabular} \\ 
Friend & N & \begin{tabular}[c]{@{}l@{}} Score about users that liked or \\ retweeted tweets by the user.\end{tabular} \\ 
Language & C & \begin{tabular}[c]{@{}l@{}} Language used.\end{tabular} \\ 
User & N & \begin{tabular}[c]{@{}l@{}} Score about the number of followers' user\\ name, and consistency of shared language \\between the tweets.\end{tabular} \\ 
\midrule
\multicolumn{1}{c}{\textbf{Propaganda}} & & \multicolumn{1}{c}{\textbf{Description}} \\
\midrule
Prta & N
& \begin{tabular}[c]{@{}l@{}} Sentence-level Prta propaganda score\end{tabular} \\
Proppy & N
& \begin{tabular}[c]{@{}l@{}} Article-level Proppy propaganda score\end{tabular} \\
\bottomrule
\end{tabular}
}
\caption{Features modeling social context, botness, and propaganda. The middle column shows the type of feature: B is Boolean, C is categorical, and N is numerical.}
\label{tab:social_features}
\end{table}

\clearpage
\newpage
\section{Detailed Result by Language}
\label{sec:appendix_results}

The main text of the paper had only weighted F1 scores; here we report also accuracy (Acc) and macro-F1 (M-F1) for English, Arabic, Bulgarian and Dutch are shown in Tables \ref{tab:results_english_binary_multi_labels}, \ref{tab:results_arabic_binary_multi_labels}, \ref{tab:results_bulgarian_binary_multi_labels} and \ref{tab:results_dutch_binary_multi_labels}, respectively.

\begin{table}[tbh]
\centering
\scalebox{0.90}{
\begin{tabular}{@{}lrrr|rrr@{}}
\toprule
\multicolumn{1}{c}{\textbf{}} & \multicolumn{3}{c}{\textbf{Binary}} & \multicolumn{3}{|c}{\textbf{Multiclass}} \\ \midrule
\multicolumn{1}{c}{\textbf{Q}} & \multicolumn{1}{c}{\textbf{Acc}} & \multicolumn{1}{c}{\textbf{M-F1}} & \multicolumn{1}{c|}{\textbf{W-F1}} & \multicolumn{1}{c}{\textbf{Acc}} & \multicolumn{1}{c}{\textbf{M-F1}} & \multicolumn{1}{c}{\textbf{W-F1}} \\ \midrule
\multicolumn{7}{c}{\textbf{Majority}} \\ \midrule
Q1 & 63.0 & 38.7 & 48.7 &  &  &  \\
Q2 & 94.3 & 48.5 & 91.6 & 77.7 & 17.5 & 67.9 \\
Q3 & 97.6 & 49.4 & 96.3 & 85.5 & 18.4 & 78.9 \\
Q4 & 76.8 & 43.4 & 66.7 & 36.9 & 10.8 & 19.9 \\
Q5 & 77.5 & 43.7 & 67.7 & 61.5 & 19.0 & 46.8 \\
Q6 & 91.0 & 47.6 & 86.7 & 89.1 & 11.8 & 84.0 \\
Q7 & 85.1 & 46.0 & 78.3 & 85.0 & 9.2 & 78.1 \\ \midrule
\multicolumn{7}{c}{\textbf{FastText}} \\ \midrule
Q1 & 79.3 & 65.9 & 77.7 &  &  &  \\
Q2 & 90.8 & 61.3 & 89.0 & 46.3 & 28.9 & 44.7 \\
Q3 & 69.5 & 66.9 & 69.3 & 65.0 & 33.5 & 57.4 \\
Q4 & 97.0 & 54.5 & 96.3 & 78.0 & 20.7 & 69.2 \\
Q5 & 85.4 & 65.2 & 83.8 & 89.4 & 14.3 & 84.9 \\
Q6 & 93.0 & 58.8 & 92.1 & 72.9 & 68.7 & 71.7 \\
Q7 & 81.9 & 71.3 & 80.6 & 86.8 & 23.6 & 82.4 \\ \midrule
\multicolumn{7}{c}{\textbf{BERT}} \\ \midrule
Q1 & 76.8 & 74.5 & 76.5 &  &  &  \\
Q2 & 92.8 & 60.1 & 92.1 & 73.0 & 25.5 & 69.2 \\
Q3 & 97.2 & 54.8 & 96.4 & 85.2 & 27.0 & 82.5 \\
Q4 & 85.9 & 79.3 & 85.6 & 56.4 & 40.3 & 56.0 \\
Q5 & 81.5 & 71.0 & 80.6 & 64.8 & 37.2 & 62.0 \\
Q6 & 90.2 & 62.3 & 88.9 & 88.3 & 22.2 & 86.5 \\
Q7 & 87.0 & 68.5 & 85.5 & 85.2 & 27.7 & 83.4 \\ \midrule
\multicolumn{7}{c}{\textbf{RoBERTa}} \\ \midrule
Q1 & 78.8 & 76.8 & 78.6 &  &  &  \\
Q2 & 93.2 & 63.6 & 92.7 & 71.1 & 37.9 & 70.6 \\
Q3 & 97.6 & 60.5 & 96.9 & 83.3 & 33.2 & 82.8 \\
Q4 & 89.1 & 84.5 & 89.0 & 58.7 & 43.9 & 58.0 \\
Q5 & 84.7 & 77.4 & 84.4 & 71.4 & 51.3 & 70.0 \\
Q6 & 91.4 & 68.6 & 90.5 & 88.7 & 26.2 & 87.7 \\
Q7 & 86.7 & 71.3 & 86.1 & 86.1 & 33.7 & 85.3 \\ \bottomrule
\end{tabular}
}
\caption{Classification results on the \textbf{test set for English} using various models including a majority class baseline for different questions. Acc. is Accuracy, M-F1 is macro F1, and W-F1 is weighted average F1. }
\label{tab:results_english_binary_multi_labels}
\end{table}

\begin{table}[tbh]
\centering
\scalebox{0.90}{
\begin{tabular}{@{}lrrr|rrr@{}}
\toprule
\multicolumn{1}{c}{\textbf{}} & \multicolumn{3}{c}{\textbf{Binary}} & \multicolumn{3}{|c}{\textbf{Multiclass}} \\ \midrule
\multicolumn{1}{c}{\textbf{Q}} & \multicolumn{1}{c}{\textbf{Acc}} & \multicolumn{1}{c}{\textbf{M-F1}} & \multicolumn{1}{c|}{\textbf{W-F1}} & \multicolumn{1}{c}{\textbf{Acc}} & \multicolumn{1}{c}{\textbf{M-F1}} & \multicolumn{1}{c}{\textbf{W-F1}} \\ \midrule
\multicolumn{7}{c}{\textbf{Majority}} \\ \midrule
Q1 & 69.4 & 41.0 & 56.8 &  &  &  \\
Q2 & 78.0 & 43.8 & 68.3 & 74.0 & 17.0 & 62.9 \\
Q3 & 97.5 & 49.4 & 96.3 & 59.5 & 14.9 & 44.4 \\
Q4 & 77.1 & 43.5 & 67.2 & 45.2 & 12.4 & 28.1 \\
Q5 & 61.5 & 38.1 & 46.8 & 56.9 & 18.1 & 41.2 \\
Q6 & 81.0 & 44.7 & 72.5 & 78.2 & 11.0 & 68.7 \\
Q7 & 70.1 & 41.2 & 57.7 & 29.9 & 4.6 & 13.8 \\ \midrule
\multicolumn{7}{c}{\textbf{FastText}} \\ \midrule
Q1 & 64.4 & 60.0 & 63.1 &  &  &  \\
Q2 & 84.5 & 66.6 & 81.7 & 57.0 & 30.4 & 53.3 \\
Q3 & 82.6 & 78.2 & 82.0 & 81.1 & 22.7 & 75.6 \\
Q4 & 97.2 & 49.3 & 96.2 & 56.6 & 21.0 & 54.2 \\
Q5 & 75.9 & 67.2 & 74.0 & 54.7 & 27.7 & 52.6 \\
Q6 & 81.5 & 67.2 & 79.3 & 75.3 & 26.6 & 71.5 \\
Q7 & 83.7 & 71.5 & 81.6 & 43.7 & 28.0 & 40.8 \\ \midrule
\multicolumn{7}{c}{\textbf{AraBERT}} \\ \midrule
Q1 & 84.1 & 80.7 & 83.8 &  &  &  \\
Q2 & 84.7 & 75.7 & 84.0 & 78.1 & 30.6 & 75.6 \\
Q3 & 96.5 & 53.0 & 96.0 & 54.4 & 22.9 & 53.7 \\
Q4 & 90.4 & 86.3 & 90.3 & 47.6 & 34.0 & 46.9 \\
Q5 & 66.3 & 63.7 & 65.9 & 53.3 & 34.7 & 52.6 \\
Q6 & 89.2 & 81.4 & 88.9 & 82.8 & 32.3 & 82.2 \\
Q7 & 77.8 & 72.6 & 77.4 & 57.8 & 37.3 & 57.5 \\ \midrule
\multicolumn{7}{c}{\textbf{XLM-RoBERTa}} \\ \midrule
Q1 & 84.6 & 81.0 & 84.2 &  &  &  \\
Q2 & 84.0 & 74.4 & 83.1 & 78.7 & 31.4 & 76.2 \\
Q3 & 97.5 & 49.4 & 96.3 & 60.6 & 23.7 & 59.5 \\
Q4 & 89.1 & 84.3 & 89.0 & 52.1 & 36.5 & 50.6 \\
Q5 & 67.1 & 64.5 & 66.7 & 55.3 & 31.7 & 52.4 \\
Q6 & 89.8 & 83.3 & 89.8 & 85.1 & 36.4 & 84.8 \\
Q7 & 77.8 & 72.6 & 77.4 & 61.7 & 40.8 & 61.6 \\ \bottomrule
\end{tabular}%
}
\caption{Classification results on the \textbf{test set for Arabic} using various models including a majority class baseline for different questions. Acc. is Accuracy, M-F1 is macro F1, and W-F1 is weighted average F1. }
\label{tab:results_arabic_binary_multi_labels}
\end{table}

\begin{table}[tbh]
\centering
\scalebox{0.90}{
\begin{tabular}{@{}lrrr|rrr@{}}
\toprule
\multicolumn{1}{c}{\textbf{}} & \multicolumn{3}{c}{\textbf{Binary}} & \multicolumn{3}{|c}{\textbf{Multiclass}} \\ \midrule
\multicolumn{1}{c}{\textbf{Q}} & \multicolumn{1}{c}{\textbf{Acc}} & \multicolumn{1}{c}{\textbf{M-F1}} & \multicolumn{1}{c|}{\textbf{W-F1}} & \multicolumn{1}{c}{\textbf{Acc}} & \multicolumn{1}{c}{\textbf{M-F1}} & \multicolumn{1}{c}{\textbf{W-F1}} \\ \midrule
\multicolumn{7}{c}{\textbf{Majority}} \\ \midrule
Q1 & 70.5 & 41.4 & 58.3 &  &  &  \\
Q2 & 96.6 & 49.1 & 95.0 & 84.4 & 18.3 & 77.3 \\
Q3 & 97.7 & 49.4 & 96.5 & 75.0 & 21.4 & 64.2 \\
Q4 & 91.1 & 47.7 & 86.8 & 70.9 & 16.6 & 58.8 \\
Q5 & 79.6 & 44.3 & 70.5 & 52.4 & 17.2 & 36.0 \\
Q6 & 88.6 & 47.0 & 83.2 & 84.0 & 11.4 & 76.6 \\
Q7 & 86.4 & 46.4 & 80.1 & 86.4 & 10.3 & 80.1 \\ \midrule
\multicolumn{7}{c}{\textbf{FastText}} \\ \midrule
Q1 & 78.8 & 58.1 & 75.5 &  &  &  \\
Q2 & 88.7 & 55.9 & 85.2 & 84.0 & 16.6 & 78.8 \\
Q3 & 80.6 & 74.0 & 79.3 & 78.5 & 73.5 & 78.2 \\
Q4 & 97.7 & 49.4 & 96.5 & 76.1 & 26.5 & 69.0 \\
Q5 & 86.4 & 51.7 & 81.5 & 86.4 & 13.6 & 81.5 \\
Q6 & 96.6 & 49.1 & 95.0 & 85.2 & 27.7 & 79.6 \\
Q7 & 91.1 & 49.7 & 87.2 & 73.6 & 24.2 & 66.8 \\ \midrule
\multicolumn{7}{c}{\textbf{mBERT}} \\ \midrule
Q1 & 84.5 & 80.3 & 84.0 &  &  &  \\
Q2 & 96.0 & 49.0 & 94.7 & 80.9 & 27.5 & 77.8 \\
Q3 & 96.5 & 49.1 & 96.0 & 71.1 & 27.8 & 68.1 \\
Q4 & 87.8 & 62.0 & 87.7 & 68.2 & 26.8 & 65.6 \\
Q5 & 81.7 & 68.2 & 80.5 & 59.5 & 41.9 & 58.0 \\
Q6 & 86.1 & 58.1 & 84.5 & 80.3 & 16.2 & 77.2 \\
Q7 & 82.9 & 58.1 & 81.6 & 84.4 & 17.8 & 81.7 \\ \midrule
\multicolumn{7}{c}{\textbf{XLM-RoBERTa}} \\ \midrule
Q1 & 88.0 & 84.7 & 87.6 &  &  &  \\
Q2 & 96.6 & 49.1 & 95.0 & 83.6 & 28.6 & 79.3 \\
Q3 & 97.7 & 49.4 & 96.5 & 71.3 & 28.6 & 68.8 \\
Q4 & 88.8 & 63.2 & 88.4 & 67.4 & 32.2 & 67.1 \\
Q5 & 83.6 & 72.7 & 82.9 & 63.0 & 44.7 & 61.6 \\
Q6 & 86.0 & 61.1 & 85.1 & 79.2 & 23.2 & 78.8 \\
Q7 & 82.1 & 60.5 & 81.7 & 84.4 & 18.5 & 81.8 \\ \bottomrule
\end{tabular}%
}
\caption{Classification results on the \textbf{test set for Bulgarian} using various models including a majority class baseline for different questions. Acc. is Accuracy, M-F1 is macro F1, and W-F1 is weighted average F1.}
\label{tab:results_bulgarian_binary_multi_labels}
\end{table}

\begin{table}[tbh]
\centering
\scalebox{0.90}{
\begin{tabular}{@{}lrrr|rrr@{}}
\toprule
\multicolumn{1}{c}{\textbf{}} & \multicolumn{3}{c}{\textbf{Binary}} & \multicolumn{3}{|c}{\textbf{Multiclass}} \\ \midrule
\multicolumn{1}{c}{\textbf{Q}} & \multicolumn{1}{c}{\textbf{Acc}} & \multicolumn{1}{c}{\textbf{M-F1}} & \multicolumn{1}{c|}{\textbf{W-F1}} & \multicolumn{1}{c}{\textbf{Acc}} & \multicolumn{1}{c}{\textbf{M-F1}} & \multicolumn{1}{c}{\textbf{W-F1}} \\ \midrule
\multicolumn{7}{c}{\textbf{Majority}} \\ \midrule
Q1 & 52.8 & 34.6 & 36.5 &  &  &  \\ 
Q2 & 75.4 & 43.0 & 64.9 & 52.8 & 13.8 & 36.5 \\
Q3 & 73.5 & 42.4 & 62.3 & 48.8 & 13.1 & 32.0 \\
Q4 & 74.7 & 42.8 & 63.9 & 38.1 & 11.0 & 21.0 \\
Q5 & 59.5 & 37.3 & 44.4 & 35.3 & 10.4 & 18.4 \\
Q6 & 89.6 & 47.3 & 84.7 & 82.4 & 11.3 & 74.4 \\
Q7 & 76.0 & 43.2 & 65.6 & 75.8 & 8.6 & 65.4 \\ \midrule
\multicolumn{7}{c}{\textbf{FastText}} \\ \midrule
Q1 & 63.1 & 59.7 & 61.9 &  &  &  \\
Q2 & 89.8 & 62.6 & 87.9 & 40.1 & 29.7 & 39.7 \\
Q3 & 69.9 & 69.8 & 69.9 & 81.8 & 26.6 & 77.7 \\
Q4 & 75.9 & 61.9 & 72.7 & 47.2 & 27.1 & 42.9 \\
Q5 & 76.2 & 65.1 & 75.3 & 74.5 & 15.6 & 69.6 \\
Q6 & 77.6 & 63.1 & 74.9 & 52.0 & 28.2 & 46.0 \\
Q7 & 77.6 & 62.2 & 74.1 & 47.2 & 29.4 & 45.3 \\ \midrule
\multicolumn{7}{c}{\textbf{BERTje}} \\ \midrule
Q1 & 75.5 & 75.3 & 75.4 &  &  &  \\
Q2 & 76.3 & 64.9 & 75.1 & 51.6 & 27.8 & 45.7 \\
Q3 & 78.7 & 68.5 & 76.9 & 53.2 & 36.7 & 50.9 \\
Q4 & 78.8 & 67.8 & 77.1 & 48.0 & 29.7 & 46.3 \\
Q5 & 67.1 & 65.3 & 66.8 & 40.9 & 30.3 & 40.7 \\
Q6 & 88.9 & 59.7 & 86.9 & 80.5 & 16.7 & 76.7 \\
Q7 & 78.8 & 69.5 & 78.3 & 75.5 & 19.1 & 72.2 \\ \midrule
\multicolumn{7}{c}{\textbf{XLM-RoBERTa}} \\ \midrule
Q1 & 80.0 & 80.0 & 80.0 &  &  &  \\
Q2 & 84.2 & 75.9 & 83.1 & 56.7 & 31.2 & 51.1 \\
Q3 & 79.1 & 71.1 & 78.3 & 56.3 & 38.4 & 53.9 \\
Q4 & 84.1 & 78.4 & 83.9 & 54.4 & 36.1 & 53.1 \\
Q5 & 71.0 & 69.7 & 70.9 & 46.8 & 35.0 & 46.4 \\
Q6 & 89.1 & 65.5 & 88.1 & 80.7 & 15.7 & 76.3 \\
Q7 & 79.4 & 72.3 & 79.6 & 77.0 & 21.3 & 74.1 \\ \bottomrule
\end{tabular}
}
\caption{Classification results on the \textbf{test set for Dutch} using various models including a majority class baseline for different questions. Acc. is Accuracy, M-F1 is macro F1, and W-F1 is weighted average F1. }
\label{tab:results_dutch_binary_multi_labels}
\end{table}

\clearpage
\newpage
\section{Detailed Annotation Instructions}
\label{sec:annotation_instruc_detail}

\paragraph{General Instructions:}
\begin{enumerate}
    \item For each tweet, the annotator needs to read the text, including the hashtags, and also to look at the tweet itself when necessary by going to the link (i.e., for Q2-7 it might be required to open the tweet link). The reason for not going to the tweet link for Q1 is that we wanted to reduce the complexity of the annotation task and to focus on the content of the tweet only. As for Q2, it might be important to check whether the tweet was posted by an authoritative source, and thus it might be useful for the annotator to open the tweet to get more context. After all, this is how real users perceive the tweet. Since the annotators would open the tweet's link for Q2, they can use that information for the rest of the questions as well (even though this is not required). 
    
    \item The annotators should assume the time when the tweet was posted as a reference when making judgments, e.g., \textit{``Trump thinks, that for the vast majority of Americans, the risk is very, very low.''} would be true when he made the statement but false by the time annotations were carried out for this tweet.
    
    \item The annotators may look at the images, the videos and the Web pages that the tweet links to, as well as at the tweets in the same thread when making a judgment, if needed.
    \item The annotators are not asked to complete questions Q2-Q5 if the answer to question Q1 is \lbl{NO}.

\end{enumerate}

\subsection{Verifiable Factual Claim} 
\uline{\textbf{Question 1:} Does the tweet contain a verifiable factual claim?}

A \emph{verifiable factual claim} is a sentence claiming that something is true, and this can be verified using factual verifiable information such as statistics, specific examples, or personal testimony. Factual claims include the following:\footnote{Inspired by \citep{DBLP:journals/corr/abs-1809-08193}.}
\begin{itemize}
    \item Stating a definition; 
    \item Mentioning quantity in the present or the past;
    \item Making a verifiable prediction about the future; 
    \item Statistics or specific examples;
    \item Personal experience or statement (e.g., \textit{``I spent much of the last decade working to develop an \#Ebola treatment.''})
    \item Reference to laws, procedures, and rules of operation;
    \item References (e.g., URL) to images or videos (e.g., \textit{``This is a video showing a hospital in Spain.''});
    \item Statements that can be technically classified as questions, but in fact contain a verifiable claim based on the criteria above (e.g., \textit{``Hold on - \#China Communist Party now denying \#CoronavirusOutbreak originated in China? This after Beijing's catastrophic mishandling of the virus has caused a global health crisis?''})
    \item Statements about correlation or causation. Such a correlation or causation needs to be explicit, i.e., sentences like \textit{``This is why the beaches haven't closed in Florida. https://t.co/8x2tcQeg21''} is not a claim because it does not explicitly say why, and thus it is not verifiable. 
\end{itemize}

Tweets containing personal opinions and preferences are not factual claims. 
Note that if a tweet is composed of multiple sentences or clauses, at least one full sentence or clause needs to be a claim in order for the tweet to contain a factual claim. If a claim exists in a sub-sentence or a sub-clause, then the tweet is not considered to contain a factual claim. 
For example, \textit{``My new favorite thing is Italian mayors and regional presidents LOSING IT at people violating quarantine''} is not a claim -- it is in fact an opinion. However, if we consider \textit{``Italian mayors and regional presidents LOSING IT at people violating quarantine''} it would be a claim. In addition, when answering this question, annotators should not open the tweet URL. 
Since this is a binary decision task, the answer of this question consists of two labels as defined below. 

\noindent\textbf{Labels:}
\begin{itemize}
    \item \lbl{YES:} if it contains a verifiable factual claim;
    \item \lbl{NO:} if it does not contain a verifiable factual claim;
    \item \lbl{Don't know or can't judge:} the content of the tweet does not provide enough information to make a judgment. It is recommended to categorize the tweet using this label when the content of the tweet is not understandable at all. For example, it uses a language (i.e., non-English) or references difficult to understand;
\end{itemize}

\noindent \textbf{Examples:}

\begin{enumerate}
    \item \textit{Please don't take hydroxychloroquine (Plaquenil) plus Azithromycin for \#COVID19 UNLESS your doctor prescribes it. Both drugs affect the QT interval of your heart and can lead to arrhythmias and sudden death, especially if you are taking other meds or have a heart condition.} \\
    \textbf{Label:} \lbl{YES} \\
    \textbf{Explanation:} There is a claim in the text.
    \item \textit{Saw this on Facebook today and it’s a must read for all those idiots clearing the shelves \#coronavirus \#toiletpapercrisis \#auspol} \\
    \textbf{Label:} \lbl{NO} \\
    \textbf{Explanation:} There is no claim in the text.
\end{enumerate}

\subsection{False Information}
\uline{\textbf{Question 2:} To what extent does the tweet appear to contain false information?} 

The stated claim may contain false information. This question labels the tweets with the categories mentioned below. \textit{False Information} appears on social media platforms, blogs, and news-articles to deliberately misinform or deceive readers.

\noindent\textbf{Labels:}
The labels for this question are defined on a five point Likert scale~\citep{albaum1997likert}. A higher value means that it is more likely to be false: 

\begin{enumerate}
\item \lbl{NO, definitely contains no false information}
\item \lbl{NO, probably contains no false information}
\item \lbl{Not sure}
\item \lbl{YES, probably contains false information}
\item \lbl{YES, definitely contains false information}
\end{enumerate}

To answer this question, it is recommended to open the link of the tweet and to look for additional information to determine the veracity of the claims it makes. For example, if the tweet contains a link to an article from a reputable information source (e.g., Reuters, Associated Press, France Press, Aljazeera English, BBC), then the answer could be ``\dots~contains no false info''. 
Note that answering this question is not required if the answer to Question 1 is \lbl{NO}.

\noindent \textbf{Examples:}
\begin{enumerate}
\item \textit{``Dominican Republic found the cure for Covid-19 https://t.co/1CfA162Lq3''}
\\
\textbf{Label:}\enspace\lbl{5.}\,\lbl{YES, definitely contains false information}
\\
\textbf{Explanation:} This is not correct information at the time of this tweet is posted. 

\item \textit{This is Dr. Usama Riaz. He spent past weeks screening and treating patients with Corona Virus in Pakistan. He knew there was no PPE. He persisted anyways. Today he lost his own battle with coronavirus but he gave life and hope to so many more. KNOW HIS NAME \img{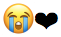} https://t.co/flSwhLCPmx}
\\
\textbf{Label:}\enspace\lbl{2.}\,\lbl{NO, probably contains no false info}
\\
\textbf{Explanation:} The content of the tweet states correct information. 
\end{enumerate}

\subsection{Interest to the General Public}
\uline{\textbf{Question 3:} Will the tweet's claim have an impact on or be of interest to the general public?}

Most often, people do not make interesting claims, which can be verified by our general knowledge. For example, though \textit{``The sky is blue''} is a claim, it is not interesting to the general public. In general, topics such as healthcare, political news, and current events are of higher interest to the general public. Using the five point Likert scale the labels are defined below.

\noindent\textbf{Labels:} The labels are on a 5-point Likert scale:

\begin{enumerate}
\item \lbl{NO, definitely not of interest}
\item \lbl{NO, probably not of interest}
\item \lbl{Not sure}
\item \lbl{YES, probably of interest}
\item \lbl{YES, definitely of interest}
\end{enumerate}

\noindent \textbf{Examples:} 
\begin{enumerate}
    \item \textit{Germany is conducting 160k Covid-19 tests a week. It has a total 35k ventilators, 10k ordered to be made by the govt. It has converted a new 1k bed hospital in Berlin.   It’s death rate is tiny bcos it’s mass testing allows quarantine and bcos it has fewer non reported cases.} \\
    \textbf{Label:}\enspace\lbl{4.}\,\lbl{YES: probably of interest} \\
    \textbf{Explanation:} This information is relevant and of high interest for the general population as it reports how a country deals with COVID-19.
    
    \item \textit{Fake news peddler Dhruv Rathee had said:  ``Corona virus won't spread outside China, we need not worry''  Has this guy ever spoke something sensible? https://t.co/siBAwIR8Pn}
    \\
    \textbf{Label:}\enspace\lbl{2.}\,\lbl{NO, probably not of interest}
    \\
    \textbf{Explanation:} The information is not interesting for the general public as it is an opinion and discusses the statement by someone else.
\end{enumerate}

\subsection{Harmfulness}
\uline{\textbf{Question 4:}  To what extent does the tweet appear to be harmful to society, person(s), company(s) or product(s)?}
The purpose of this question is to determine whether the content of the tweet aims to and can negatively affect society as a whole, specific person(s), company(s), product(s), or spread rumors about them. The content intends to harm or \textit{weaponize the information}\footnote{The use of information as a weapon to spread misinformation and mislead people.}~\citep{broniatowski2018weaponized}.
A rumor involves a form of a statement whose veracity is not quickly verifiable or ever confirmed.\footnote{\url{https://en.wikipedia.org/wiki/Rumor}}

\textbf{Labels:} To categorize the tweets in terms of their harmfulness, we defined the following labels, again using a Likert scale, where a higher value indicates a higher degree of harm:

\begin{enumerate}
\itemsep0em 
\item \lbl{NO, definitely not harmful}
\item \lbl{NO, probably not harmful}
\item \lbl{Not sure}
\item \lbl{YES, probably harmful}
\item \lbl{YES, definitely harmful}
\end{enumerate}

\noindent \textbf{Examples: }

\begin{enumerate}
\item \textit{How convenient but not the least bit surprising from Democrats! As usual they put politics over American citizens. @SpeakerPelosi withheld \#coronavirus bill so DCCC could run ads AGAINST GOP candidates! \#tcot} 
\\
\textbf{Label:}\enspace\lbl{5.}\,\lbl{YES, definitely harmful}
\\
\textbf{Explanation:} This tweet is weaponized to target Nancy Pelosi and the Democrats in general.

\item \textit{As we saw over the wkend, disinfo is being spread online about a supposed national lockdown and grounding flights. Be skeptical of rumors. Make sure you’re getting info from legitimate sources. The @WhiteHouse is holding daily briefings and @cdcgov is providing the latest.}
\\
\textbf{Label:}\enspace\lbl{1.}\,\lbl{NO, definitely not harmful}
\\
\textbf{Explanation:} This tweet is informative and gives advice. It does not attack anyone and is not harmful.

\end{enumerate}

\subsection{Need for Verification}
\uline{\textbf{Question 5:} Do you think that a professional fact-checker should verify the claim in the tweet?}

It is important that a verifiable factual check-worthy claim be verified by a professional fact-checker, as the claim may cause harm to society, specific person(s), company(s), product(s), or some government entities. However, not all factual claims are important or worth fact-checking by a professional fact-checker, as this very time-consuming. Therefore, the purpose is to categorize the tweet using the labels defined below. While doing so, the annotator can rely on the answers to the previous questions. For this question, we defined the following labels to categorize the tweets. This question is to be answered, taking the responses to the previous questions into account.

\noindent \textbf{Labels:}
\begin{enumerate}
\itemsep0em 
\item \lbl{NO, no need to check}: the tweet does not need to be fact-checked, e.g., because it is not interesting, a joke, or does not contain any claim.
\item \lbl{NO, too trivial to check}: the tweet is worth fact-checking, however, this does not require a professional fact-checker, i.e., a non-expert might be able to fact-check the claim. For example, one can verify the information using reliable sources such as the official website of the WHO, etc. An example of a claim is as follows: \textit{``The GDP of the USA grew by 50\% last year.''}
\item \lbl{YES, not urgent}: the tweet should be fact-checked by a professional fact-checker, however, this is not urgent or critical;
\item \lbl{YES, very urgent}: the tweet can cause immediate harm to a large number of people; therefore, it should be verified as soon as possible by a professional fact-checker; 
\item \lbl{Not sure}: the content of the tweet does not have enough information to make a judgment.  It is recommended to categorize the tweet using this label when the content of the tweet is not understandable at all. For example, it uses a language (i.e., non-English) or references that it is difficult to understand.
\end{enumerate}

\noindent\textbf{Examples:} 
\begin{enumerate}
\item \textit{Things the GOP has done during the Covid-19 outbreak:   - Illegally traded stocks  - Called it a hoax - Blamed it on China  - Tried to bailout big business without conditions  What they haven’t done:   - Help workers  - Help small businesses - Produced enough tests or ventilators}
\\
\textbf{Label:}\enspace\lbl{2.}\,\lbl{YES, very urgent}
\\
\textbf{Explanation:} The tweet blames the authorities, and thus, it is important to verify it quickly by a professional fact-checker. In addition, the attention of government entities might be required in order to take necessary actions.
\item \textit{ALERT \img{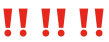} The corona virus can be spread through internationally printed albums. If you have any albums at home, put on some gloves, put all the albums in a box and put it outside the front door tonight. I'm collecting all the boxes tonight for safety. Think of your health.}
\\
\textbf{Label:}\enspace\lbl{5.}\,\lbl{NO, no need to check}
\\
\textbf{Explanation:} This is clearly a joke, and thus is does not require to be checked by a professional fact-checker. 
\end{enumerate}

\subsection{Harmful to Society}
\uline{\textbf{Question 6:} Is the tweet harmful to society and why?}

This question asks whether the content of the tweet is intended to harm or is weaponized to mislead the society. To identify that, we defined the following labels for the categorization. 

\noindent\textbf{Labels:}
\begin{enumerate}[label=\Alph*.]
\itemsep0em 
\item \lbl{NO, not harmful:} the content of the tweet would not harm the society (e.g., \textit{``I like corona beer''}).
\item \lbl{NO, joke or sarcasm:} the tweet contains a joke (e.g., \textit{``If Corona enters Spain, it’ll enter from the side of Barcelona defense''}) or sarcasm (e.g., \textit{```The corona virus is a real thing.' -- Wow, I had no idea!''}).
\item \lbl{Not sure:} if the content of the tweet is not understandable enough to judge.
\item \lbl{YES, panic:} the tweet spreads panic. The content of the tweet can cause sudden fear and anxiety for a large part of the society (e.g., \textit{``there are 50,000 cases ov COVID-19 in Qatar''}).
\item \lbl{YES, xenophobic, racist, prejudices, or hate-speech:} the tweet spreads xenophobia, racism, or prejudices. According to the dictionary\footnote{\url{https://www.dictionary.com/}} \textit{Xenophobic} refers to fear or hatred of foreigners, people from different cultures, or strangers. \textit{Racism} is the belief that groups of humans possess different behavioral traits corresponding to physical appearance and can be divided based on the superiority of one race over another.\footnote{\url{https://en.wikipedia.org/wiki/Racism}} It may also refer to prejudice, discrimination, or antagonism directed against other people because they are of a different race or ethnicity. \textit{Prejudice} is an unjustified or incorrect attitude (i.e.,~typically negative) towards an individual based solely on the individual's membership in a social group.\footnote{\url{http://www.simplypsychology.org/prejudice.html}}
Here is an example: \textit{``do not buy cucumbers from Iran''}.
\item \lbl{YES, bad cure:} the tweet promotes questionable cure, medicine, vaccine, or prevention procedures (e.g., \textit{``\dots drinking bleach can help cure coronavirus''}). 

\item \lbl{YES, rumor, or conspiracy:} the tweet spreads rumors. Rumor is defined as a ``specific (or topical) proposition for belief passed along from person to person usually by word of mouth without secure standards of evidence being present''~\citep{allport1947psychology}.  For example, \textit{``BREAKING: Trump could still own stock in a company that, according to the CDC, will play a major role in providing coronavirus test kits to the federal government, which means that Trump could profit from coronavirus testing. \#COVID-19 \#coronavirus https://t.co/Kwl3ylMZRk''}
\item \lbl{YES, other:} if the content of the tweet does not belong to any of the above categories, then this category can be chosen to label the tweet.
\end{enumerate}

\subsection{Requires Attention}
\uline{\textbf{Question 7:} Do you think that this tweet should get the attention of policy makers of government entities?}

Most often people tweet by blaming authorities, providing advice, and/or call for action. Policy makers might want to respond or to react to this. The purpose of this question is to categorize such information. It is important to note that not all information requires attention from a government entity. Therefore, even if the tweet's content belongs to any of the positive categories, it is important to understand whether it requires attention. For the annotation, it is mandatory to first decide whether attention is necessary  (i.e., \lbl{YES/NO}). If the answer is \lbl{YES}, it is obligatory to select a category from the \lbl{YES} sub-categories below.

\noindent\textbf{Labels:}
\begin{enumerate}[label=\Alph*.]
\itemsep0em 
\item \lbl{NO, not interesting:} the content of the tweet is not important or interesting for any government entity to pay attention to. 
\item \lbl{Not sure:} if the content of the tweet is not understandable enough to judge;
\item \lbl{YES, categorized as in question 6:} some government entity needs to pay attention to this tweet as it is harmful for society and it was labeled as any of the \textit{YES} sub-categories in question 6;
\item \lbl{YES, other:} if the tweet cannot be labeled as any of the above categories, then this label should be selected;
\item \lbl{YES, blames authorities:} the tweet blames  authorities, e.g., \textit{``Dear @VP Pence: Is the below true? Do you have a plan? Also, when are local jurisdictions going to get the \#Coronavirus test kits you promised?''};

\item \lbl{YES, contains advice:} the tweet contains advice about social, political, national, or international issues that requires attention from some government entity (e.g.,~\textit{The elderly \& people with pre-existing health conditions are more susceptible to \#COVID19. To stay safe, they should: \checkmark Keep distance from people who are sick \checkmark Frequently wash hands with soap \& water \checkmark  Protect their mental health});

\item \lbl{YES, calls for action:} the tweet states that some government entity should take action for a particular issue (e.g., \textit{I think the Government should close all the Barber Shops and Salons, let people buy shaving machines and other beauty gardgets keep in their houses. Salons and Barbershops might prove to be another Virus spreading channels @citizentvkenya @SenMutula @CSMutahi\_Kagwe});

\item \lbl{YES, discusses action taken:} the tweet discusses actions taken by governments, companies, individuals for any particular issue, for example, closure of bars, conferences, churches due to the corona virus (e.g., \textit{Due to the current circumstances with the Corona virus, The 4th Mediterranean Heat Treatment and Surface Engineering Conference in Istanbul postponed to 26-28 Mayıs 2021.}). 

\item \lbl{YES, discusses cure:} attention is needed by some government entity as the tweet discusses a possible cure, vaccine, or treatment for a disease;

\item \lbl{YES, asks question:} the tweet asks a question about a particular issue and it requires attention from government entities (e.g., \textit{Special thanks to all doctors and nurses, new found respect for you’ll. Is the virus going to totally disappear in the summer? I live in USA and praying that when the temperature warms up the virus will go away...is my thinking accurate?})

\end{enumerate}


\clearpage
\newpage
\section{Annotation Agreement}
\label{sec:appendix_anno_agreement}

In Tables \ref{tab:annotation_agr_english}, \ref{tab:annotation_agr_arabic}, \ref{tab:annotation_agr_bulgarian} and \ref{tab:annotation_agr_dutch}, we report the inter-annotator agreement for English,\footnote{We have already presented the table for English in the main text of the paper, but we repeat it here to facilitate cross-language comparisons.} Arabic, Bulgarian and Dutch, respectively. Overall, we can see that there is moderate to substantial agreement for all questions both for the binary and for the multi-label setting.

\begin{table}[tbh]
\centering
\scalebox{0.95}{
\setlength{\tabcolsep}{2.5pt}
\begin{tabular}{@{}lrrrrrrr@{}}
\toprule
\multicolumn{1}{c}{\textbf{Agree. Pair}} & \multicolumn{1}{c}{\textbf{Q1}} & \multicolumn{1}{c}{\textbf{Q2}} & \multicolumn{1}{c}{\textbf{Q3}} & \multicolumn{1}{c}{\textbf{Q4}} & \multicolumn{1}{c}{\textbf{Q5}} & \multicolumn{1}{c}{\textbf{Q6}} & \multicolumn{1}{c}{\textbf{Q7}} \\ \midrule
\multicolumn{8}{c}{\textbf{Multiclass}} \\ \midrule
A1 - C & 0.81 & 0.73 & 0.59 & 0.74 & 0.79 & 0.67 & 0.73 \\
A2 - C & 0.67 & 0.53 & 0.44 & 0.45 & 0.39 & 0.65 & 0.47 \\
A3 - C & 0.78 & 0.58 & 0.63 & 0.61 & 0.70 & 0.17 & 0.42 \\
\textbf{Avg} & \bf 0.75 & \bf 0.61 & \bf 0.55 & \bf 0.60 & \bf 0.63 & \bf 0.50 & \bf 0.54 \\\midrule
\multicolumn{8}{c}{\textbf{Binary}} \\\midrule
A1 - C & 0.81 & 0.73 & 0.77 & 0.85 & 0.84 & 0.77 & 0.92 \\
A2 - C & 0.67 & 0.58 & 0.53 & 0.43 & 0.52 & 0.33 & 0.57 \\
A3 - C & 0.78 & 0.70 & 0.63 & 0.70 & 0.74 & 0.11 & 0.57 \\ 
\textbf{Avg} & \bf 0.75 & \bf 0.67 & \bf 0.64 & \bf 0.66 & \bf 0.70 & \bf 0.40 & \bf 0.69 \\ \bottomrule
\end{tabular}
}
\caption{Inter-annotator agreement using Fleiss Kappa ($\kappa$) for the \textbf{English} dataset. \emph{A} refers to annotator, and \emph{C} refers to consolidation.}
\label{tab:annotation_agr_english}
\end{table}

\begin{table}[tbh]
\centering
\scalebox{0.95}{
\setlength{\tabcolsep}{2.5pt}
\begin{tabular}{@{}lrrrrrrr@{}}
\toprule
\multicolumn{1}{c}{\textbf{Agree. Pair}} & \multicolumn{1}{c}{\textbf{Q1}} & \multicolumn{1}{c}{\textbf{Q2}} & \multicolumn{1}{c}{\textbf{Q3}} & \multicolumn{1}{c}{\textbf{Q4}} & \multicolumn{1}{c}{\textbf{Q5}} & \multicolumn{1}{c}{\textbf{Q6}} & \multicolumn{1}{c}{\textbf{Q7}} \\ \midrule
\multicolumn{8}{c}{\textbf{Multiclass}} \\ \midrule
A1 - C & 0.58 & 0.5 & 0.52 & 0.53 & 0.4 & 0.61 & 0.47 \\
A2 - C & 0.59 & 0.52 & 0.52 & 0.55 & 0.44 & 0.62 & 0.4 \\
A3 - C & 0.57 & 0.44 & 0.48 & 0.37 & 0.36 & 0.4 & 0.3 \\
\textbf{Avg} & \textbf{0.58} & \textbf{0.49} & \textbf{0.51} & \textbf{0.48} & \textbf{0.4} & \textbf{0.54} & \textbf{0.39} \\ \midrule
\multicolumn{8}{c}{\textbf{Binary}} \\ \midrule
A1 - C & 0.58 & 0.52 & 0.53 & 0.58 & 0.47 & 0.65 & 0.45 \\
A2 - C & 0.59 & 0.57 & 0.57 & 0.59 & 0.47 & 0.67 & 0.36 \\
A3 - C & 0.57 & 0.48 & 0.53 & 0.47 & 0.39 & 0.46 & 0.29 \\
\textbf{Avg} & \textbf{0.58} & \textbf{0.52} & \textbf{0.54} & \textbf{0.55} & \textbf{0.44} & \textbf{0.59} & \textbf{0.37} \\ \bottomrule
\end{tabular}
}
\caption{Inter-annotator agreement using Fleiss Kappa ($\kappa$) for the \textbf{Arabic} dataset. \emph{A} refers to annotator, and \emph{C} refers to consolidation.}
\label{tab:annotation_agr_arabic}
\end{table}

\begin{table}[tbh]
\centering
\scalebox{0.95}{
\setlength{\tabcolsep}{2.5pt}
\begin{tabular}{@{}lrrrrrrr@{}}
\toprule
\multicolumn{1}{c}{\textbf{Agree. Pair}} & \multicolumn{1}{c}{\textbf{Q1}} & \multicolumn{1}{c}{\textbf{Q2}} & \multicolumn{1}{c}{\textbf{Q3}} & \multicolumn{1}{c}{\textbf{Q4}} & \multicolumn{1}{c}{\textbf{Q5}} & \multicolumn{1}{c}{\textbf{Q6}} & \multicolumn{1}{c}{\textbf{Q7}} \\ \midrule
\multicolumn{8}{c}{\textbf{Multiclass}} \\ \midrule
A1 - C & 0.77 & 0.44 & 0.64 & 0.53 & 0.49 & 0.53 & 0.51 \\
A2 - C & 0.51 & 0.40 & 0.59 & 0.49 & 0.44 & 0.56 & 0.53 \\
A3 - C & 0.47 & 0.38 & 0.57 & 0.49 & 0.38 & 0.53 & 0.40 \\
\textbf{Avg} & \textbf{0.58} & \textbf{0.41} & \textbf{0.60} & \textbf{0.50} & \textbf{0.44} & \textbf{0.54} & \textbf{0.48} \\ \midrule
\multicolumn{8}{c}{\textbf{Binary}} \\ \midrule
A1 - C & 0.77 & 0.41 & 0.71 & 0.56 & 0.61 & 0.47 & 0.50 \\
A2 - C & 0.51 & 0.39 & 0.64 & 0.52 & 0.57 & 0.51 & 0.53 \\
A3 - C & 0.47 & 0.34 & 0.62 & 0.52 & 0.54 & 0.47 & 0.38 \\
\textbf{Avg} & \textbf{0.58} & \textbf{0.38} & \textbf{0.66} & \textbf{0.53} & \textbf{0.57} & \textbf{0.48} & \textbf{0.47} \\\bottomrule
\end{tabular}
}
\caption{Inter-annotator agreement using Fleiss Kappa ($\kappa$) for the \textbf{Bulgarian} dataset. \emph{A} refers to annotator, and \emph{C} refers to consolidation.}
\label{tab:annotation_agr_bulgarian}
\end{table}

\begin{table}[tbh]
\centering
\scalebox{0.95}{
\setlength{\tabcolsep}{2.5pt}
\begin{tabular}{@{}lrrrrrrr@{}}
\toprule
\multicolumn{1}{c}{\textbf{Agree. Pair}} & \multicolumn{1}{c}{\textbf{Q1}} & \multicolumn{1}{c}{\textbf{Q2}} & \multicolumn{1}{c}{\textbf{Q3}} & \multicolumn{1}{c}{\textbf{Q4}} & \multicolumn{1}{c}{\textbf{Q5}} & \multicolumn{1}{c}{\textbf{Q6}} & \multicolumn{1}{c}{\textbf{Q7}} \\ \midrule
\multicolumn{8}{c}{\textbf{Multiclass}} \\  \midrule
A1 - C & 0.63 & 0.54 & 0.58 & 0.58 & 0.54 & 0.66 & 0.63 \\
A2 - C & 0.83 & 0.69 & 0.68 & 0.70 & 0.65 & 0.59 & 0.62 \\
A3 - C & 0.76 & 0.64 & 0.59 & 0.59 & 0.62 & 0.51 & 0.59 \\
\textbf{Avg} & \textbf{0.74} & \textbf{0.62} & \textbf{0.62} & \textbf{0.62} & \textbf{0.60} & \textbf{0.59} & \textbf{0.61} \\ \midrule
\multicolumn{8}{c}{\textbf{Binary}} \\ \midrule
A1 - C & 0.63 & 0.62 & 0.63 & 0.60 & 0.60 & 0.68 & 0.69 \\
A2 - C & 0.83 & 0.73 & 0.76 & 0.77 & 0.75 & 0.63 & 0.69 \\
A3 - C & 0.76 & 0.68 & 0.68 & 0.66 & 0.69 & 0.53 & 0.65 \\
\textbf{Avg} & \textbf{0.74} & \textbf{0.67} & \textbf{0.69} & \textbf{0.68} & \textbf{0.68} & \textbf{0.61} & \textbf{0.68} \\ \bottomrule
\end{tabular}
}
\caption{Inter-annotator agreement using Fleiss Kappa ($\kappa$) for the \textbf{Dutch} dataset. \emph{A} refers to annotator, and \emph{C} refers to consolidation.}
\label{tab:annotation_agr_dutch}
\end{table}

\clearpage
\newpage
\section{Class Label Distribution}
\label{sec:appendix_label_dist}

\begin{figure*}[tbh]
\centering
        \begin{subfigure}[b]{0.75\textwidth}
        \includegraphics[width=\textwidth]{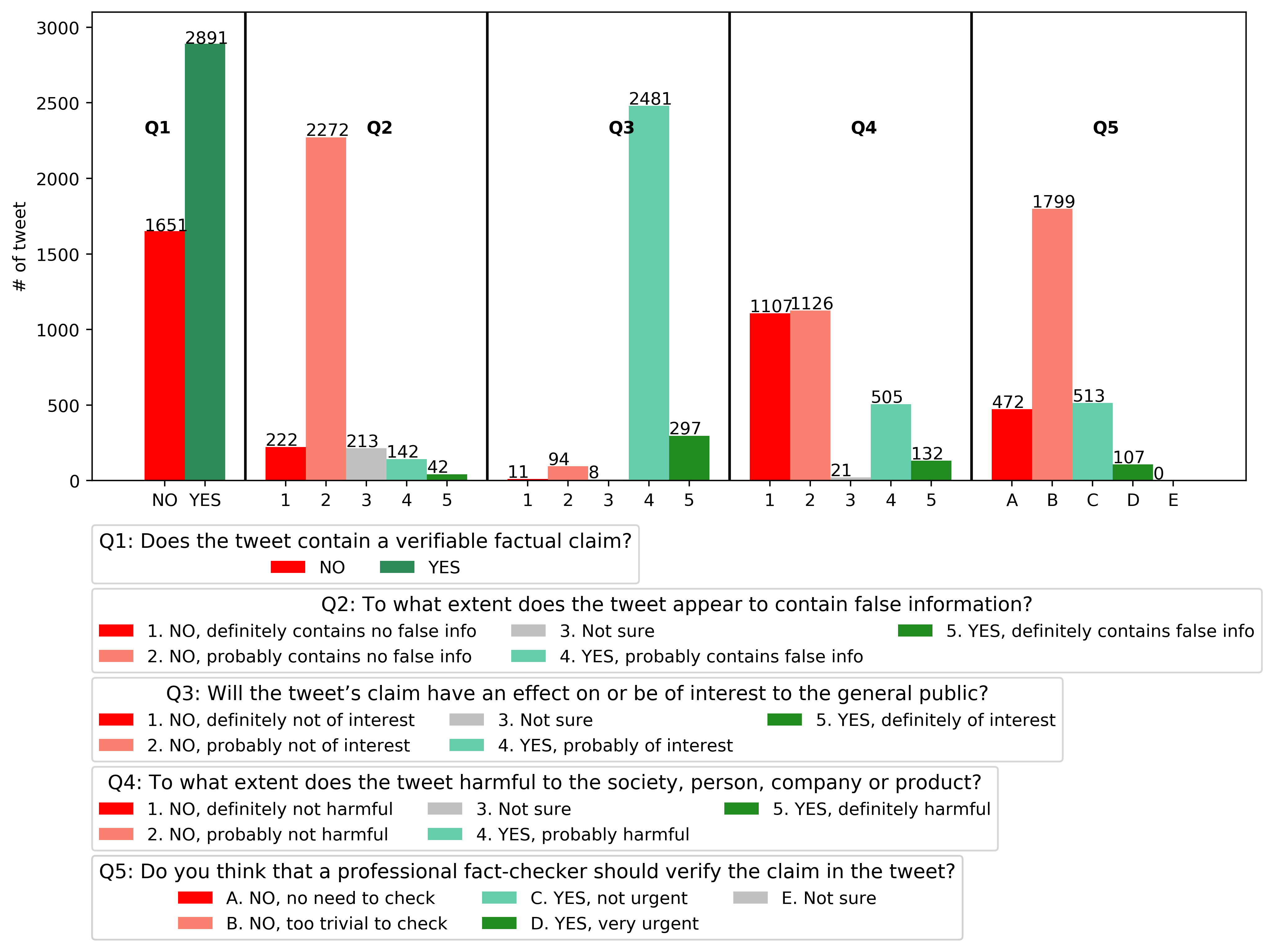}
        \caption{Questions (Q1-5).}
        \label{fig:data_dist_group1}
    \end{subfigure}
    \hfill
    \begin{subfigure}[b]{0.75\textwidth}
        \includegraphics[width=\textwidth]{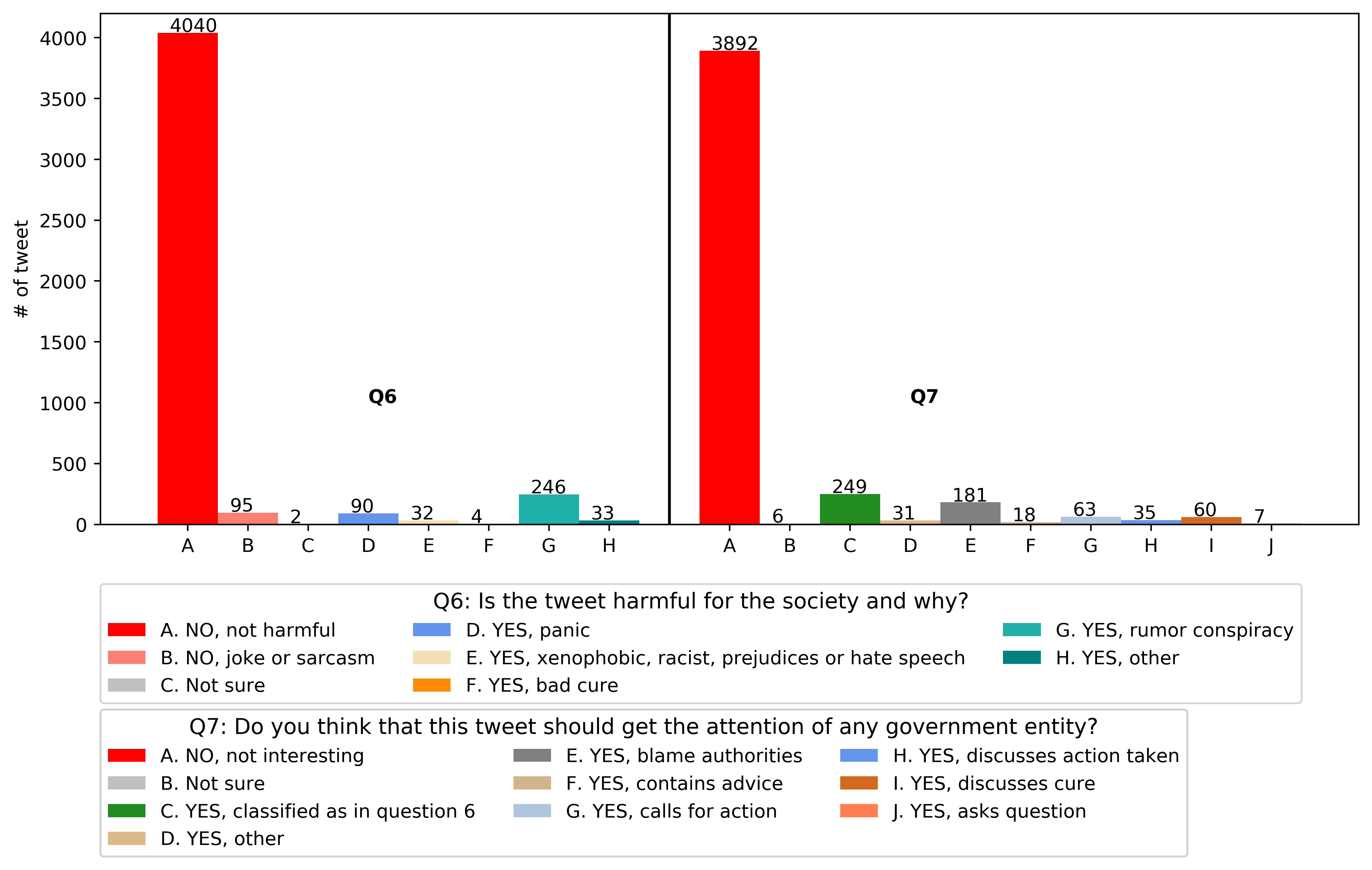}
        \caption{Questions (Q6-7).}
        \label{fig:data_dist_group2}
    \end{subfigure}
    \caption{Distribution of class labels for \textbf{English tweets}}
    \label{fig:data_dist_english_all}
\end{figure*}

\begin{figure*}[tbh]
\centering
    \begin{subfigure}[b]{0.75\textwidth}
        \includegraphics[width=\textwidth]{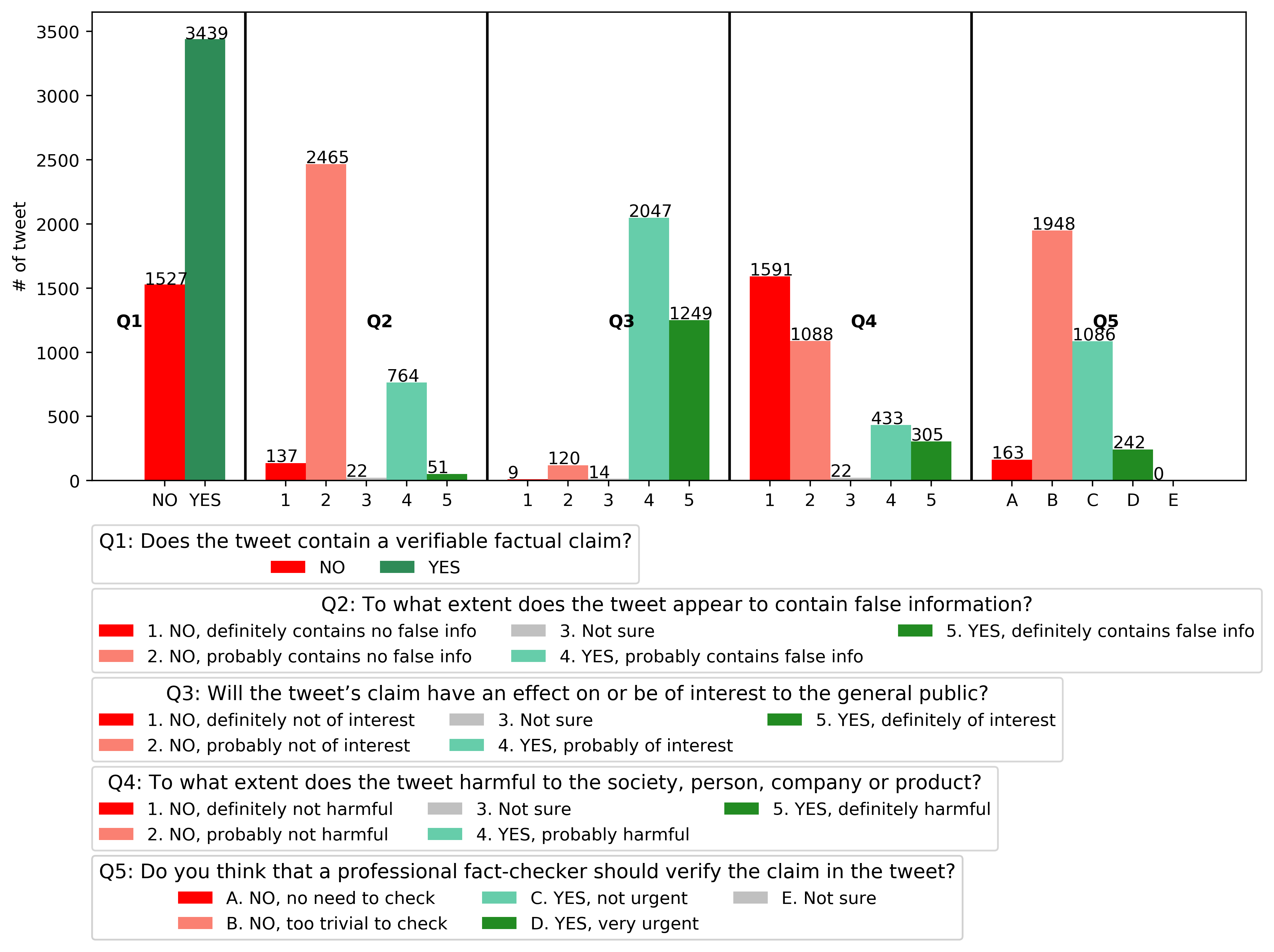}
        \caption{Questions (Q1-5).}
        \label{fig:data_dist_group1_arabic}
    \end{subfigure}
    \hfill
    \begin{subfigure}[b]{0.75\textwidth}    
        \includegraphics[width=\textwidth]{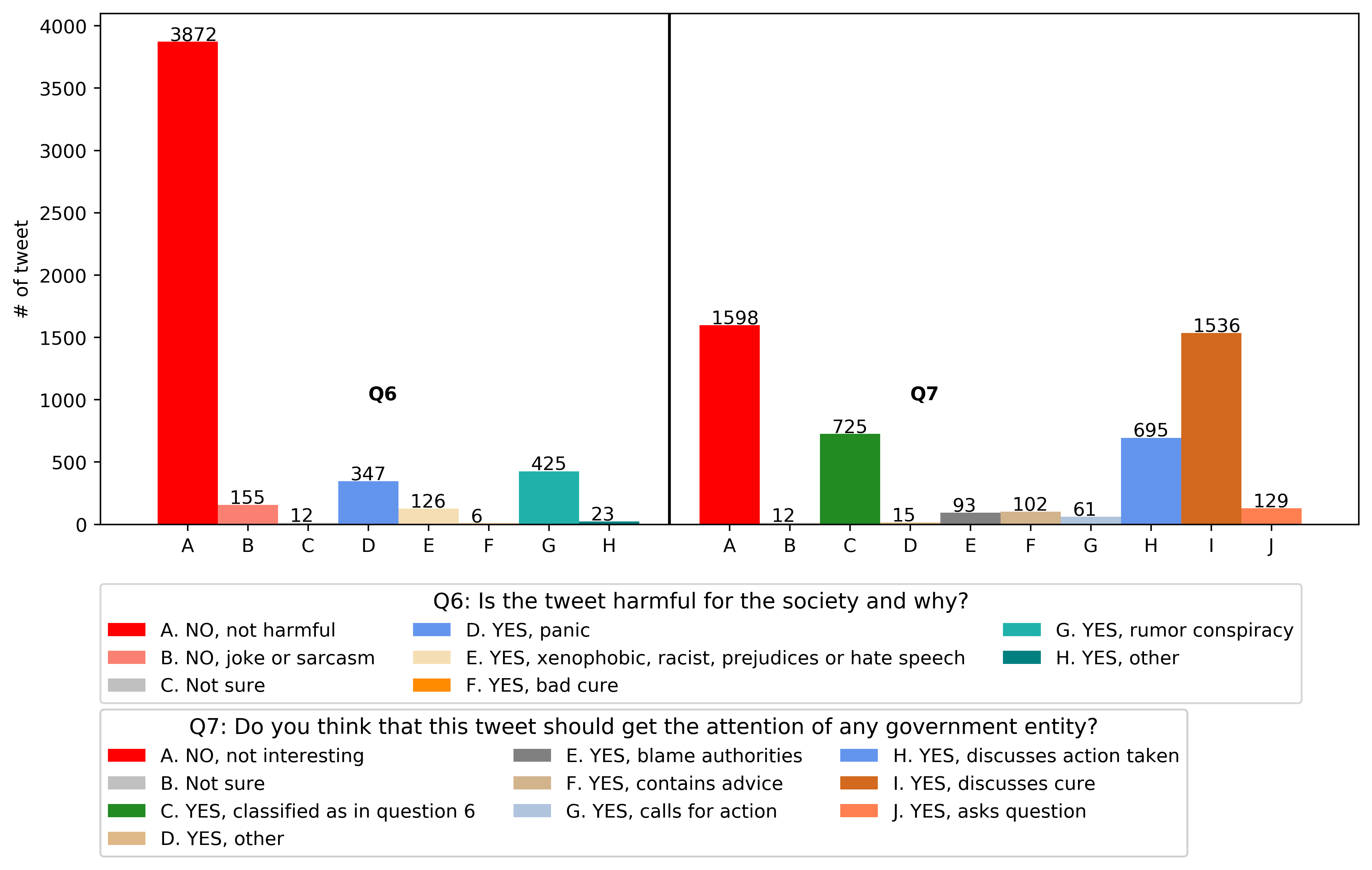}
        \caption{Questions (Q6-7).}
        \label{fig:data_dist_group2_arabic}
    \end{subfigure}
    \caption{Distribution of class labels for \textbf{Arabic tweets}}
    \label{fig:data_dist_arabic_all}
\end{figure*}

\begin{figure*}[tbh]
\centering
    \begin{subfigure}[b]{0.75\textwidth}
        \includegraphics[width=\textwidth]{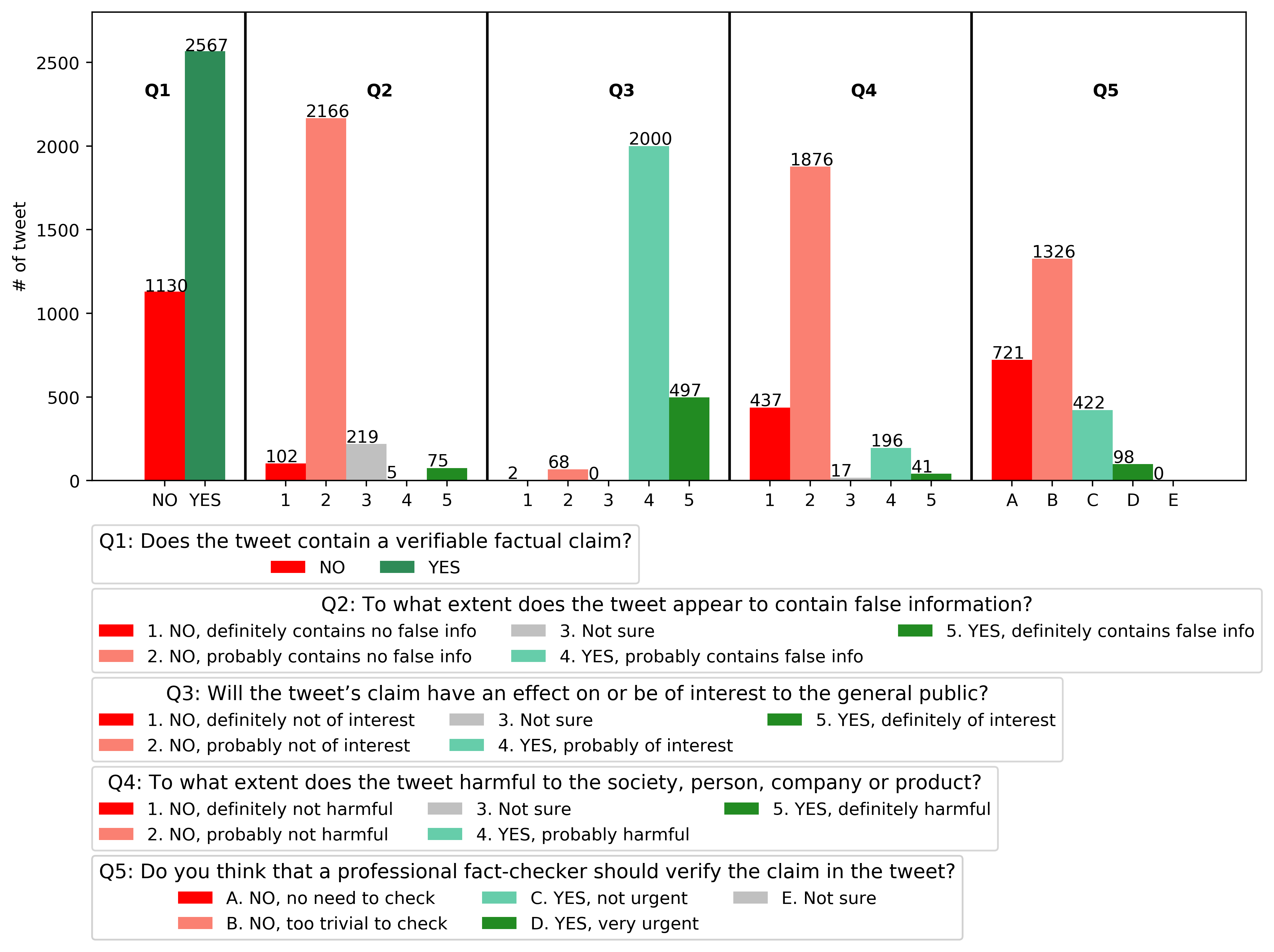}
        \caption{Questions (Q1-5).}
        \label{fig:data_dist_group1_bugarian}
    \end{subfigure}
    \hfill
    \begin{subfigure}[b]{0.75\textwidth}    
        \includegraphics[width=\textwidth]{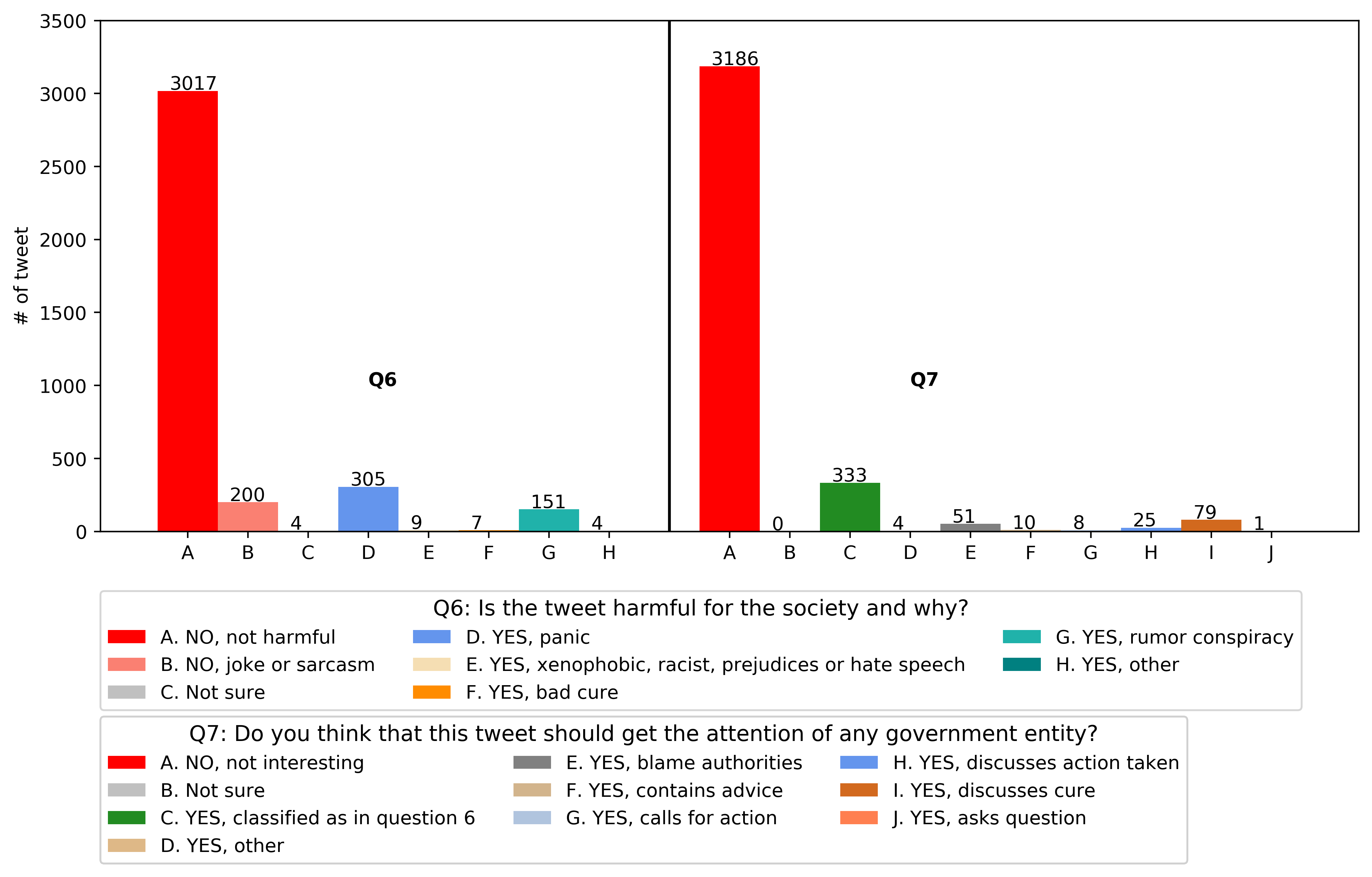}
        \caption{Questions (Q6-7).}
        \label{fig:data_dist_group2_bugarian}
    \end{subfigure}
    \caption{Distribution of class labels for \textbf{Bulgarian tweets}}
    \label{fig:data_dist_bugarian_all}
\end{figure*}

\begin{figure*}[tbh]
\centering
    \begin{subfigure}[b]{0.75\textwidth}
        \includegraphics[width=\textwidth]{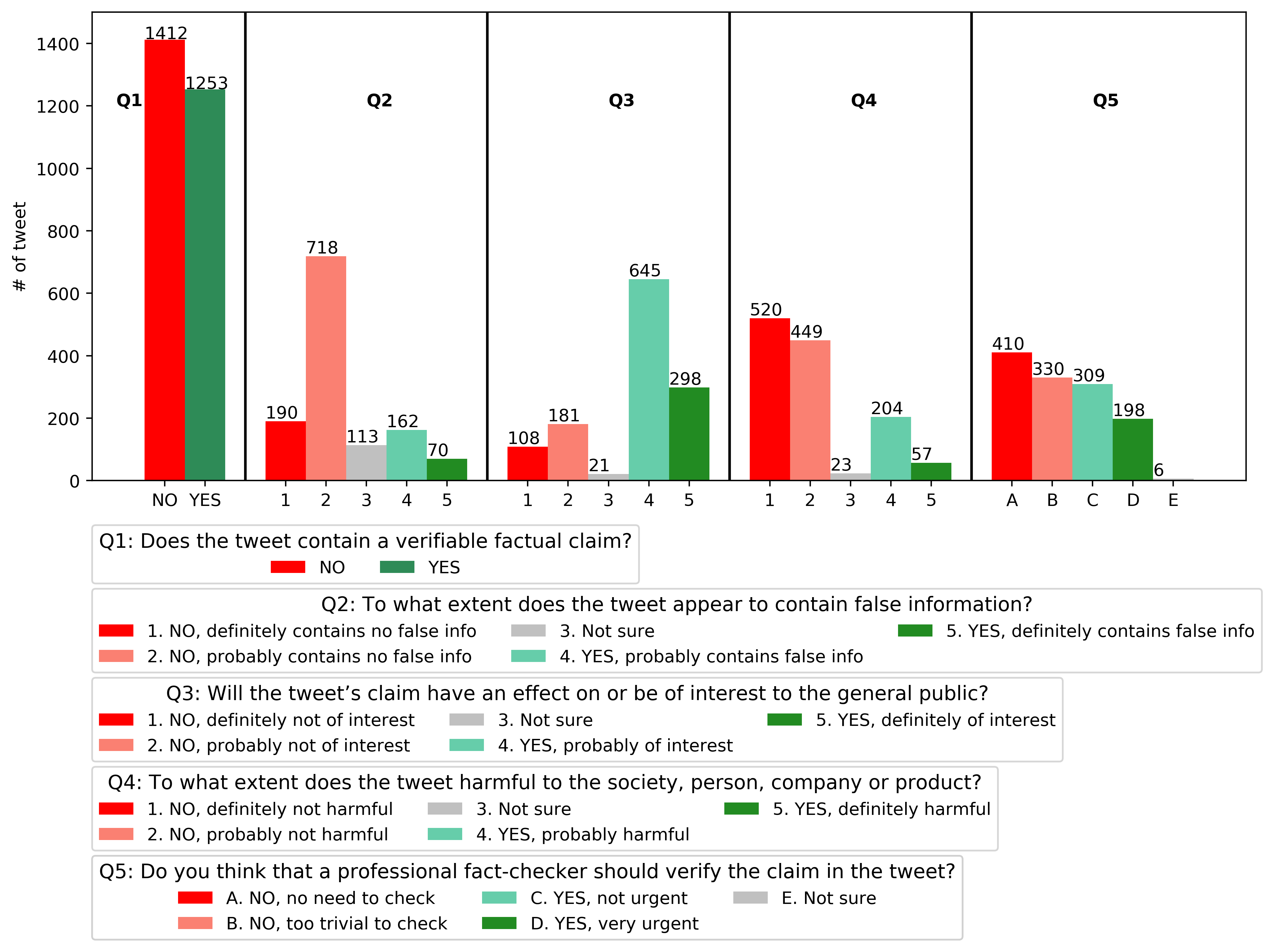}
        \caption{Questions (Q1-5).}
        \label{fig:data_dist_group1_dutch}
    \end{subfigure}
    \hfill
    \begin{subfigure}[b]{0.75\textwidth}    
        \includegraphics[width=\textwidth]{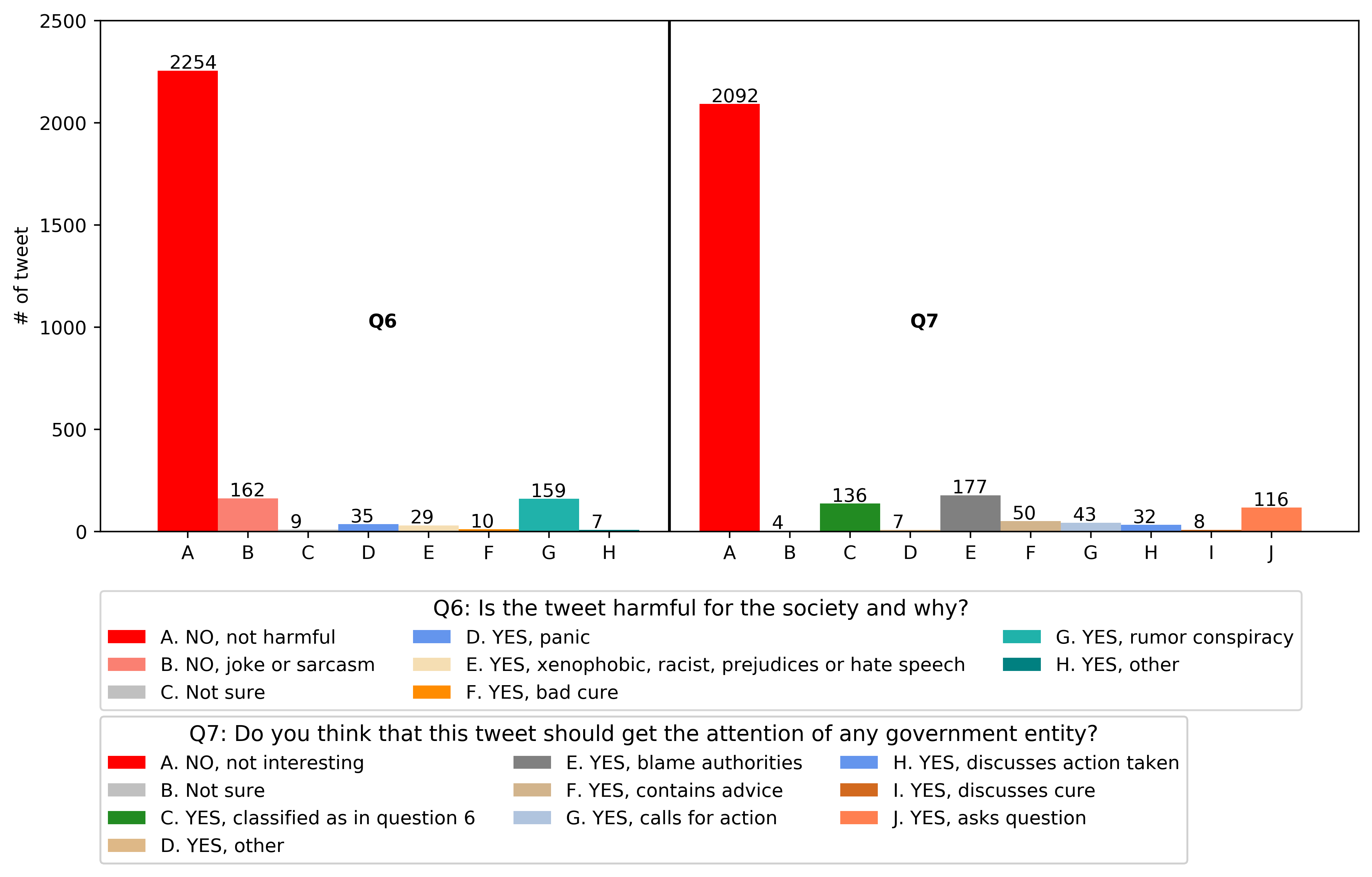}
        \caption{Questions (Q6-7).}
        \label{fig:data_dist_group2_dutch}
    \end{subfigure}
    \caption{Distribution of class labels for \textbf{Dutch tweets}}
    \label{fig:data_dist_dutch_all}
\end{figure*}

Figures \ref{fig:data_dist_english_all}, \ref{fig:data_dist_arabic_all}, \ref{fig:data_dist_bugarian_all} and \ref{fig:data_dist_dutch_all} report detailed statistics about the label distribution for the manual annotations for each question in English, Arabic, Bulgarian, and Dutch, respectively. 

\clearpage
\newpage
\section{Correlation Between Questions}
\label{sec:appendix_correlation}

\begin{figure*}[tbh]
\centering
    \begin{subfigure}[b]{0.4\textwidth}
        \includegraphics[width=\textwidth]{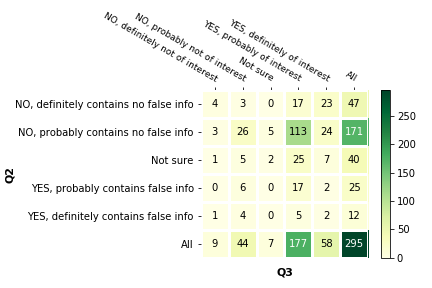}
        \caption{Heatmap for Q2 and Q3.}
        \label{fig:contingency_table_q2_q3}
    \end{subfigure}%
    \begin{subfigure}[b]{0.4\textwidth}    
        \includegraphics[width=\textwidth]{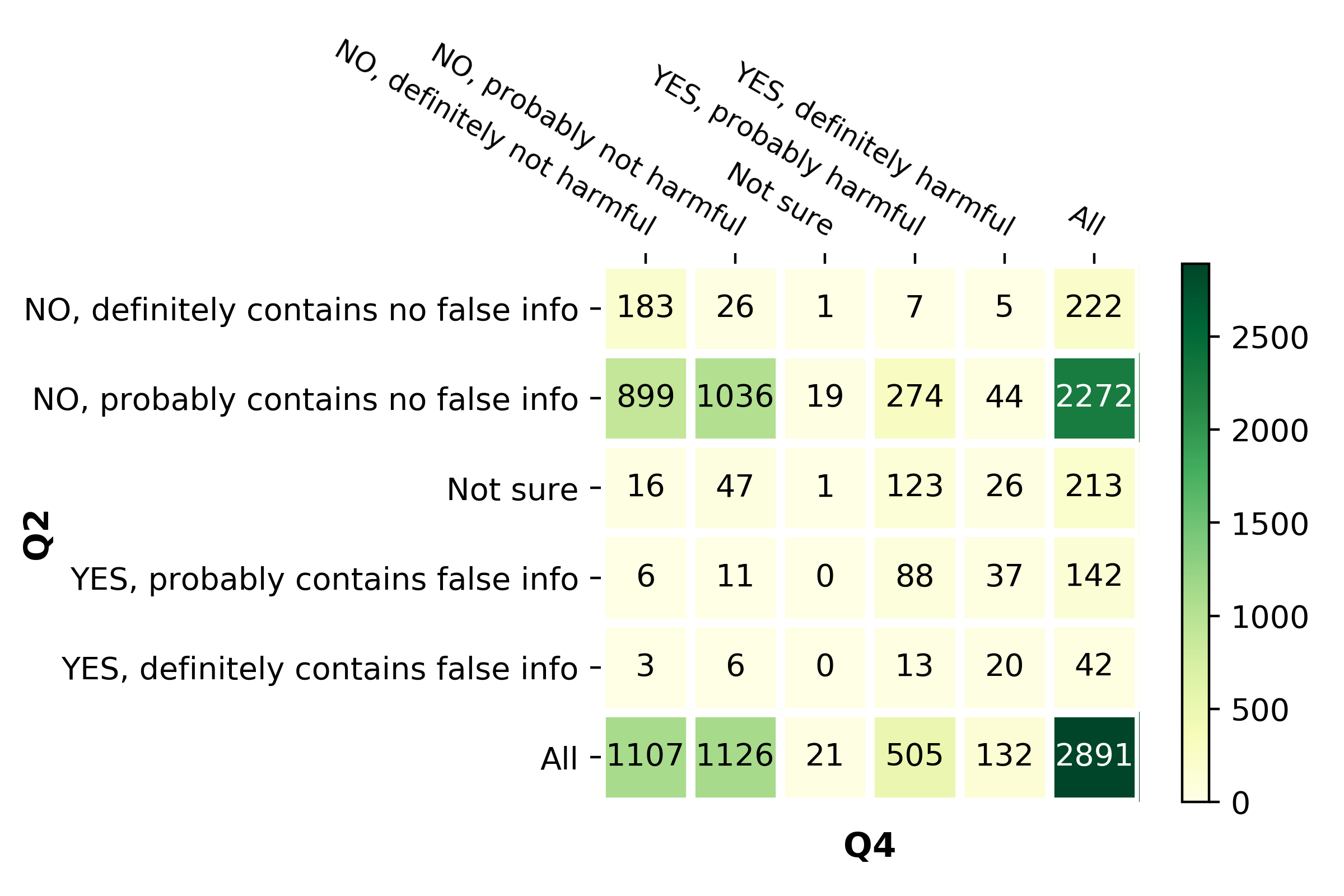}
        \caption{Heatmap for Q2 and Q4.}
        \label{fig:contingency_table_q2_q4}    
    \end{subfigure} 
    \hfill
    \begin{subfigure}[b]{0.4\textwidth}    
        \includegraphics[width=\textwidth]{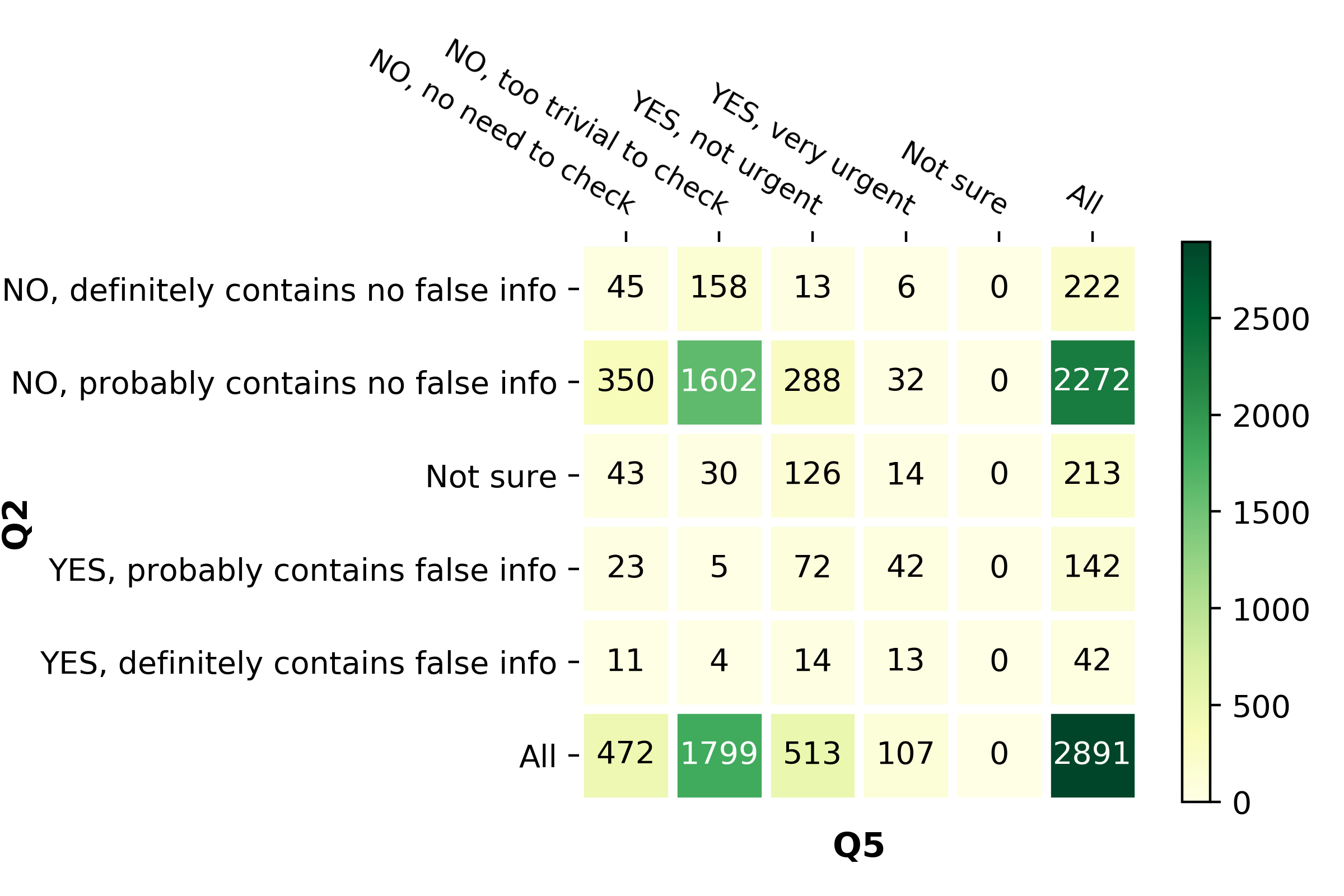}
        \caption{Heatmap for Q2 and Q5.}
        \label{fig:contingency_table_q2_q5}    
    \end{subfigure}%
    \begin{subfigure}[b]{0.4\textwidth}    
        \includegraphics[width=\textwidth]{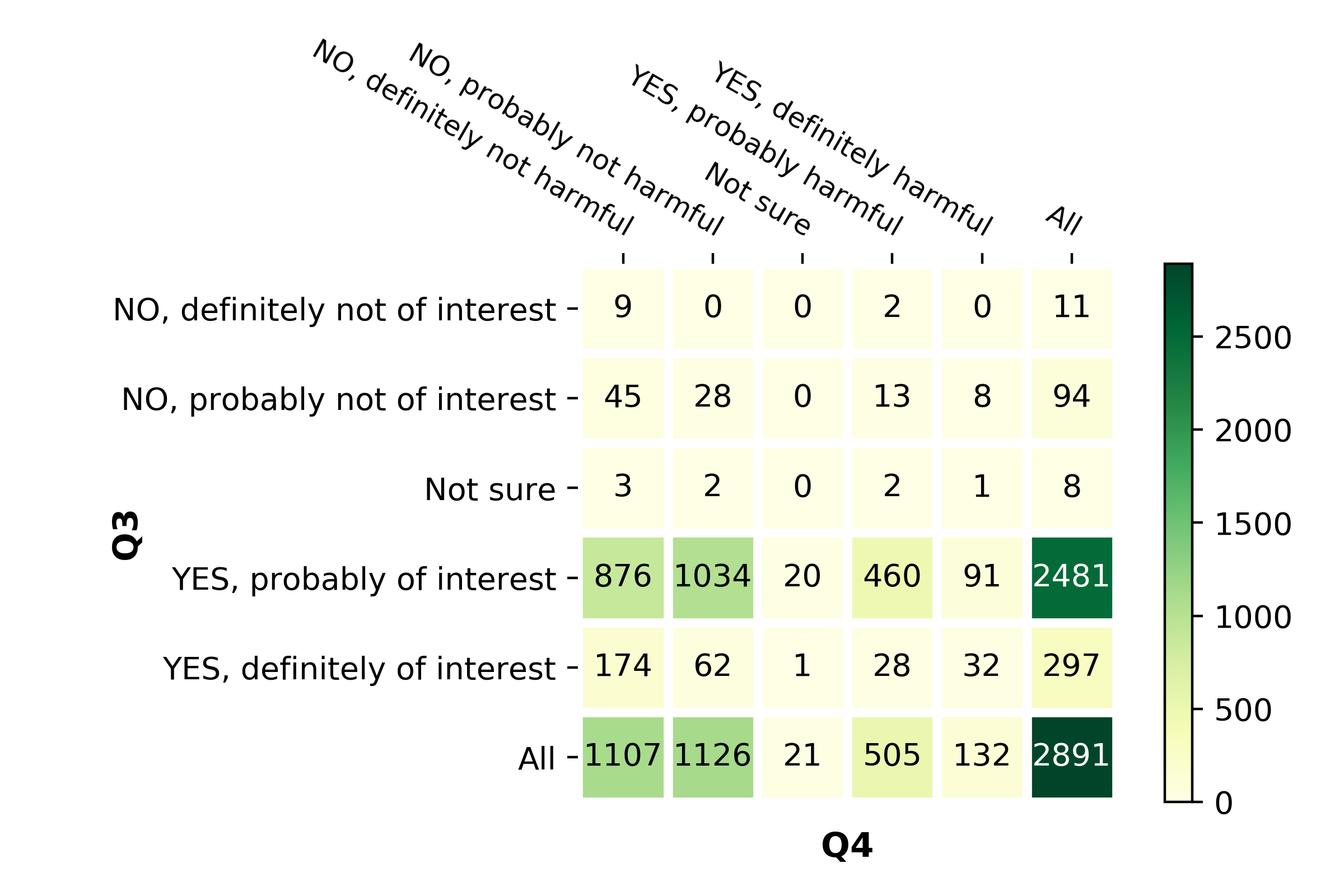}
        \caption{Heatmap for Q3 and Q4.}
        \label{fig:contingency_table_q3_q4}    
    \end{subfigure}
    \hfill
    \begin{subfigure}[b]{0.4\textwidth}    
        \includegraphics[width=\textwidth]{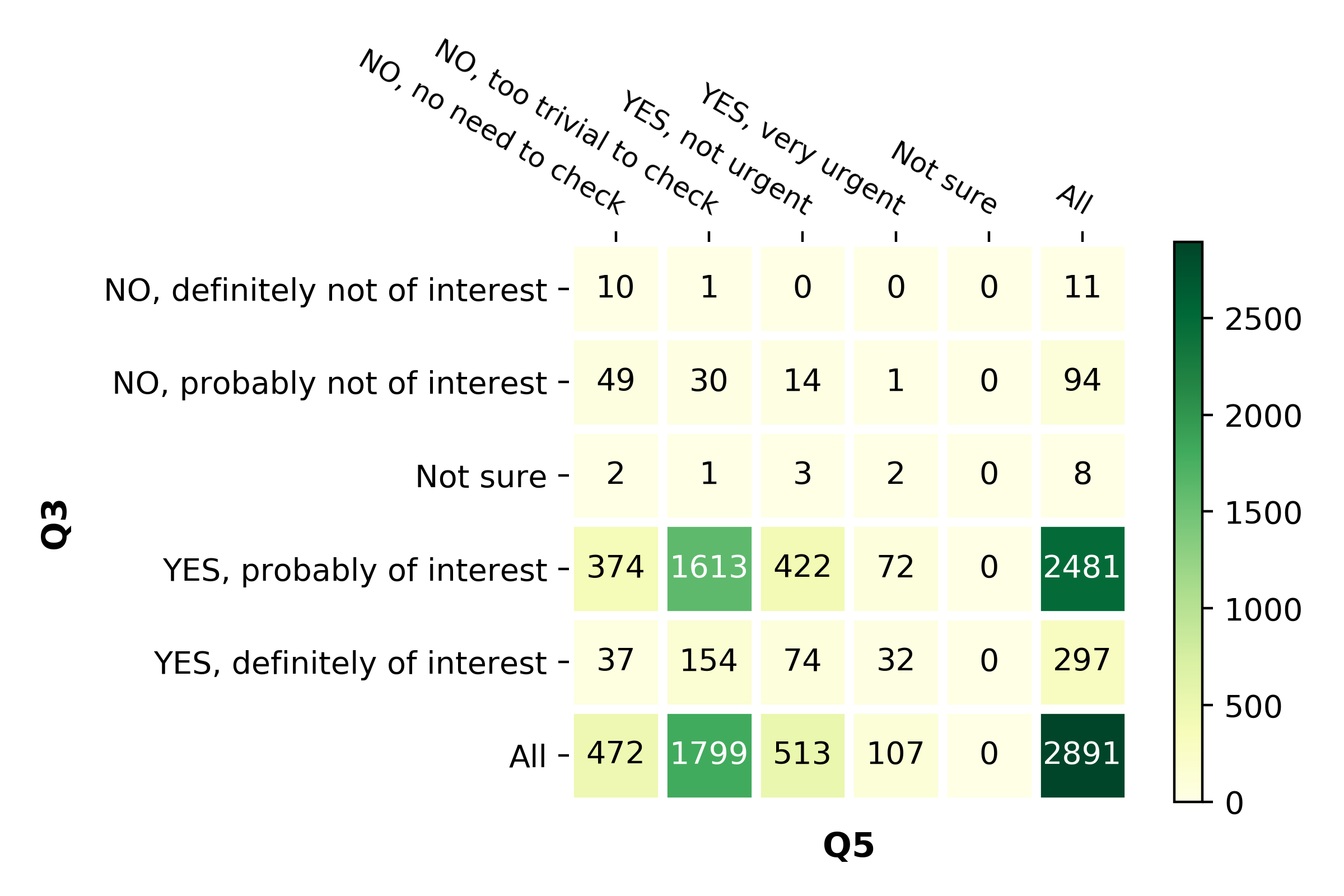}
        \caption{Heatmap for Q3 and Q5.}
        \label{fig:contingency_table_q3_q5}    
    \end{subfigure}%
    \begin{subfigure}[b]{0.4\textwidth}    
        \includegraphics[width=\textwidth]{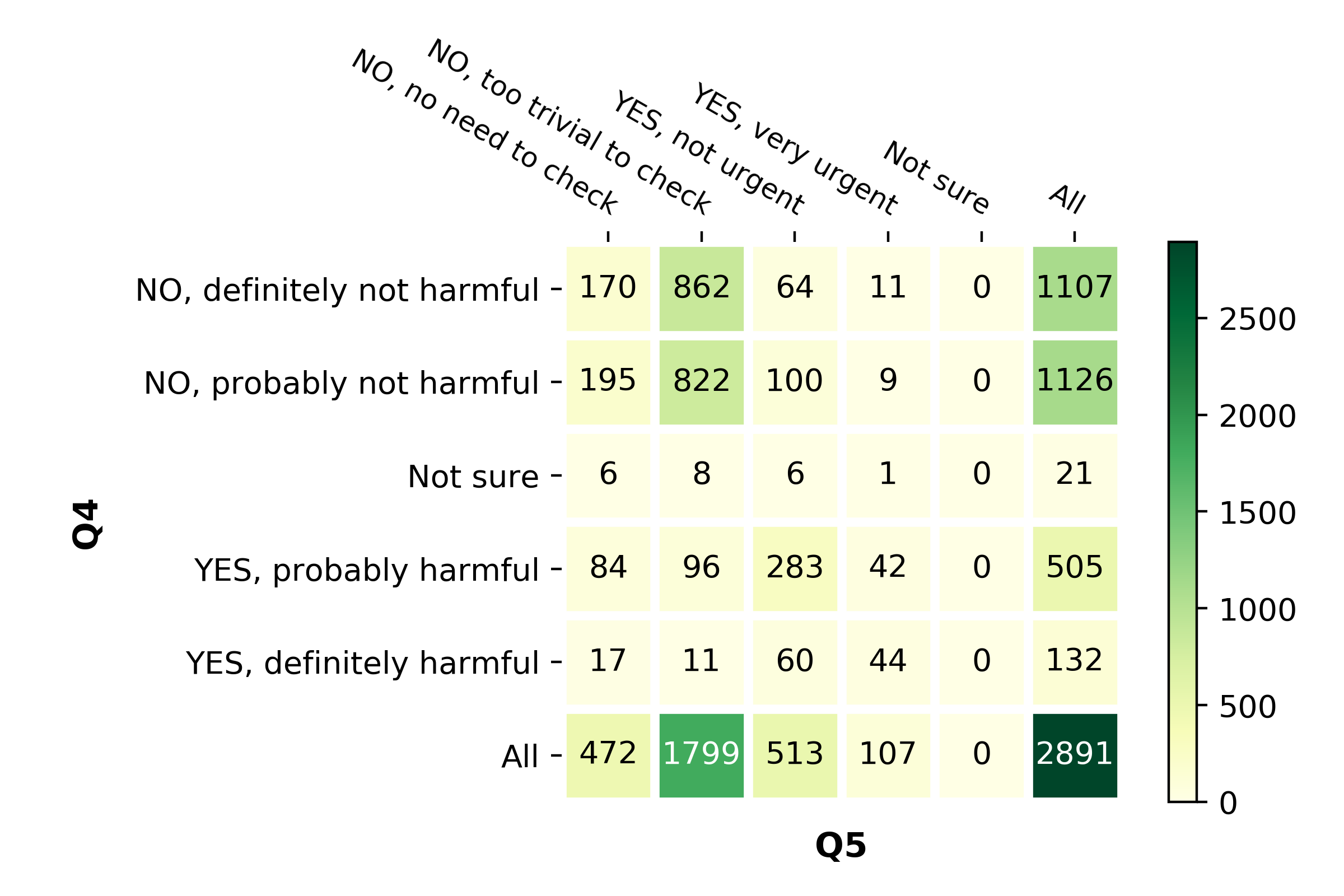}
        \caption{Heatmap for Q4 and Q5.}
        \label{fig:contingency_table_q4_q5}    
    \end{subfigure}     
    \hfill
    \begin{subfigure}[b]{0.6\textwidth}    
        \includegraphics[width=\textwidth]{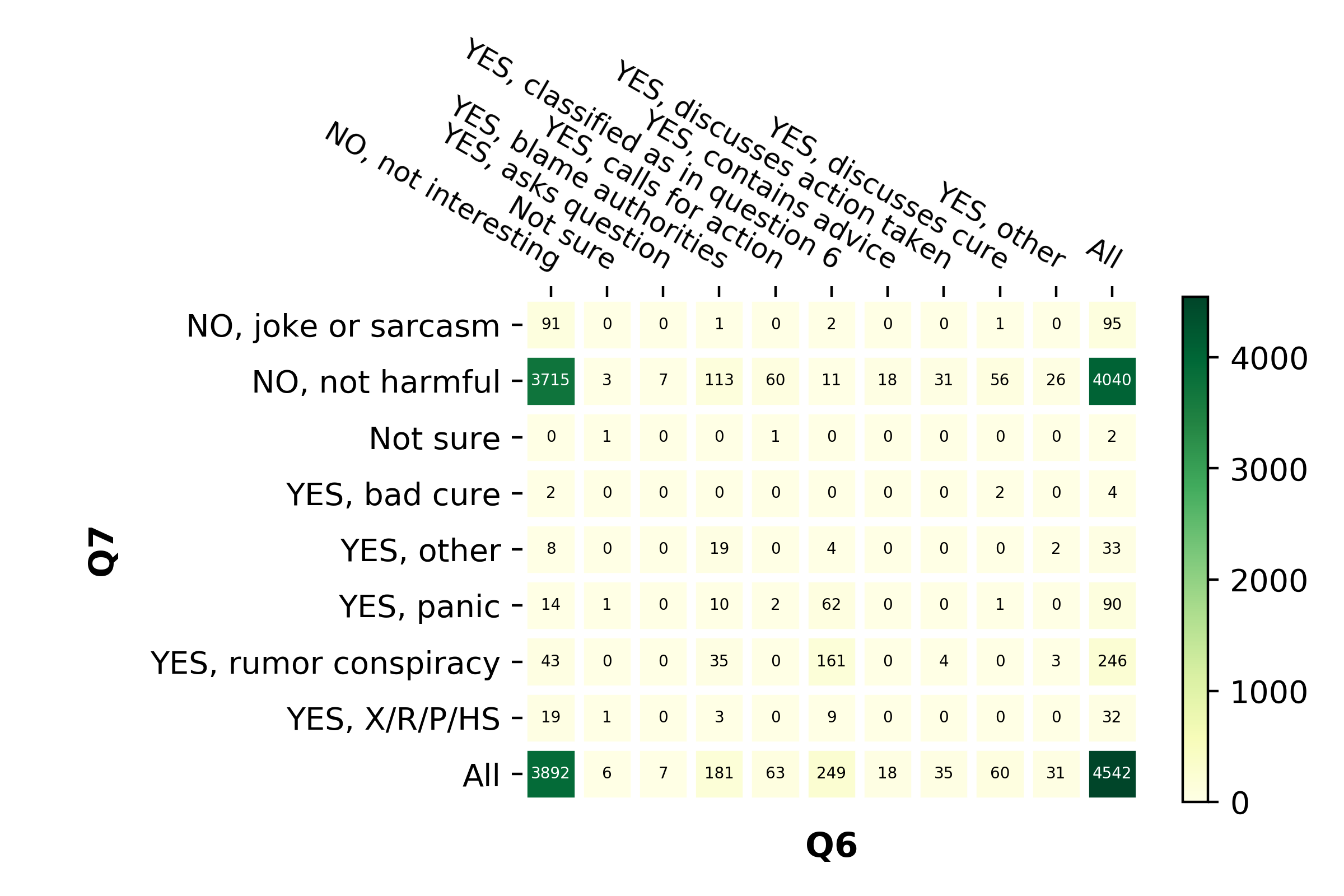}
        \caption{Heatmap for Q6 and Q7. YES, X/R/P/HS -- YES, xenophobic, racist, prejudices or hate speech}
        \label{fig:contingency_table_q6_q7}    
    \end{subfigure}
    \caption{Contingency and correlation heatmaps for the \textbf{English tweets} for different question pairs.}
    \label{fig:contingency_all}
\end{figure*}

\begin{figure*}[tbh]
\centering
    \begin{subfigure}[b]{0.4\textwidth}
        \includegraphics[width=\textwidth]{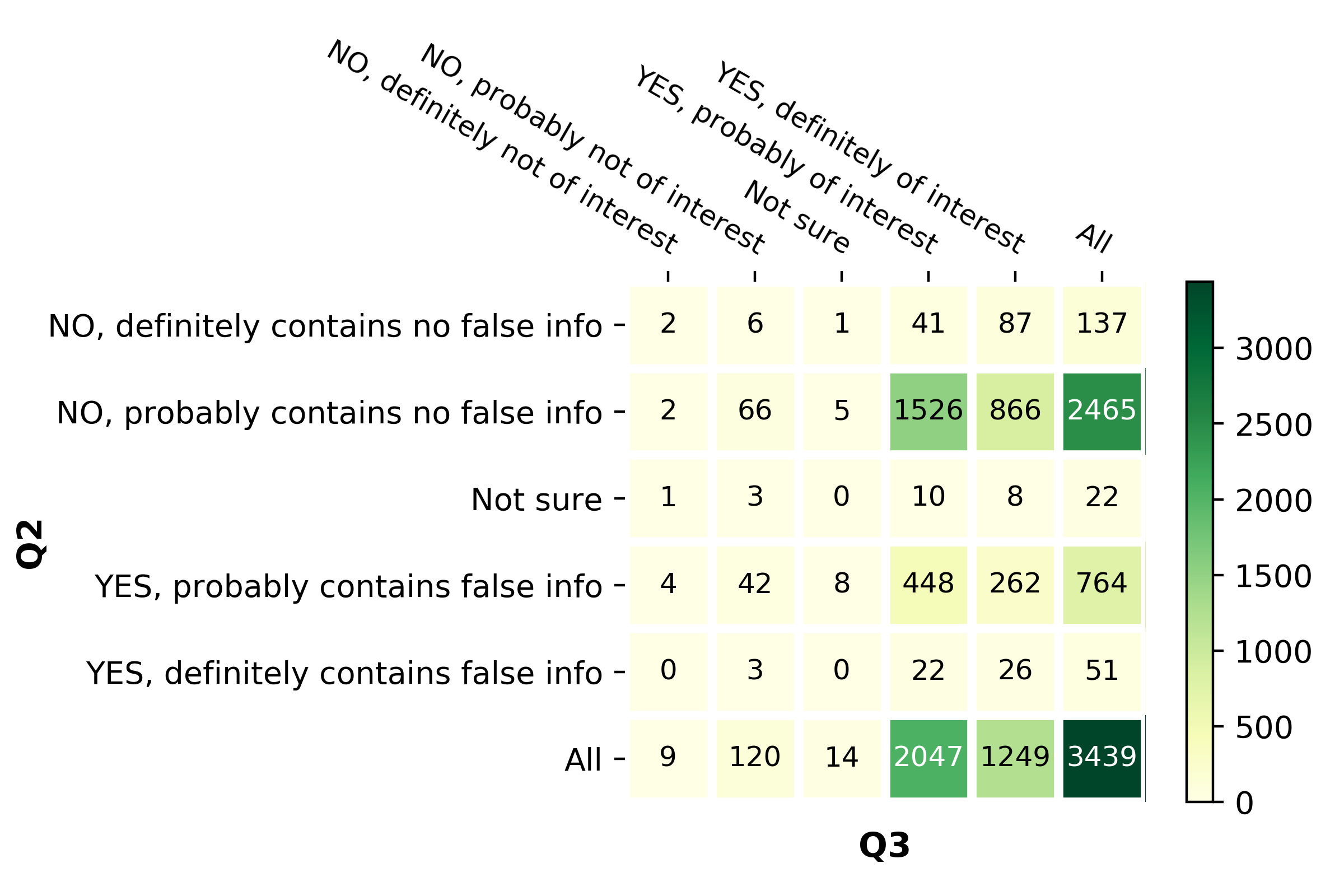}
        \caption{Heatmap for Q2 and Q3.}
        \label{fig:arabic_contingency_table_q2_q3}
    \end{subfigure}%
    \begin{subfigure}[b]{0.4\textwidth}    
        \includegraphics[width=\textwidth]{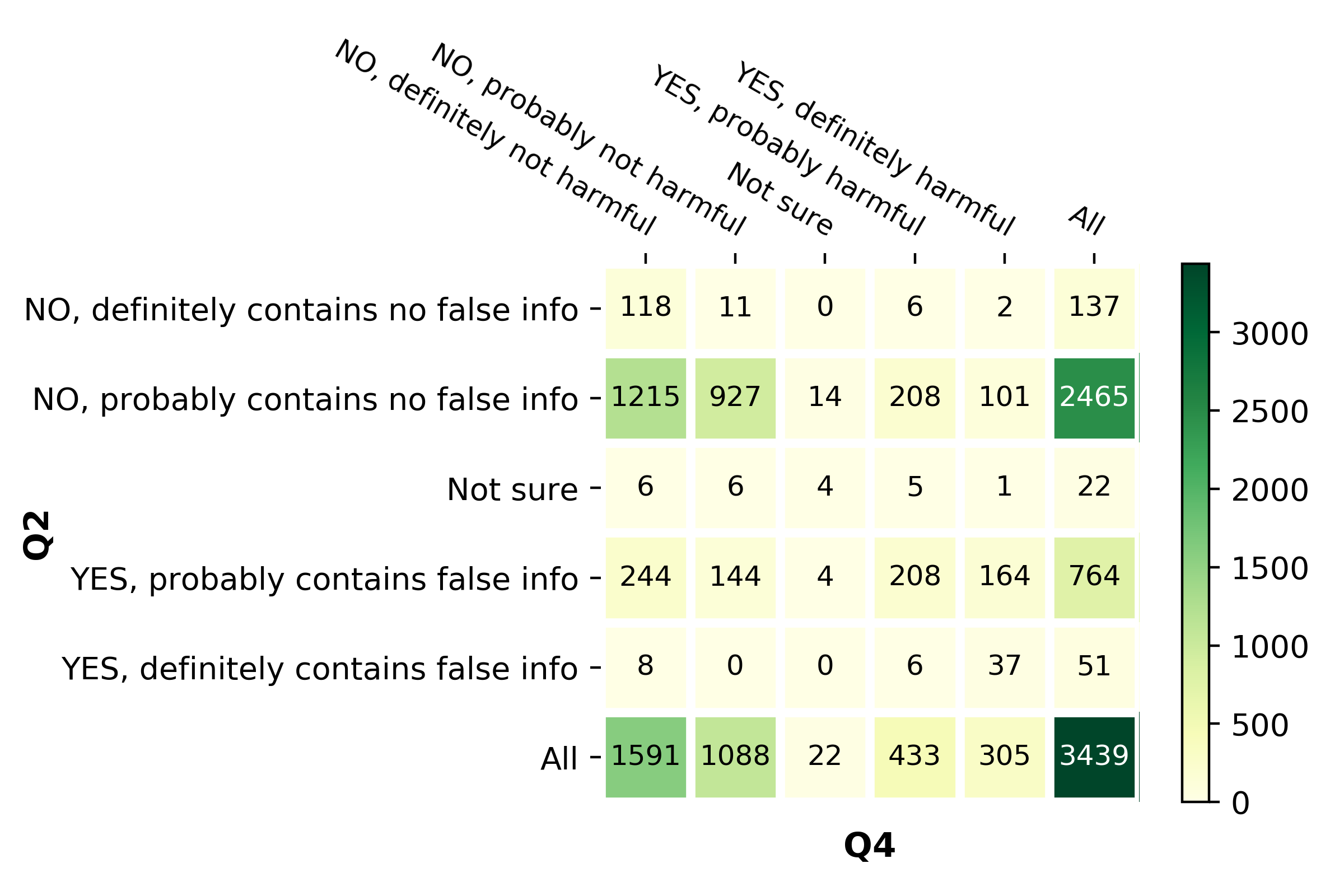}
        \caption{Heatmap for Q2 and Q4.}
        \label{fig:arabic_contingency_table_q2_q4}    
    \end{subfigure} 
    \hfill
    \begin{subfigure}[b]{0.4\textwidth}    
        \includegraphics[width=\textwidth]{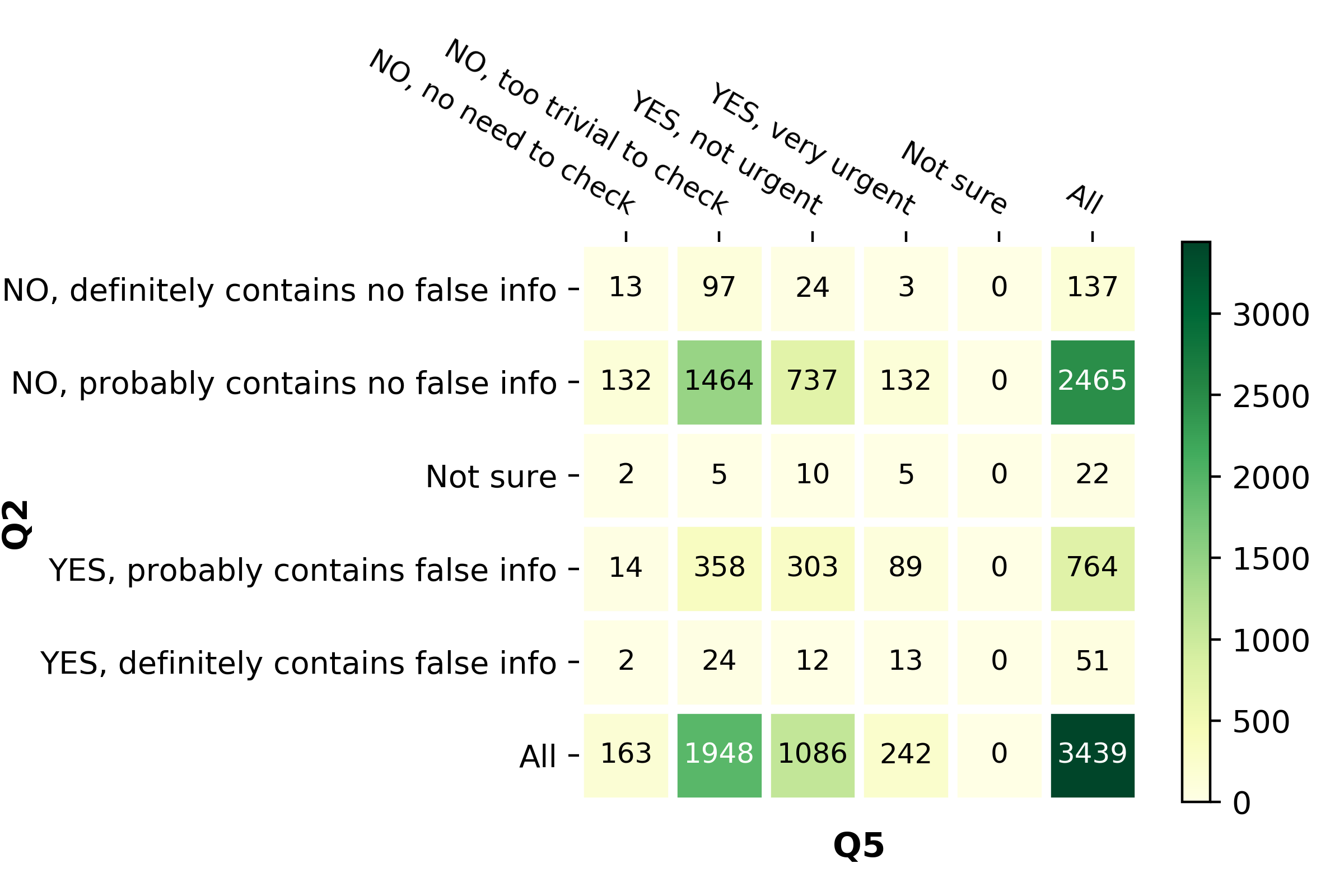}
        \caption{Heatmap for Q2 and Q5.}
        \label{fig:arabic_contingency_table_q2_q5}    
    \end{subfigure}%
    \begin{subfigure}[b]{0.4\textwidth}    
        \includegraphics[width=\textwidth]{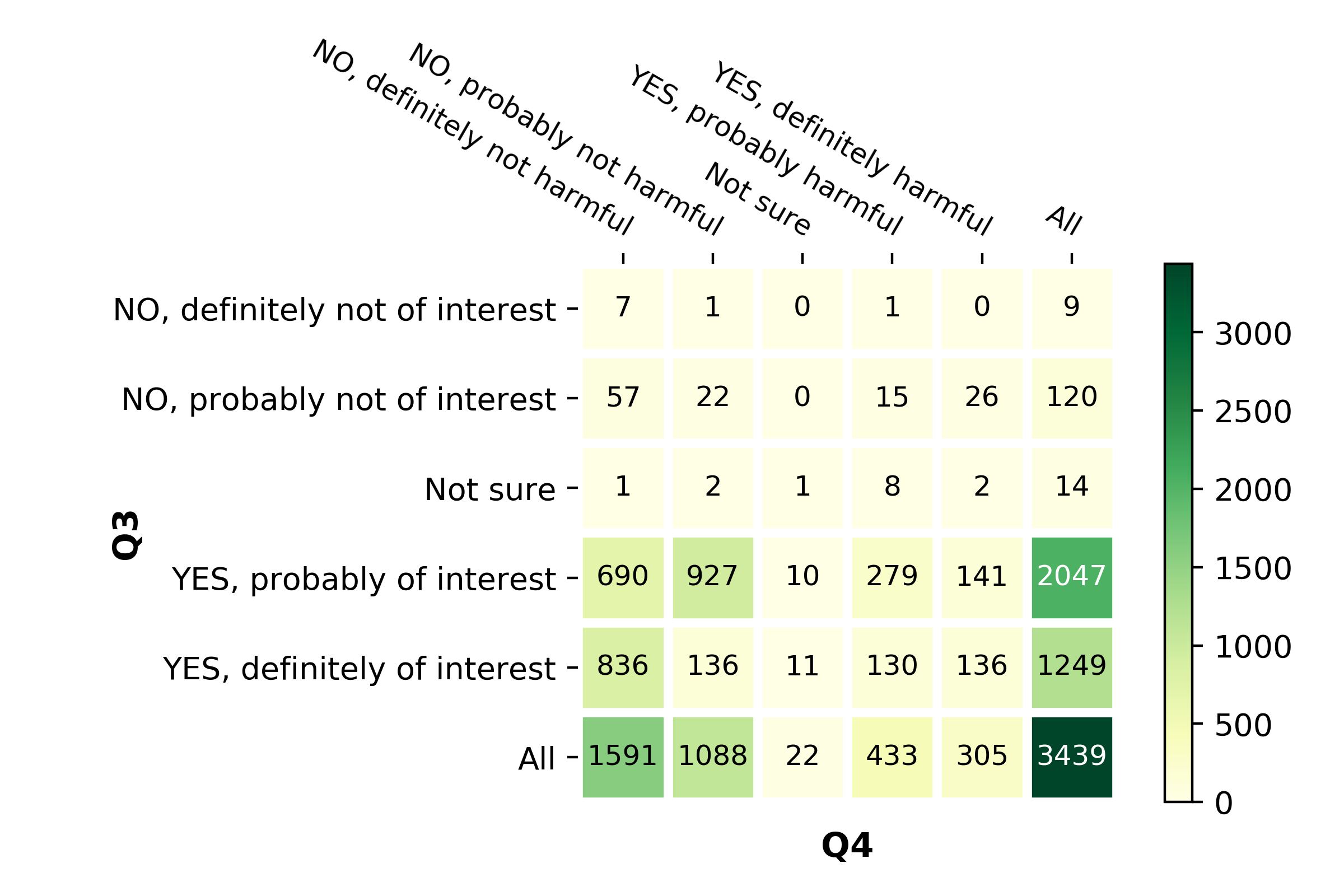}
        \caption{Heatmap for Q3 and Q4.}
        \label{fig:arabic_contingency_table_q3_q4}    
    \end{subfigure}
    \hfill
    \begin{subfigure}[b]{0.4\textwidth}    
        \includegraphics[width=\textwidth]{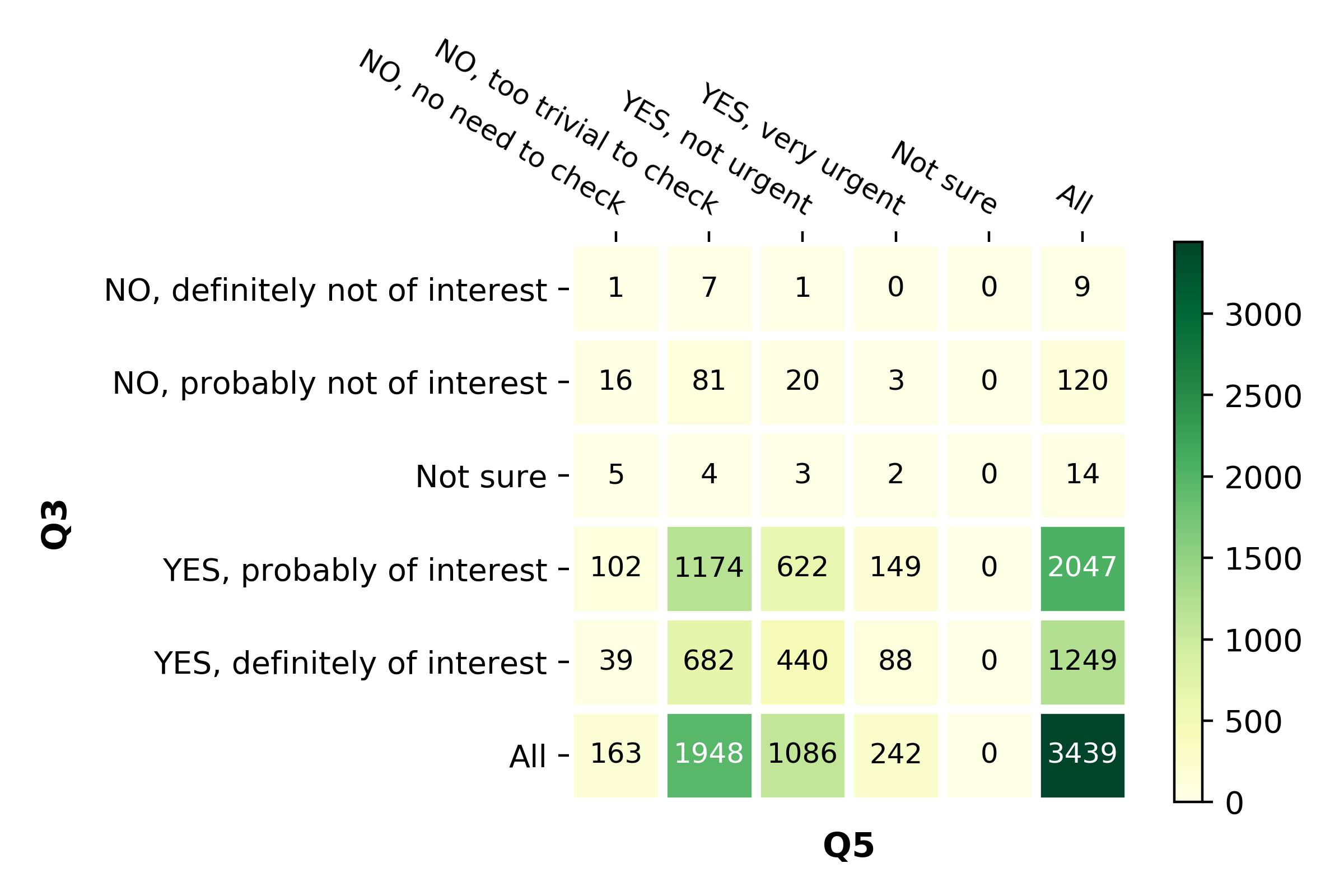}
        \caption{Heatmap for Q3 and Q5.}
        \label{fig:arabic_contingency_table_q3_q5}    
    \end{subfigure}%
    \begin{subfigure}[b]{0.4\textwidth}    
        \includegraphics[width=\textwidth]{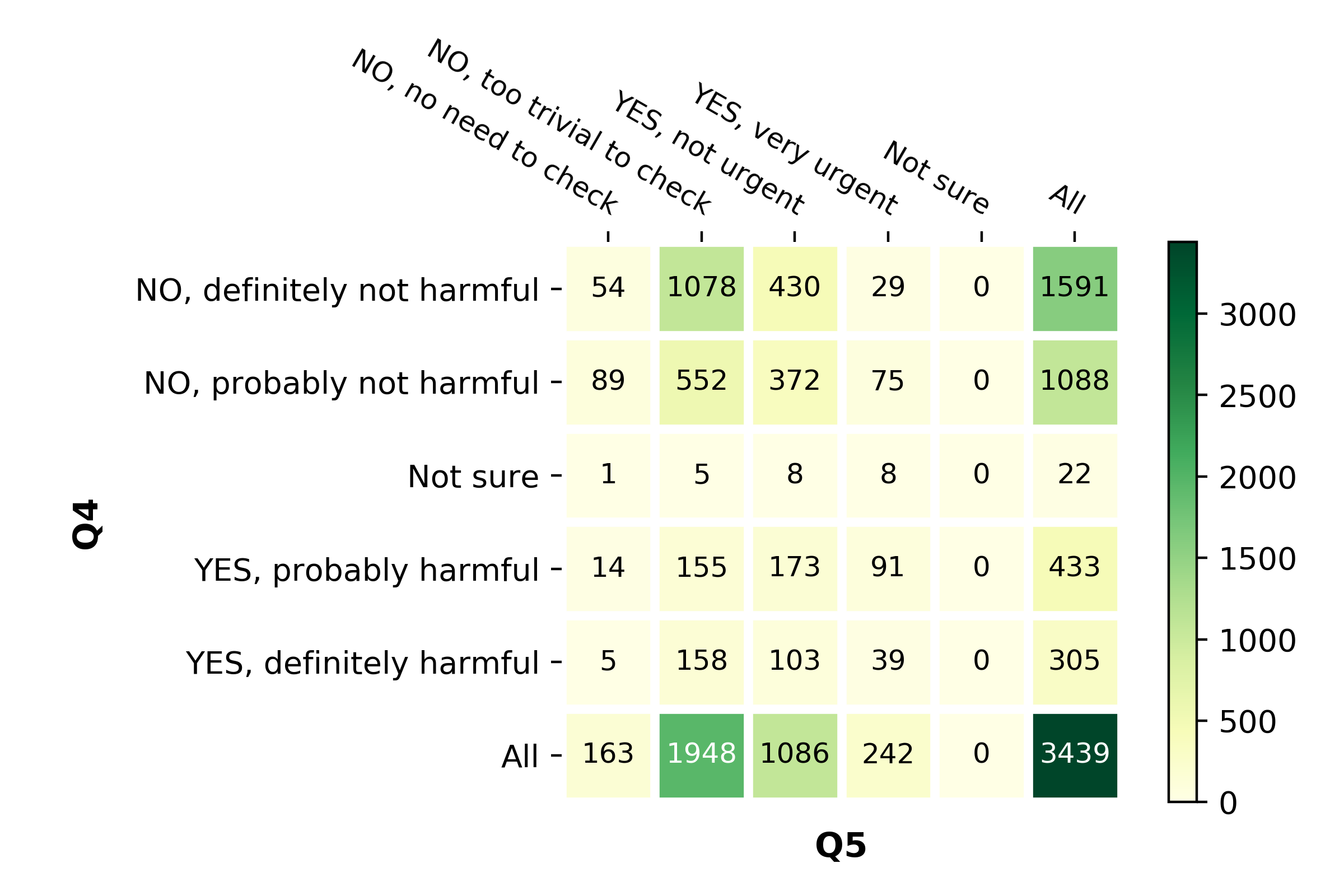}
        \caption{Heatmap for Q4 and Q5.}
        \label{fig:arabic_contingency_table_q4_q5}    
    \end{subfigure}     
    \hfill
    \begin{subfigure}[b]{0.6\textwidth}    
        \includegraphics[width=\textwidth]{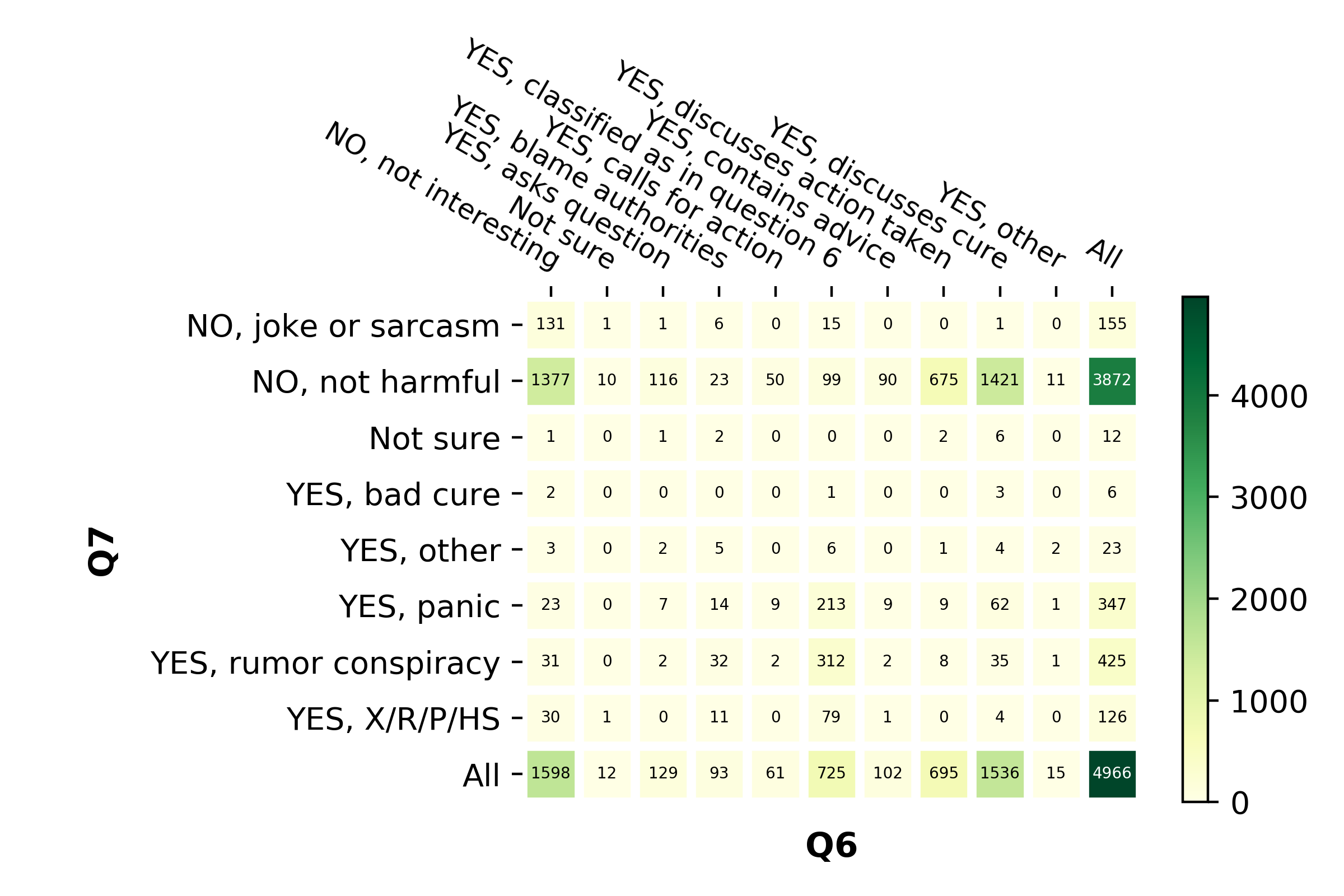}
        \caption{Heatmap for Q6 and Q7. YES, X/R/P/HS -- YES, xenophobic, racist, prejudices or hate speech}
        \label{fig:arabic_contingency_table_q6_q7} 
    \end{subfigure}
    \caption{Contingency and correlation heatmaps for the \textbf{Arabic tweets} for different question pairs.}
    \label{fig:arabic_contingency_all}
\end{figure*}

\begin{figure*}[tbh]
\centering
    \begin{subfigure}[b]{0.4\textwidth}
        \includegraphics[width=\textwidth]{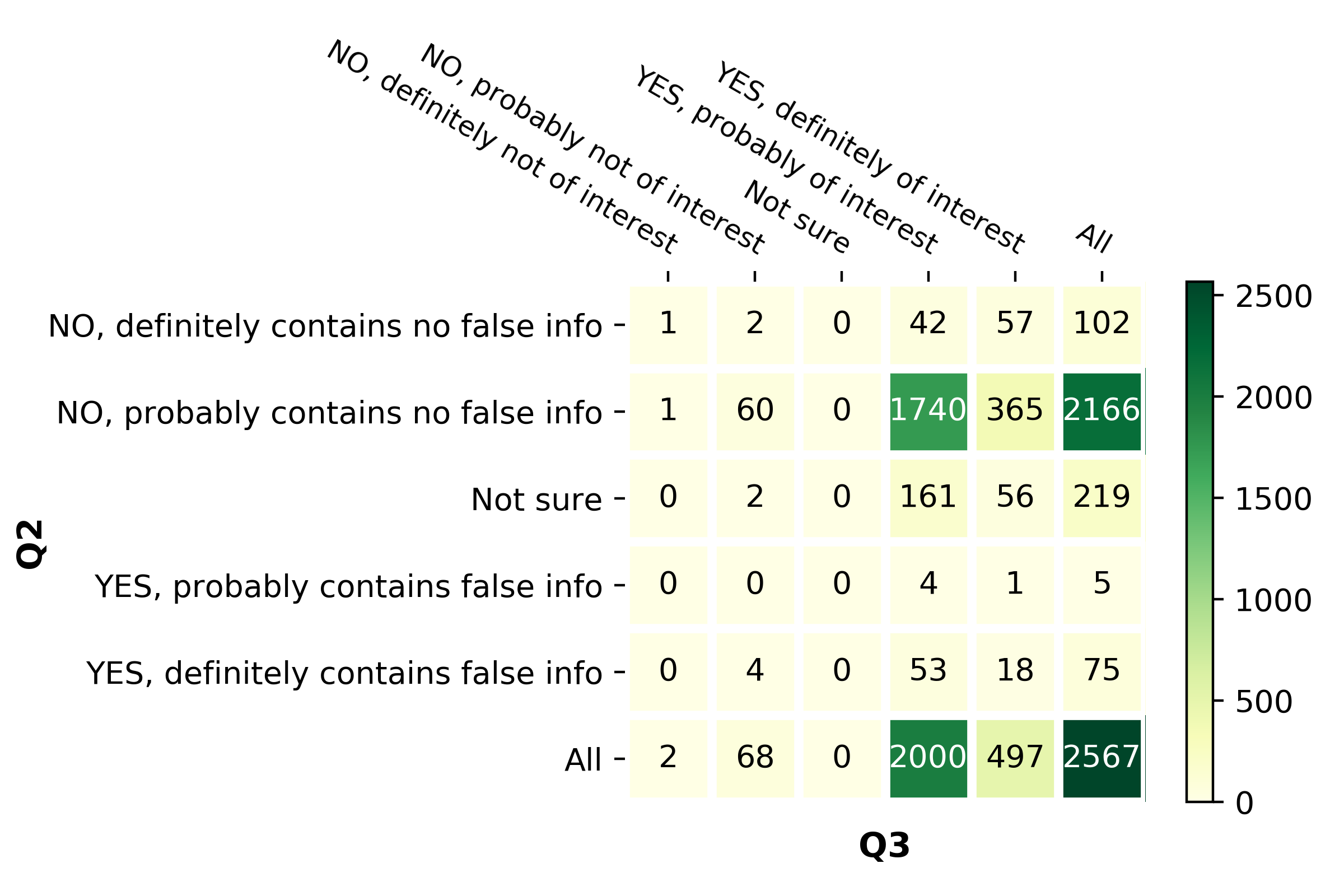}
        \caption{Heatmap for Q2 and Q3.}
        \label{fig:bulgarian_contingency_table_q2_q3}
    \end{subfigure}%
    \begin{subfigure}[b]{0.4\textwidth}    
        \includegraphics[width=\textwidth]{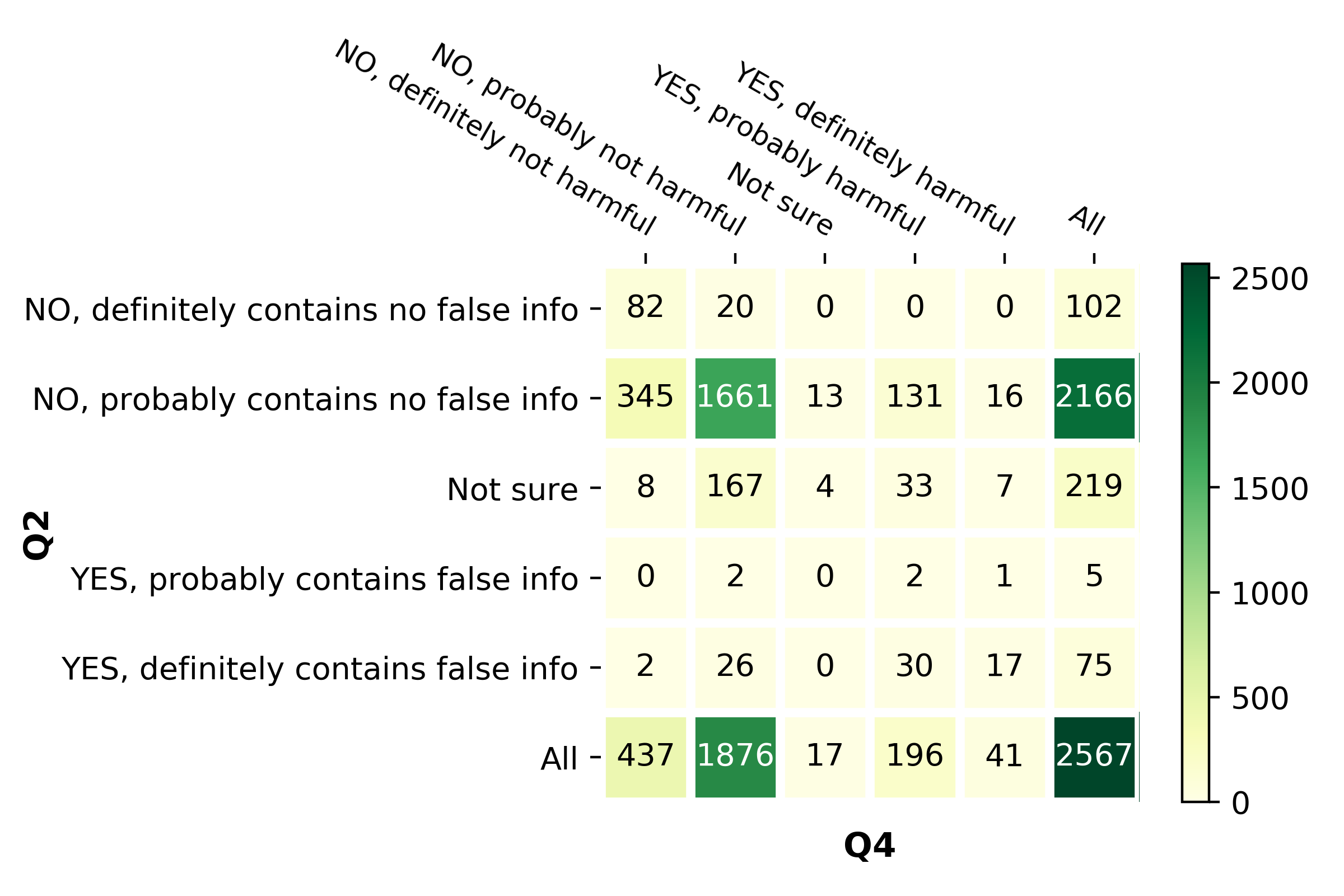}
        \caption{Heatmap for Q2 and Q4.}
        \label{fig:bulgarian_contingency_table_q2_q4}    
    \end{subfigure} 
    \hfill
    \begin{subfigure}[b]{0.4\textwidth}    
        \includegraphics[width=\textwidth]{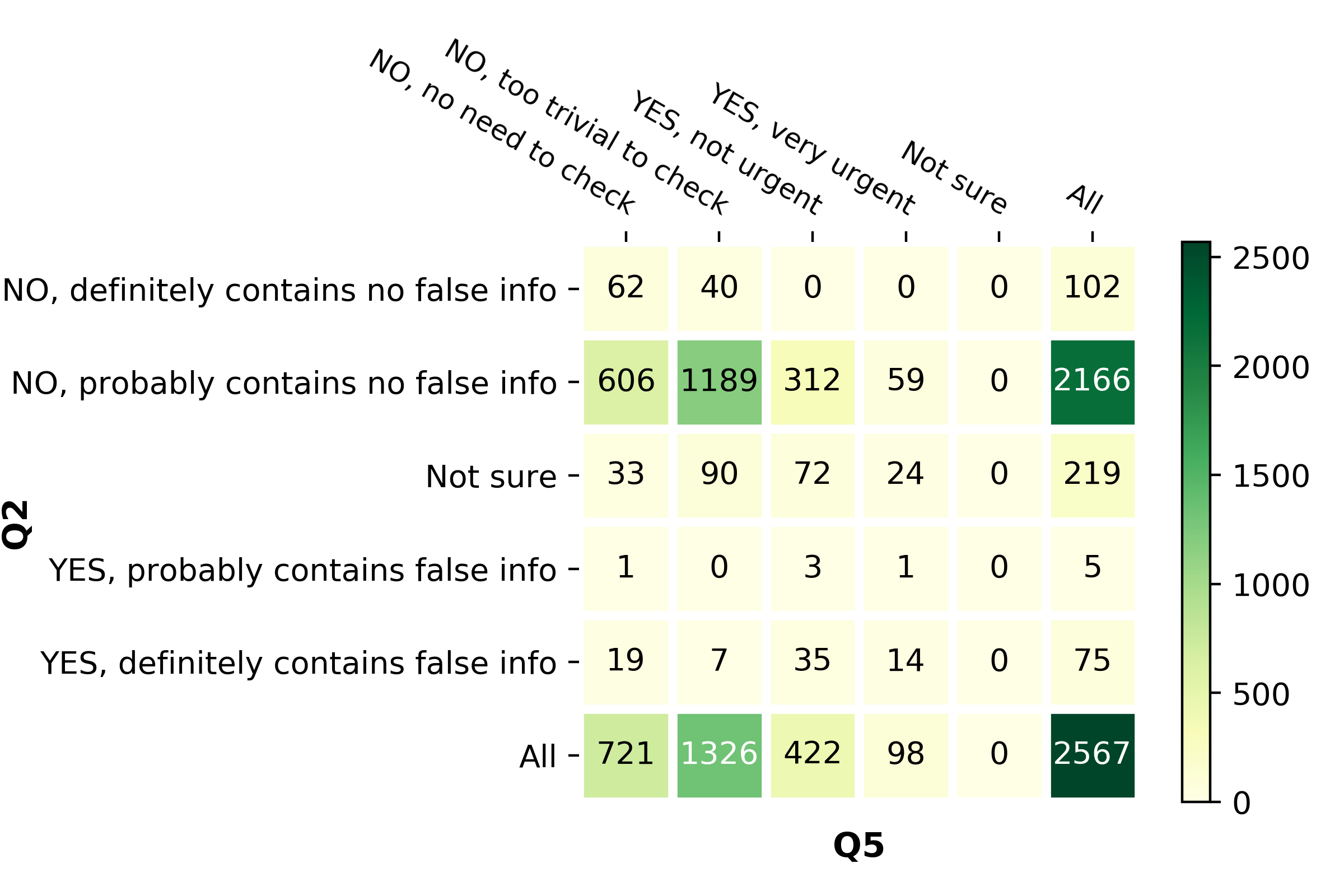}
        \caption{Heatmap for Q2 and Q5.}
        \label{fig:bulgarian_contingency_table_q2_q5}    
    \end{subfigure}%
    \begin{subfigure}[b]{0.4\textwidth}    
        \includegraphics[width=\textwidth]{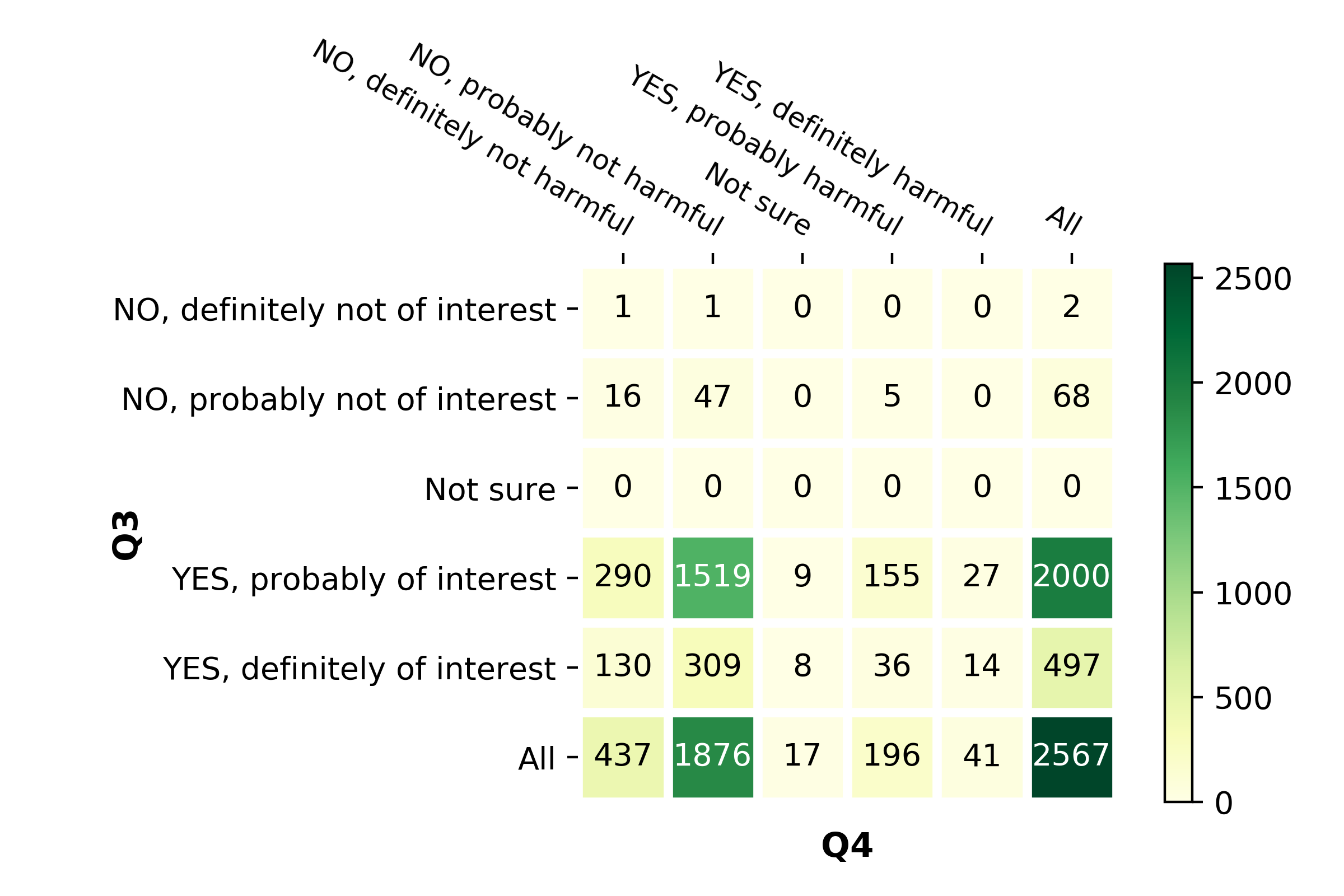}
        \caption{Heatmap for Q3 and Q4.}
        \label{fig:bulgarian_contingency_table_q3_q4}    
    \end{subfigure}
    \hfill
    \begin{subfigure}[b]{0.4\textwidth}    
        \includegraphics[width=\textwidth]{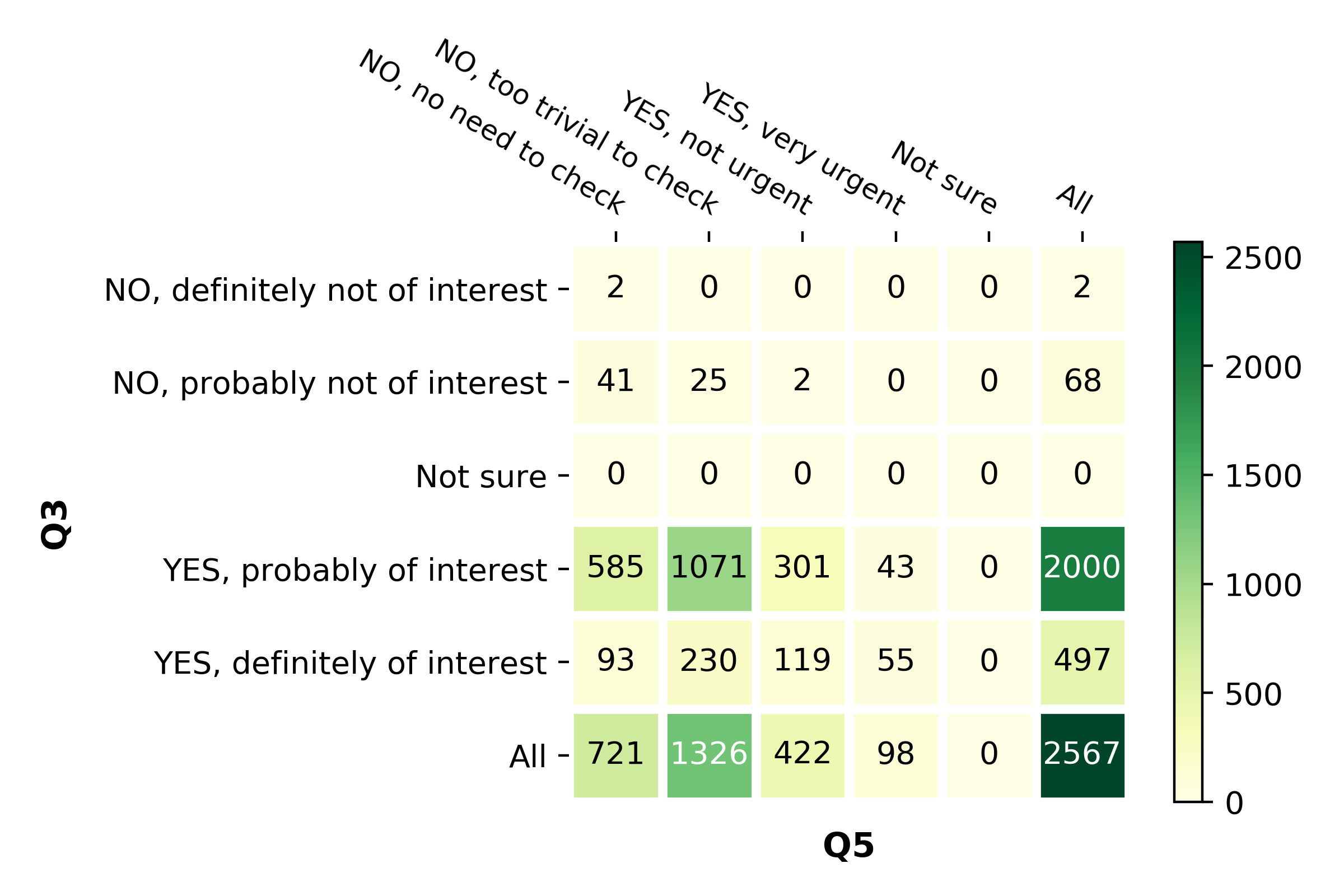}
        \caption{Heatmap for Q3 and Q5.}
        \label{fig:bulgarian_contingency_table_q3_q5}    
    \end{subfigure}%
    \begin{subfigure}[b]{0.4\textwidth}    
        \includegraphics[width=\textwidth]{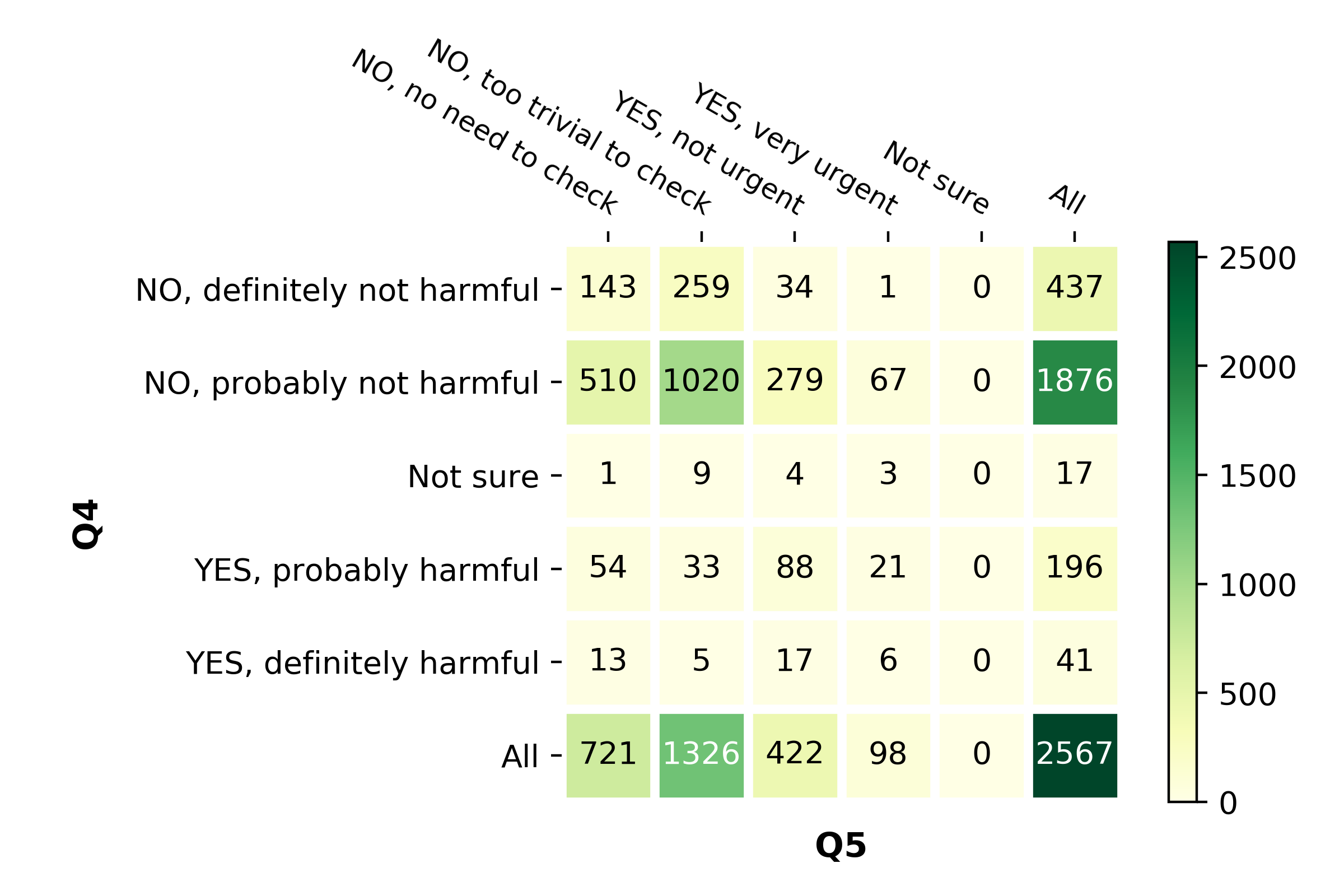}
        \caption{Heatmap for Q4 and Q5.}
        \label{fig:bulgarian_contingency_table_q4_q5}    
    \end{subfigure}     
    \hfill
    \begin{subfigure}[b]{0.6\textwidth}    
        \includegraphics[width=\textwidth]{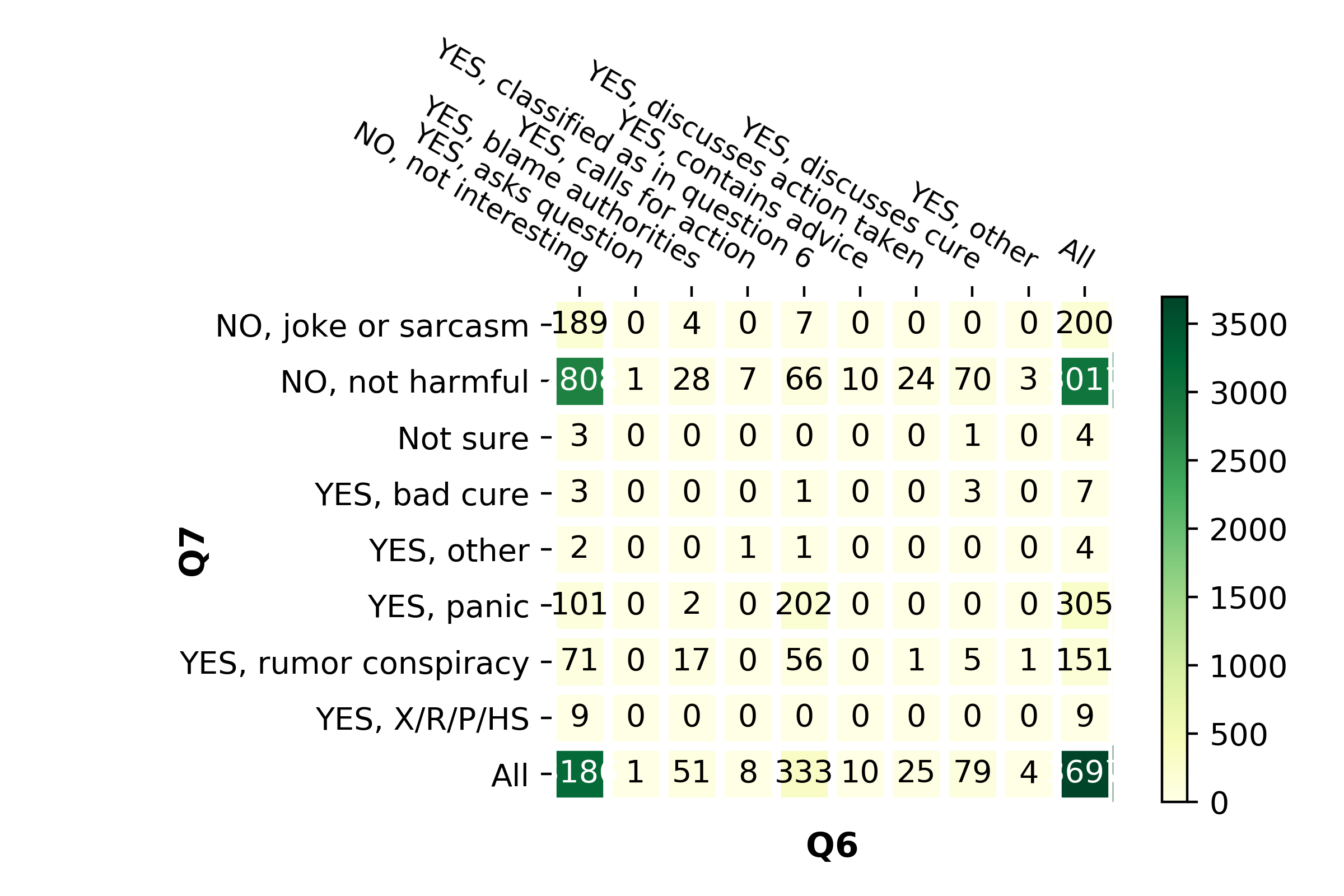}
        \caption{Heatmap for Q6 and Q7. YES, X/R/P/HS -- YES, xenophobic, racist, prejudices or hate speech}
        \label{fig:bulgarian_contingency_table_q6_q7} 
    \end{subfigure}
    \caption{Contingency and correlation heatmaps for the \textbf{Bulgarian tweets} for different question pairs.}
    \label{fig:bulgarian_contingency_all}
\end{figure*}

\begin{figure*}[tbh]
\centering
    \begin{subfigure}[b]{0.4\textwidth}
        \includegraphics[width=\textwidth]{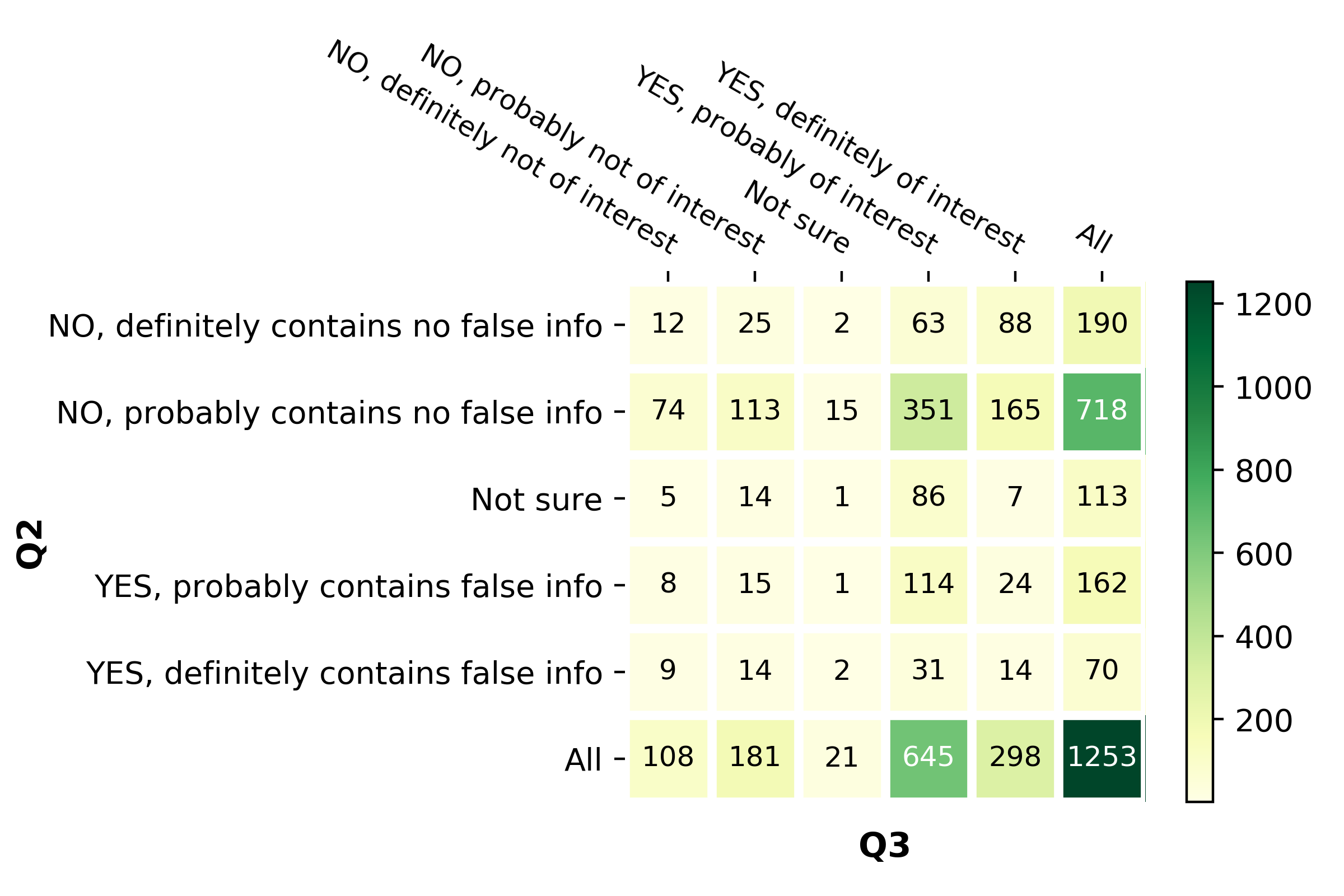}
        \caption{Heatmap for Q2 and Q3.}
        \label{fig:dutch_contingency_table_q2_q3}
    \end{subfigure}%
    \begin{subfigure}[b]{0.4\textwidth}    
        \includegraphics[width=\textwidth]{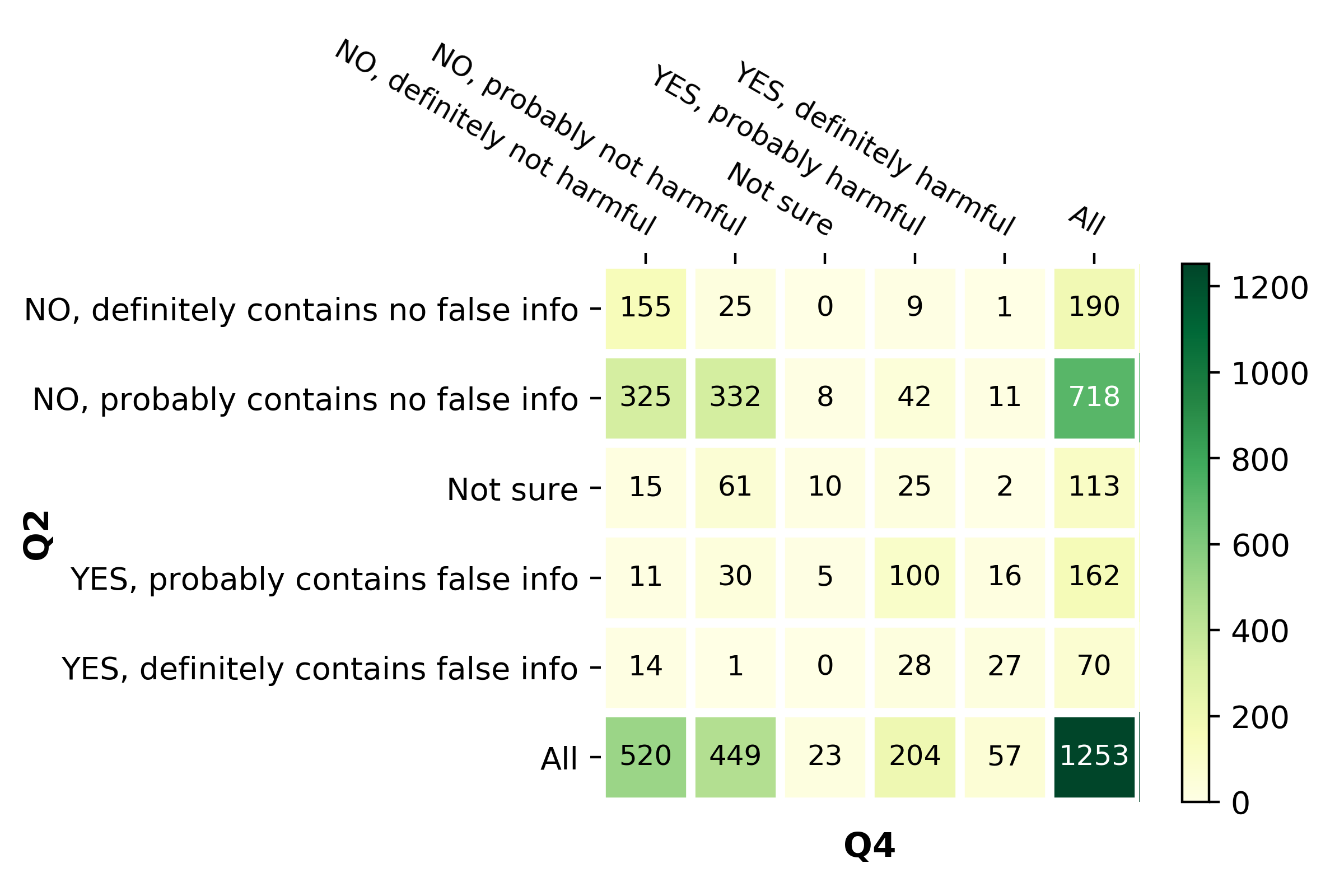}
        \caption{Heatmap for Q2 and Q4.}
        \label{fig:dutch_contingency_table_q2_q4}    
    \end{subfigure} 
    \hfill
    \begin{subfigure}[b]{0.4\textwidth}    
        \includegraphics[width=\textwidth]{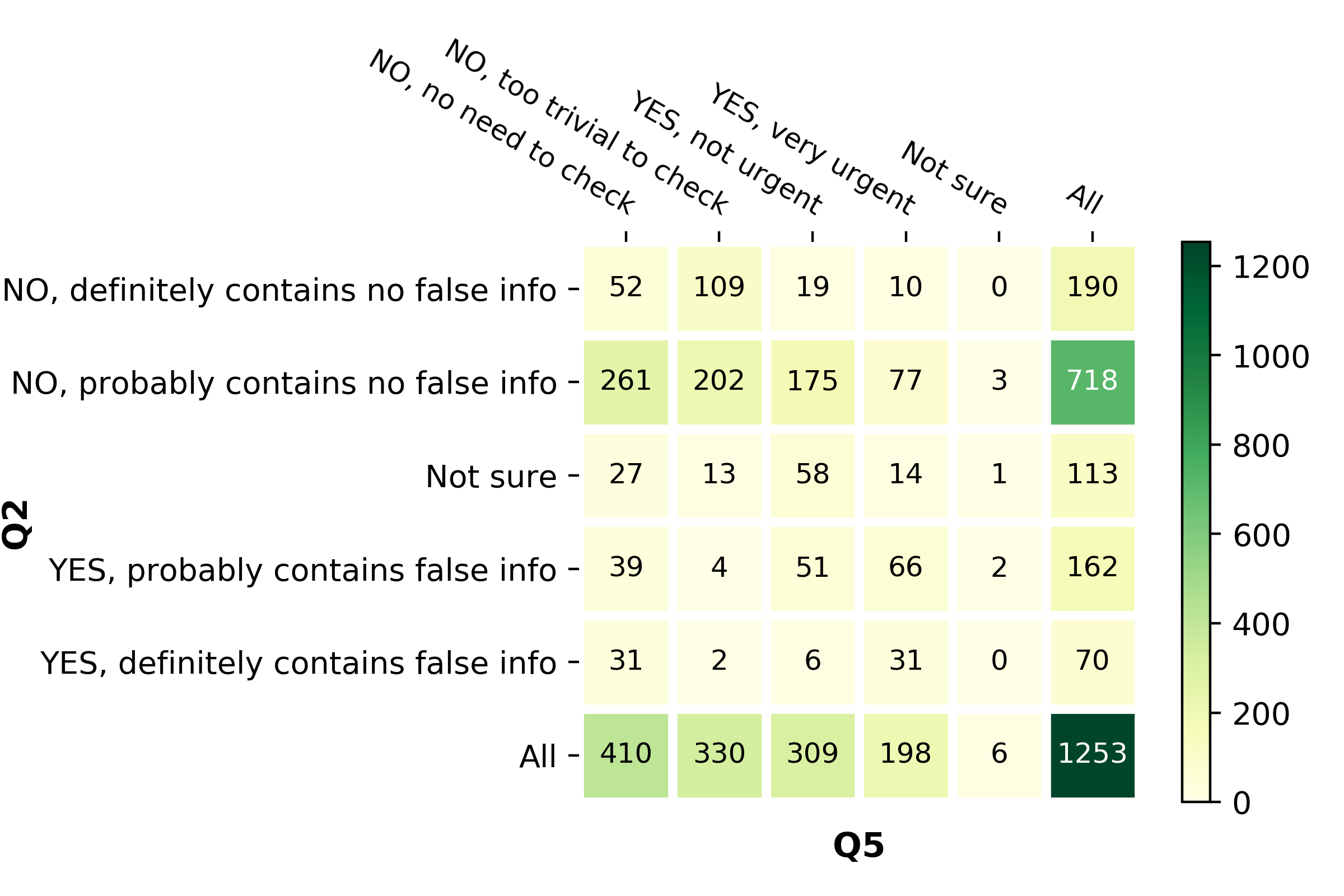}
        \caption{Heatmap for Q2 and Q5.}
        \label{fig:dutch_contingency_table_q2_q5}    
    \end{subfigure}%
    \begin{subfigure}[b]{0.4\textwidth}    
        \includegraphics[width=\textwidth]{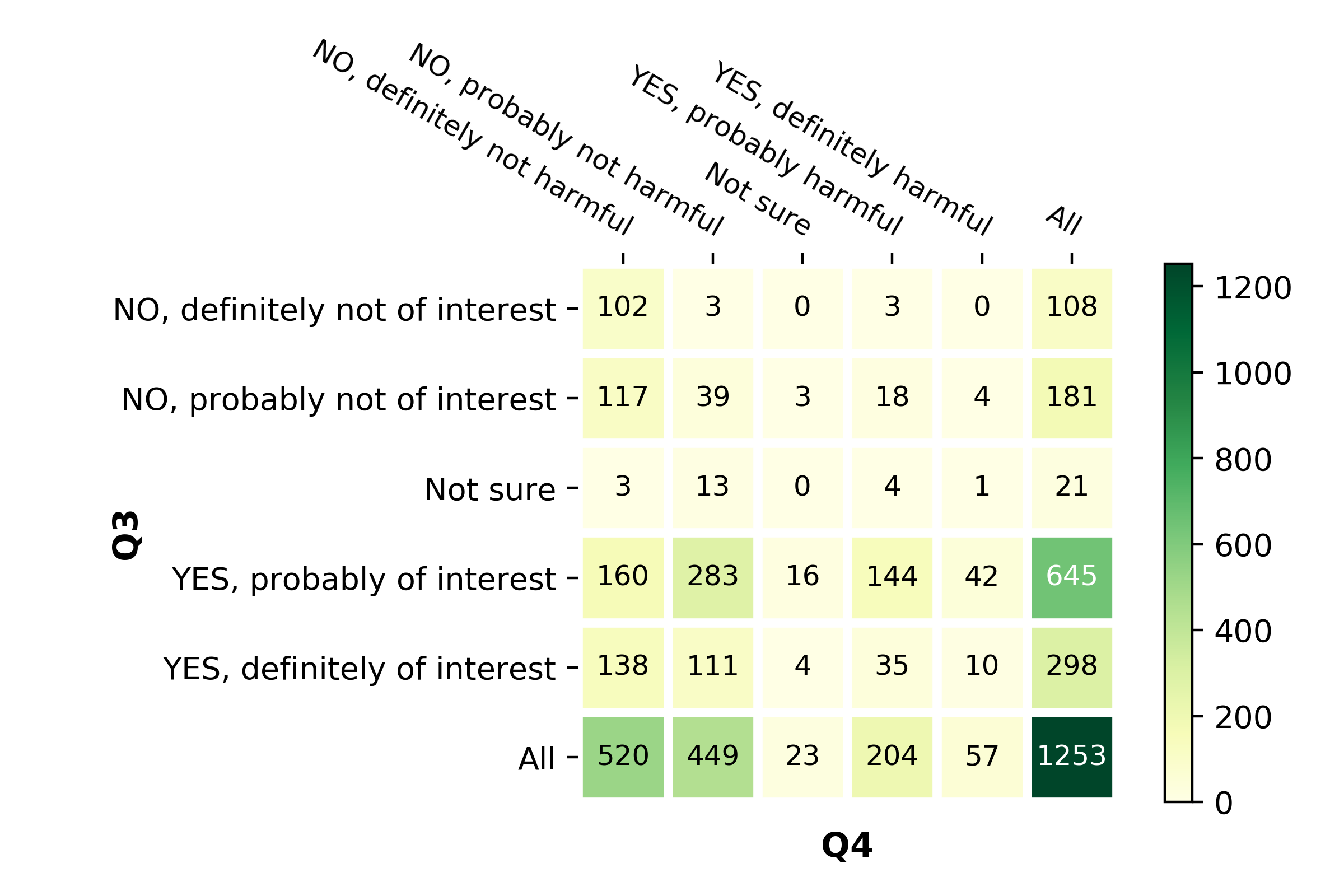}
        \caption{Heatmap for Q3 and Q4.}
        \label{fig:dutch_contingency_table_q3_q4}    
    \end{subfigure}
    \hfill
    \begin{subfigure}[b]{0.4\textwidth}    
        \includegraphics[width=\textwidth]{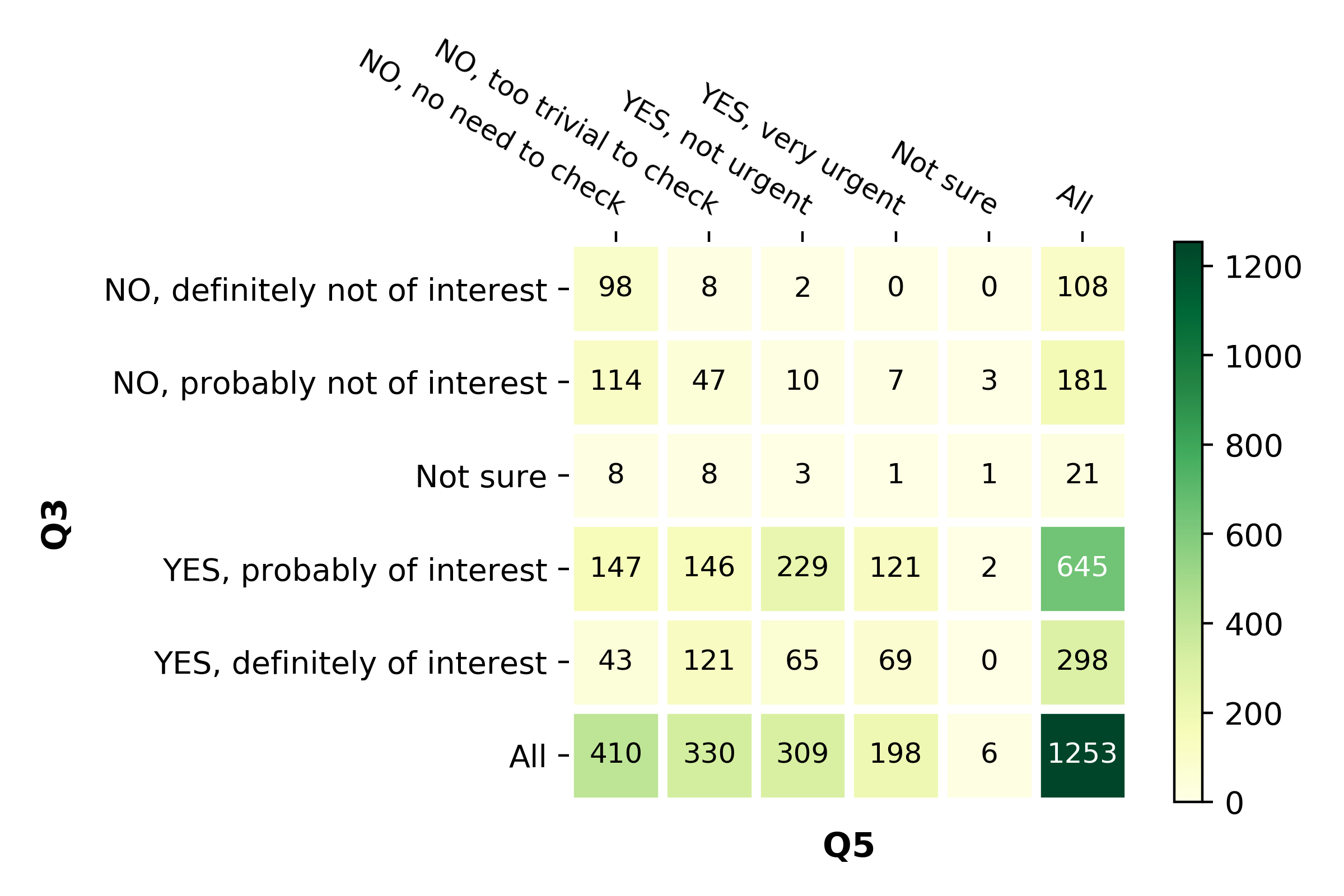}
        \caption{Heatmap for Q3 and Q5.}
        \label{fig:dutch_contingency_table_q3_q5}    
    \end{subfigure}%
    \begin{subfigure}[b]{0.4\textwidth}    
        \includegraphics[width=\textwidth]{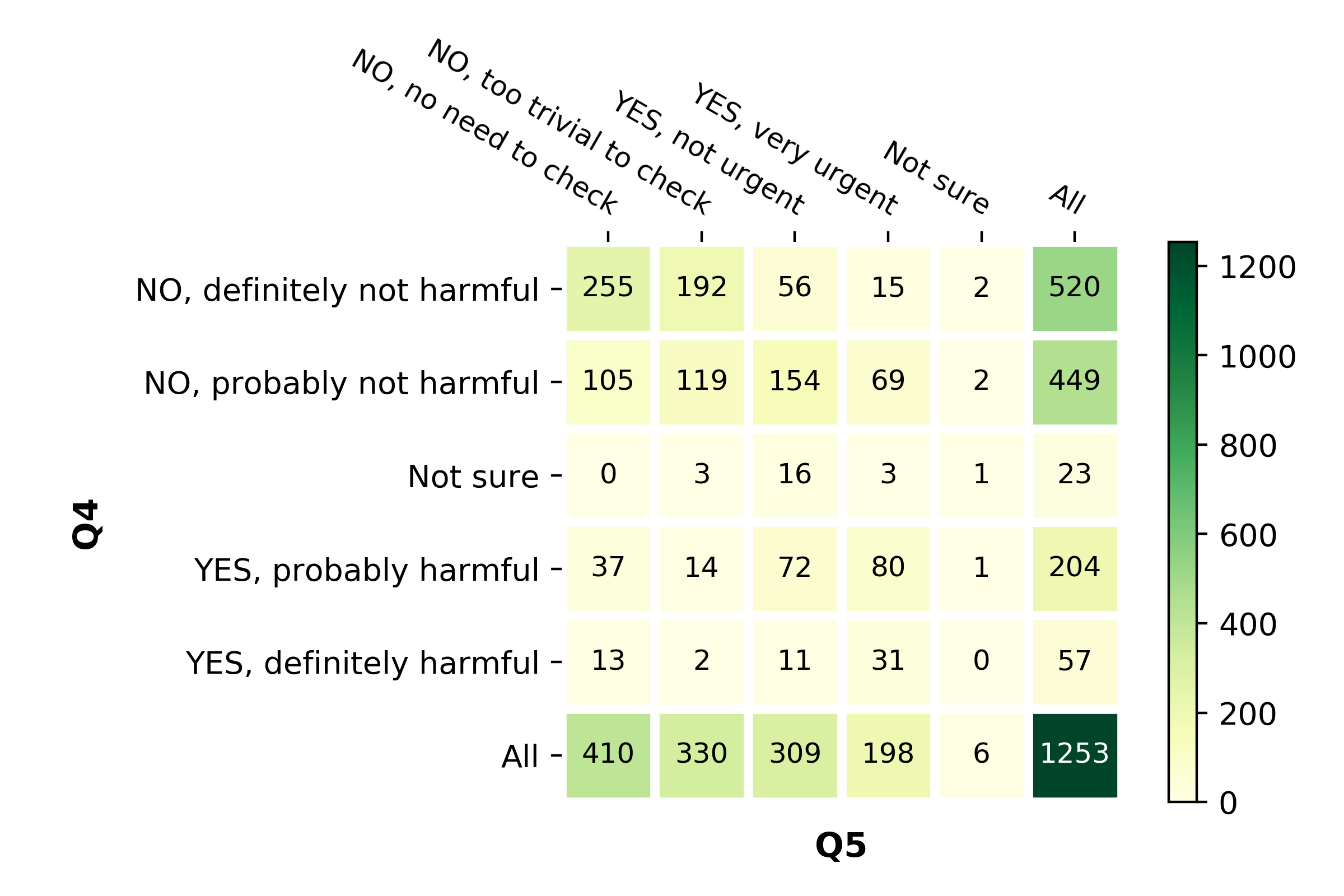}
        \caption{Heatmap for Q4 and Q5.}
        \label{fig:dutch_contingency_table_q4_q5}    
    \end{subfigure}     
    \hfill
    \begin{subfigure}[b]{0.6\textwidth}    
        \includegraphics[width=\textwidth]{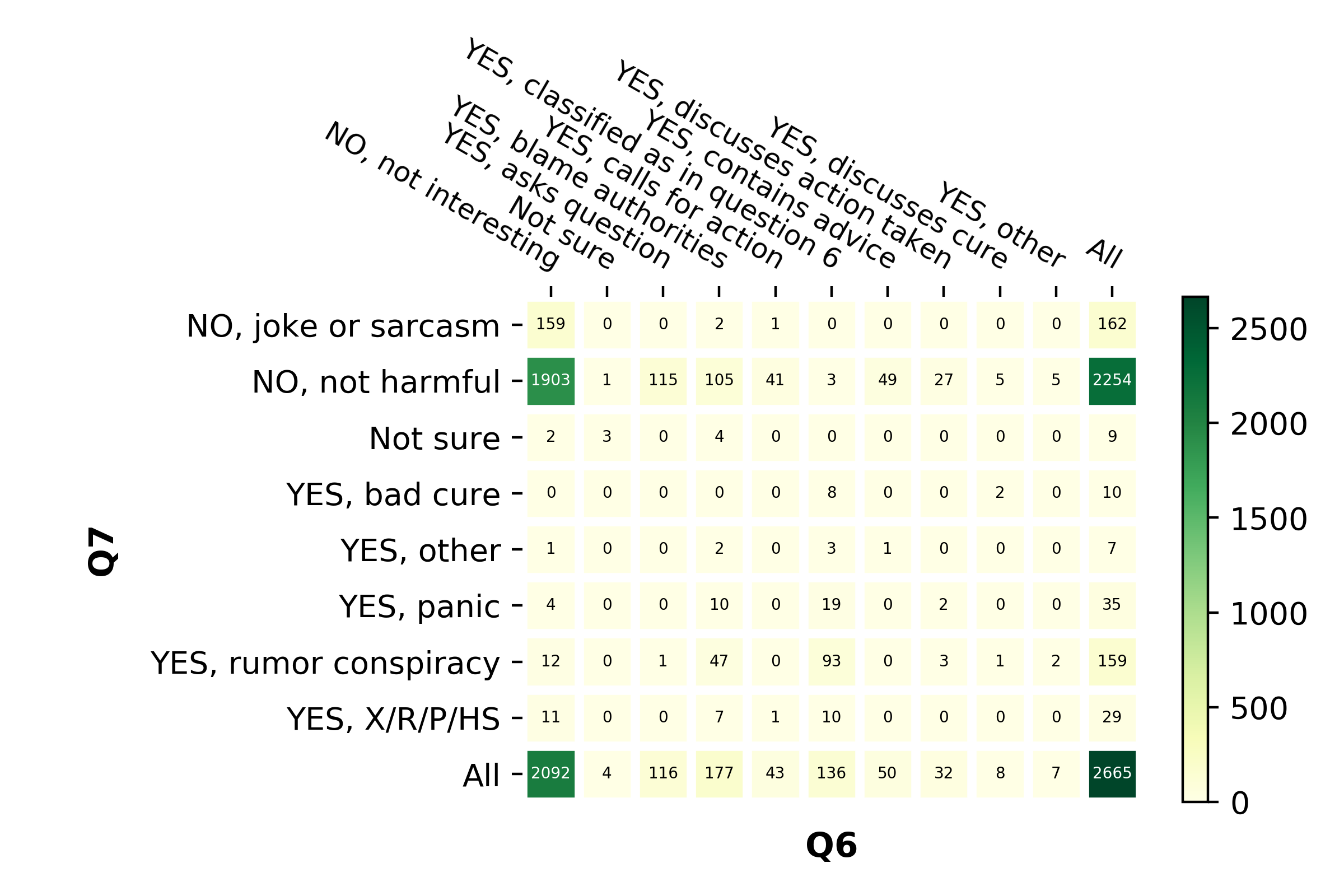}
        \caption{Heatmap for Q6 and Q7. YES, X/R/P/HS -- YES, xenophobic, racist, prejudices or hate speech}
        \label{fig:dutch_contingency_table_q6_q7} 
    \end{subfigure}
    \caption{Contingency and correlation heatmaps for the \textbf{Dutch tweets} for different question pairs.}
    \label{fig:dutch_contingency_all}
\end{figure*}

Finally, we study the correlation between the labels for different questions across for each of the four languages.

\subsection{English Tweets}

Figure \ref{fig:contingency_all} shows contingency and correlation tables in a form of a heatmap for different question pairs obtained from the English tweet dataset. For questions Q2-3, it appears that there is a high association between ``\dots no false info'' and the general public interest as shown in Figure~\ref{fig:contingency_table_q2_q3}. For questions Q2 and Q4
(Figure \ref{fig:contingency_table_q2_q4}), strong association can be observed between ``\dots~no false info'' and ``\dots~not harmful'' (86\%) compared to ``harmful'' (13\%) for either an individual, a products or government entities. By analyzing questions Q2 and Q5 (Figure \ref{fig:contingency_table_q2_q5}), we conclude that ``\dots no false info'' is associated with either ``no need to check'' or ``too trivial to check'', highlighting the fact that a professional fact-checker does not need to spend time on them. From questions Q3 and Q4 (Figure \ref{fig:contingency_table_q3_q4}), it appears that when the content of the tweets is ``not harmful'' the general public interest is higher (74\%) than when it is ``harmful'' (22\%).
From question Q3 and Q5 (Figure \ref{fig:contingency_table_q3_q5}), we see an interesting phenomenon, namely that tweets of high general public interest have a greater association with a professional fact-checker having to verify them (22\%) compared to either ``too trivial to check'' or ``no need to check'' (78\%). Questions Q4 and Q5 (Figure \ref{fig:contingency_table_q4_q5}) show that ``harmful'' tweets require an attention (69\%) from a professional fact-checkers than ``not harmful'' tweets (30\%). Our findings for Q6 and Q7 (Figure \ref{fig:contingency_table_q6_q7}) suggest that the majority of the tweets are not harmful to society, which also requires less attention from government entities. The third most common tweet label for Q7 blames the authorities, even though they are mostly not harmful for society.    

We computed the correlation using the Likert scale values (i.e., 1-5) that we defined for these questions. We observed that overall Q2 and Q3 are negatively correlated, which suggests that if the claim contains no false information, it is of high interest to the general public. This can be also observed in Figure~\ref{fig:contingency_table_q2_q3}. Questions Q2 and Q4 exhibit positive correlation, which might be due to their high association with ``\dots no false info'' and ``\dots not harmful''.

\subsection{Arabic Tweets}

Figure \ref{fig:arabic_contingency_all} shows similar heatmaps for the Arabic tweets. For Q2 and Q3 (Figure \ref{fig:arabic_contingency_table_q2_q3}), we can observe that the association between ``\dots contains no false info'' and general public interest is higher (76\%) than ``\dots contains false info'' (23\%). From questions Q2 and Q4 (Figure \ref{fig:arabic_contingency_table_q2_q4}), we can conclude that ``\dots contains no false info'' is associated with ``\dots not harmful'' and ``\dots contains false info'' is associated with ``\dots harmful''.
From the relation between Q2 and Q5 (Figure \ref{fig:arabic_contingency_table_q2_q5}), we can observe that in the majority of the cases ``\dots contains no false info'' is associated with either ``no need to check'' or ``too trivial to check'', which means that a professional fact-checker does not need to verify them. The analysis between questions Q3 and Q4 suggests that the general public interest is higher when the content of the tweets is not harmful (79\%) than when it is harmful (21\%) (Figure \ref{fig:arabic_contingency_table_q3_q4}). From questions Q3 and Q5, we can observe that the general public interest is higher when the claim(s) in the tweets are either ``no need to check'' or ``too trivial to check'' (Figure \ref{fig:arabic_contingency_table_q3_q5}). 
The analysis between question Q4 and Q5 shows that ``not harmful'' tweets are either ``no need to check'' or ``too trivial to check'' by a professional fact-checker (Figure \ref{fig:arabic_contingency_table_q4_q5}). From questions Q6 and Q7, we notice that in the majority of the cases the tweets are not harmful for society and hence they are not interesting for government entities (Figure \ref{fig:arabic_contingency_table_q6_q7}).  

\subsection{Bulgarian and Dutch Tweets}
Figures \ref{fig:bulgarian_contingency_all} and \ref{fig:dutch_contingency_all} show the same kinds of heatmaps for the Bulgarian and the Dutch datasets, respectively. The observations are very similar.

\clearpage
\newpage
\section{Geographical Distribution: English and Arabic}
\label{sec:appendix_geo_dist}

\begin{figure}[tbh]
\centering
    \begin{subfigure}[b]{0.45\textwidth}
        \includegraphics[width=\textwidth]{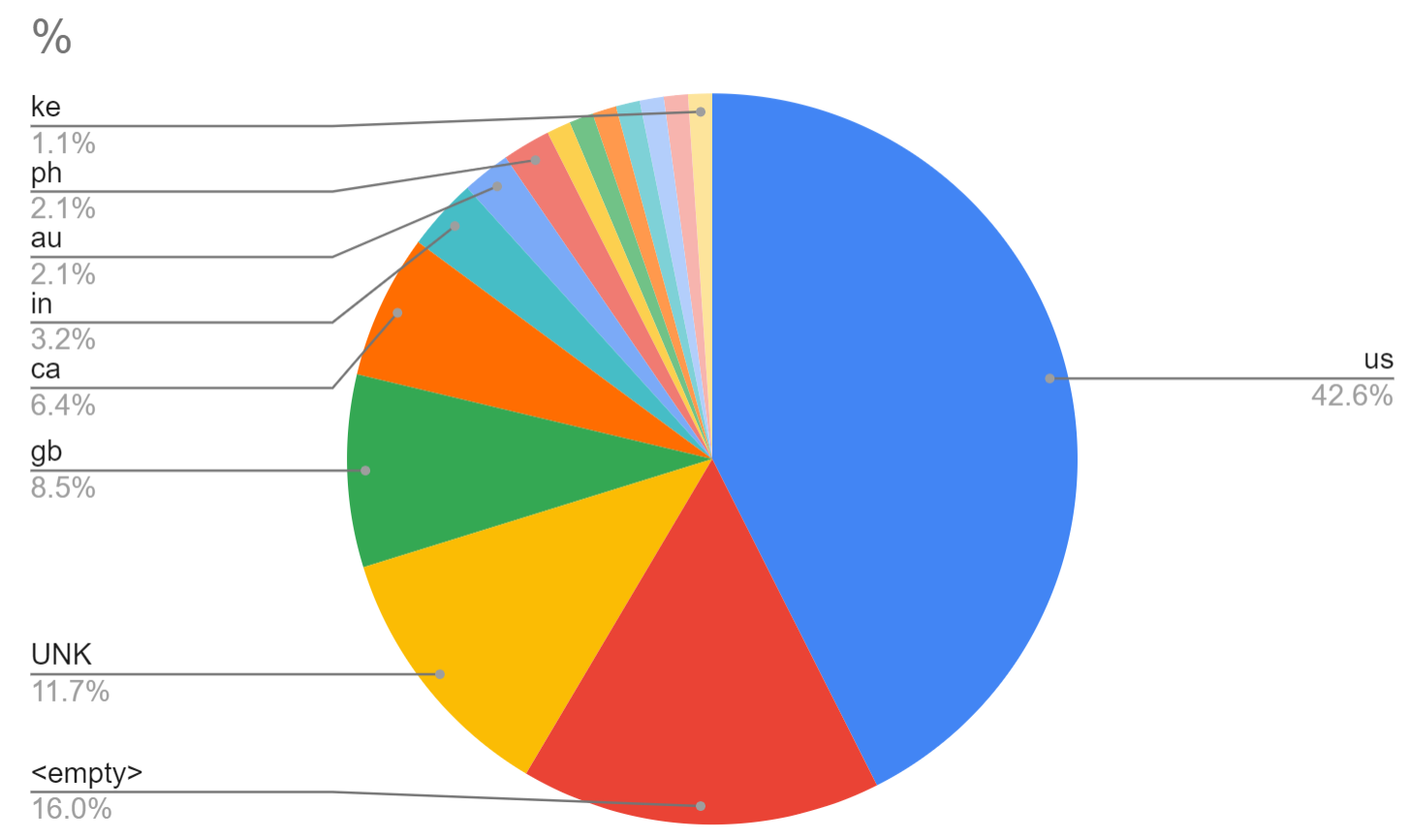}
        \caption{English dataset}
        \label{fig:countries_en}
    \end{subfigure}
    \begin{subfigure}[b]{0.45\textwidth}    
        \includegraphics[width=\textwidth]{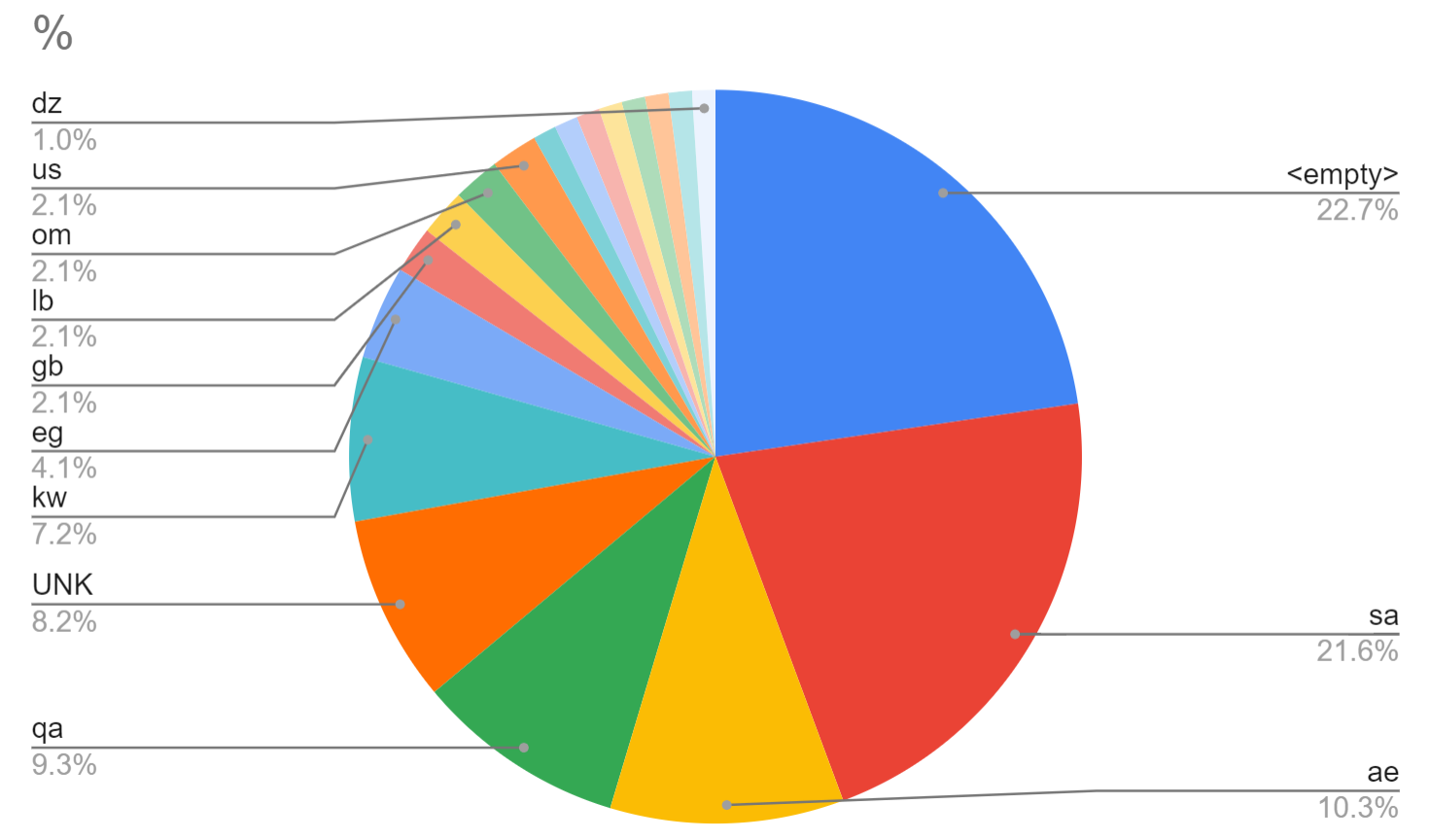}
        \caption{Arabic dataset}
        \label{fig:countries_en_ara}    
    \end{subfigure} 
\caption{Distribution by country for English and Arabic tweets.}
\label{fig:tweets_country_dist}
\end{figure}

Figure \ref{fig:tweets_country_dist} shows the geographical distribution of the annotated tweets for English and Arabic. We consider the country of the tweet author or the original author in case of a retweet. We can see that most English tweets come from USA, UK, Canada, and India, while most Arabic tweets come from the Gulf region (KSA, UAE, Qatar, and Kuwait). Yet, for both languages, we have tweets from multiple countries, which means that there is good diversity of interests, topics, style, etc.

We did not perform this analysis for Bulgarian and Dutch, as these are less international languages, and their speakers are mostly concentrated in Bulgaria and the Netherlands, respectively.

\clearpage
\newpage

\begin{figure*}[tbh]
\centering
\includegraphics[width=\textwidth]{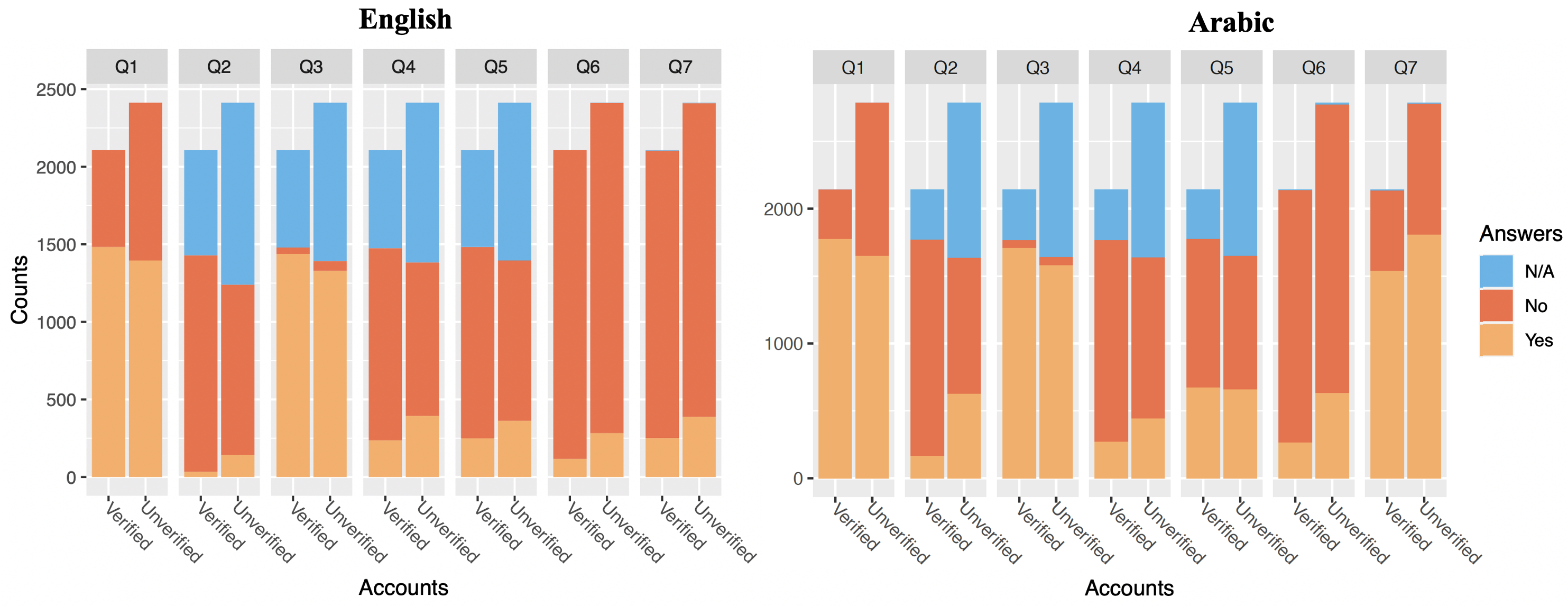}
\caption{Verified vs. non-verified account distribution for English and Arabic across the different questions. NA refers to tweets that have not been labeled for those questions, they are identical to the tweets categorized with the label NO in Q1.}
\label{fig:user_account_label_dist}
\end{figure*}

\section{Verified and Unverified Accounts: English and Arabic}
\label{sec:appendix_verified_unverified_account}
We study the correlation between tweet labels and whether or not the original author of a tweet has a verified account. Verified accounts include such by government entities, public figures, celebrities, etc., which have a large number of followers, and thus their tweets typically have higher impact.

Figure \ref{fig:user_account_label_dist} shows that verified accounts tend to post more tweets that contain factual claims than unverified accounts (Q1), and their tweets are less likely to contain false information (Q2), are more likely to be of interest to the general public (Q3), and are less likely to be harmful (Q4 and Q6). 

Based on this study, we have added corresponding features for our models.

\clearpage
\newpage

\begin{figure*}[tbh]
\centering
\includegraphics[width=\textwidth]{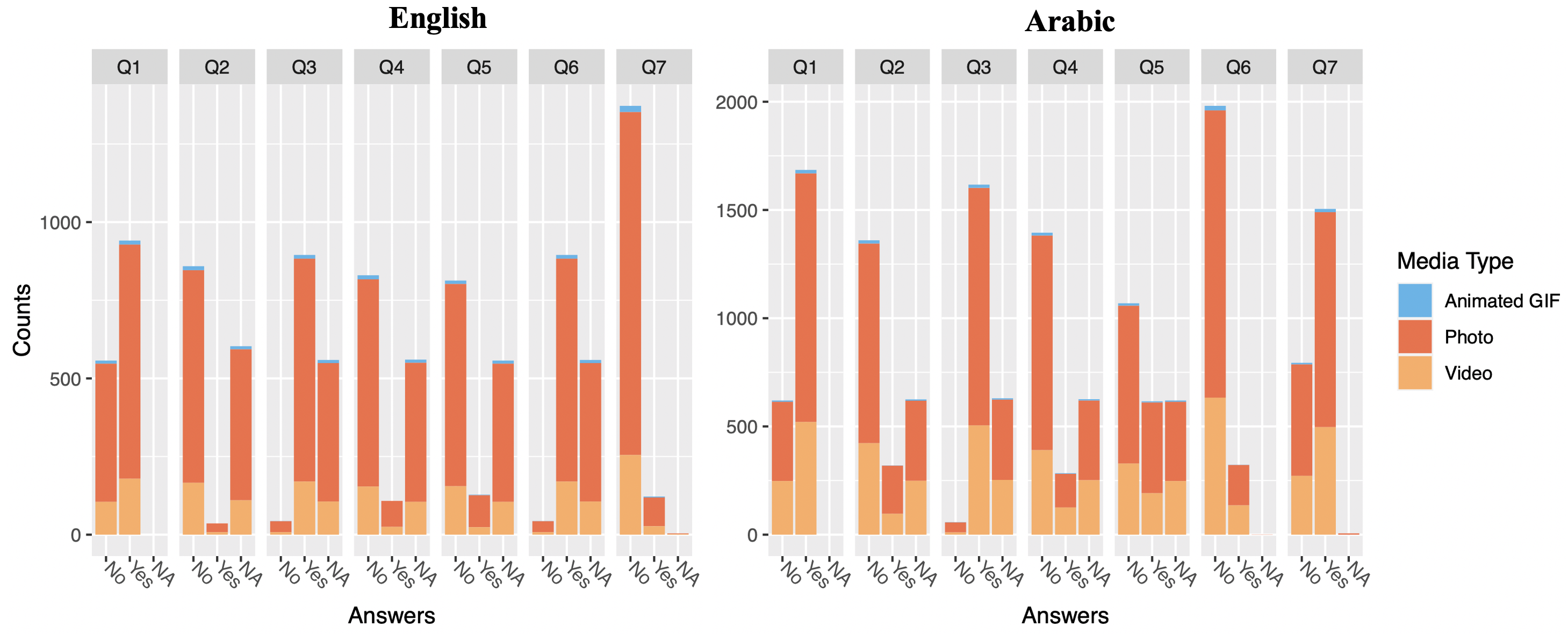}
\caption{Distribution of media types in English and Arabic tweets.}
\label{fig:media_dist}
\end{figure*}

\section{Multimedia in Tweets: English and Arabic}

\label{sec:appendix_multimedia_in_tweets}
In this subsection, we study the correlation between whether a tweet contains multimedia (image or video) and the annotation labels. Generally, people trust videos more than images or plain texts, which suggests that tweets with video could have a higher impact. Figure~\ref{fig:media_dist} shows the distribution of media types for English and Arabic. 

We can see that if a tweet contains multimedia content, it is likely to contain a factual claim (Q1), to have a higher impact to the general public (Q3), but it is less likely to contain false information (Q2) or to be harmful to the society (Q4). These observations in part motivated us to model the use of multimedia as part of our features.

\end{document}